\documentclass{article}

\usepackage{arxiv}

\usepackage[utf8]{inputenc} % allow utf-8 input
\usepackage[T1]{fontenc}    % use 8-bit T1 fonts
\usepackage{hyperref}       % hyperlinks
\usepackage{url}            % simple URL typesetting
\usepackage{booktabs}       % professional-quality tables
\usepackage{amsfonts}       % blackboard math symbols
\usepackage{nicefrac}       % compact symbols for 1/2, etc.
\usepackage{microtype}      % microtypography
\usepackage{lipsum}
\usepackage{graphicx}
\graphicspath{ {./images/} }

\usepackage{amsmath}

\title{Spatio-temporal stochastic graph-based learning for infectious disease forecasting}

\author{
 Luz Stefani Sotomayor Valenzuela, Susanna Cramb, Darren Wraith\\
 \textit{School of Public Health and Social Work, Queensland University of Technology, Brisbane, Australia}\\
 \textit{QUT Centre for Data Science, Queensland University of Technology, Brisbane, Australia}
 }

\begin{document}
\maketitle
\begin{abstract}
Spatio-temporal graph-based models have typically been used to forecast new cases of infectious diseases such as COVID-19 and chickenpox outbreaks. However, the use of stochastic modelling into their learning process has been surprisingly under-investigated and rarely considered entire data sets of large countries. As a result, it is unknown whether these models would provide accurate forecasts in real-world disease spread scenarios. In this work, we propose a spatio-temporal stochastic graph-based architecture that integrates a stochastic formulation and uncertainty approximation process to forecast new infectious disease cases. We find that our approach can adapt to encode large and small population geographical networks within a single model architecture. Using two real-world data sets, COVID-19 in the US and chickenpox in Hungary, we report an enhanced effect of the proposed architecture across predictions of the 2022 first wave for COVID-19 in the US and comparative results of chickenpox waves during 2012-2014 in Hungary. By benchmarking with four spatio-temporal graph-based models, quantitative results show competitive overall weekly performance of the proposed approach on forecasting new cases for all 3,218 US counties and all 20 Hungary counties. The proposed approach can represent overall epidemic progression relative to baselines, though with a one-step delay; while exhibiting a reduced sensitivity to high-frequency and low-amplitude variability.
\end{abstract}

% keywords can be removed

\section{Introduction}
\label{sec:4intro}

Epidemic forecasting models are paramount tools for understanding and mitigating the impact of infectious disease spread in populations around the world. Recent global epidemics have highlighted the need for robust architectures capable of modelling single and nonlinear distributions while adapting to rapidly evolving epidemiological patterns. These complex factors illustrate that the spread of infectious diseases also depends on the variability and heterogeneity of human behaviour distributed across interconnected regions \cite{kaw21}, where features such as stochasticity and spatio-temporal dependencies can benefit the epidemic forecasting task.

The development of effective epidemic models is of great significance for representing the dynamics of disease spread across fine- and coarse-grained landscapes to analyse them at multiple scales. In epidemic forecasting, knowledge-driven approaches have focused on fitting the spatio-temporal features of infectious disease spread by calibrating the model parameters and using mechanistic structures \cite{chaSyd20,koo2020,phamVT21,liuwan2024}. However, their ability is limited to not being able to capture nonlinear relationships in the data, depending on fixed assumptions, or scaling to large geographical networks. Although these models can be used in certain contexts, they often require extensive manual fine-tuning and are computationally intensive.

In recent years, data-driven approaches based on machine learning and deep learning methods have become promising tools for the prediction of epidemics \cite{shorten21}. Advances in graph-based learning models have enabled the learning of geographical network structures and sequential data relationships along with other characteristics of the disease. These structures use graph neural networks (GNNs) \cite{scar09, wuz20}, which learn the underlying spatio-temporal dependencies while enhancing prediction accuracy.

Their epidemiological applications can be categorised into three main streams of models based on: i) message passing neural networks (MPNNs) \cite{pan21}; ii) graph convolutional neural networks (GCNs) \cite{fritz22,kap20,wang22,yu23}; and iii) graph attention networks (GATs) \cite{gao21,mur21}.

MPNNs leverage the graph discrimination property to learn network structures. Their complex message-passing schemes are suitable for varying types of node and edge attributes. GCNs use convolution operations with graph filters to learn irregular spatial distributions and offer robustness to small perturbations \cite{zul20} and their features are well-suited for nodes and edges that have the same type of attributes. GATs include the learning of attributes of the graph edges - when the data are available - and computes attention coefficients for each edge. In terms of interpretability, MPNNs and GATs may be more challenging to interpret due to their more complex encoding mechanisms and architectures, while GCN-based models offer better interpretability and scalability to make predictions as they focus on local neighbourhood aggregation, which can support a stable learning process of disease spreading evolution.

Despite the useful properties of GNNs, the intrinsic variability of disease spread in the human population has rarely been included in the learning process of such architectures \cite{dav21,lagatta20,li21}. In real-world epidemiological networks, the trajectory of disease spreading behaviour may diverge due to variability in individual-level characteristics or random events across populations \cite{kaw21}. These random changes lead to stochastic effects in the epidemiological learning system; hence, leading to random prediction outcomes.

While it is important to encompass the GNNs spatio-temporal learning with stochastic, random processes, its practical implementation also involves different scales and resolutions of the geographical extent at which the evolving epidemiological patterns are taking place. Many GNN-based works consider either small or medium size geographical networks, at state- or county-level. For instance, forecasting of COVID-19 is often studied with geographical networks at county- or state-level that include approximately 20 to 400 regions \cite{dav21,lagatta20,fritz22,gao21,kap20,mur21,pan21,yu23}. Few works analyse large geographical networks within a country, e.g. forecasting of COVID-19 spread in the US at a county-level is generally studied in a network of more than 1,300 counties \cite{wang22,li21,guo21,spa21}. However, these works typically cover training and testing periods that exhibit relatively limited variability of the disease spreading trajectory. And such scenarios involve predictable patterns that facilitate the modelling process.

In complex epidemiological networks, the trending behaviours of disease spread are characterised by periods of low and high complexity. That is, spatio-temporal graph-based learning models should be implemented over high variable epidemic trajectories in the testing phase that are mismatched with the relatively consistent trajectories with minimal fluctuations used in the training phase, therefore leading to a more challenging forecasting process. In this work, we propose a spatio-temporal stochastic graph-based learning model for infectious disease forecasting to account for the temporal evolution of highly and less variable disease spread over smaller and larger geographical networks. We analyse the mismatch of complex disease spreading behaviours between the testing and training stages and the multilocality progression of the disease spread over time using minimally processed data. To characterise the sequential evolution of disease spread, the temporal epidemic data are organised as features of the graph nodes distributed across the geographical network. Since the disease spread randomness arises due to external factors, the model is trained on stochastic outcomes. We estimate uncertainty using ensembles to simulate multiple possible trajectories of predictions.

This research is inspired by the complex nature of real-world epidemiological networks. We use epidemic and geographical data to analyse entire, multi-locality sequential patterns of disease evolution in heterogeneous settings. These neighbourhoods make the learned model robust to disease spread fluctuations encountered during testing. We also develop ablation studies to numerically analyse the associated variability interval of the model predictions and the influence of the model components. Furthermore, this architecture requires a minimal set of internal trainable parameters that minimise the propagation of uncertainty through the model. The main contributions of this paper are:

\begin{itemize}
    \item We design a spatio-temporal stochastic graph-based learning architecture for infectious disease forecasting. The architecture is based on a GCNs layer that accounts for epidemic time-varying progression, geographical distribution, and a stochastic formulation layer that generates stochastic predictions used during learning. We provide two versions of this approach: (1) a PrGCN${_{rmsle}}$ version based on a GCNs layer; and (2) a PrGLSTM${_{rmsle}}$ version based on GCNs and recurrent layers.
    \item We apply the proposed model in a real-world large epidemiological network where the dynamics of disease spread exhibit highly variable heterogeneity over time and space. We analyse the long-time dynamics of COVID-19 spread in the US regions, for a period that includes 690 days from January 22nd 2020 to December 11th 2021 for training and 297 days from December 12th 2021 to October 4th 2022 for testing. The multi-locality analysis involves a geographical network of 3,218 counties, and the simultaneously stochastic predictions of weekly new disease cases at county-level.
    \item We also apply the proposed model in a real-world relatively small epidemiological network where disease transmission dynamics exhibit variability over time and space.\\
    We analyse the long-time dynamics of Chickenpox spread in Hungary, for a period that includes 7 years from January 3rd 2005 to January 2nd 2012 for training. Testing phase includes 3 years from January 3rd 2012 to December 29th 2014. The multi-locality analysis involves a geographical network of 20 counties, and the simultaneously stochastic predictions of weekly new disease cases at county-level.
    \item We report weekly and global performances of the proposed model in comparison to four spatio-temporal graph-based learning models, and ablation models. These benchmarking and ablation analyses identify that our proposed model achieves competitive results in high- and low-complexity epidemic trajectories. Each model is run 100 times for uncertainty estimation, and performance results are reported in five metrics: mean squared error (MSE), mean absolute error (MAE), Huber error, root mean square error (RMSE), and root mean square logarithmic error (RMSLE).
\end{itemize}

The paper is organised as follows. Section \ref{sec:4math} presents our proposed spatio-temporal stochastic graph-based learning model for infectious disease forecasting, components, and experiments in detail. Sections \ref{sec:4find} and \ref{sec:4dis} outline the experimental results and the discussion with related work. Finally, Section \ref{sec:4fut} provides the conclusion of this study, limitations, and future directions.

\section[Methods]{Methods}
\label{sec:4math}

\begin{figure}%[!h]
    \centering
    
        \includegraphics[width=0.6\columnwidth]{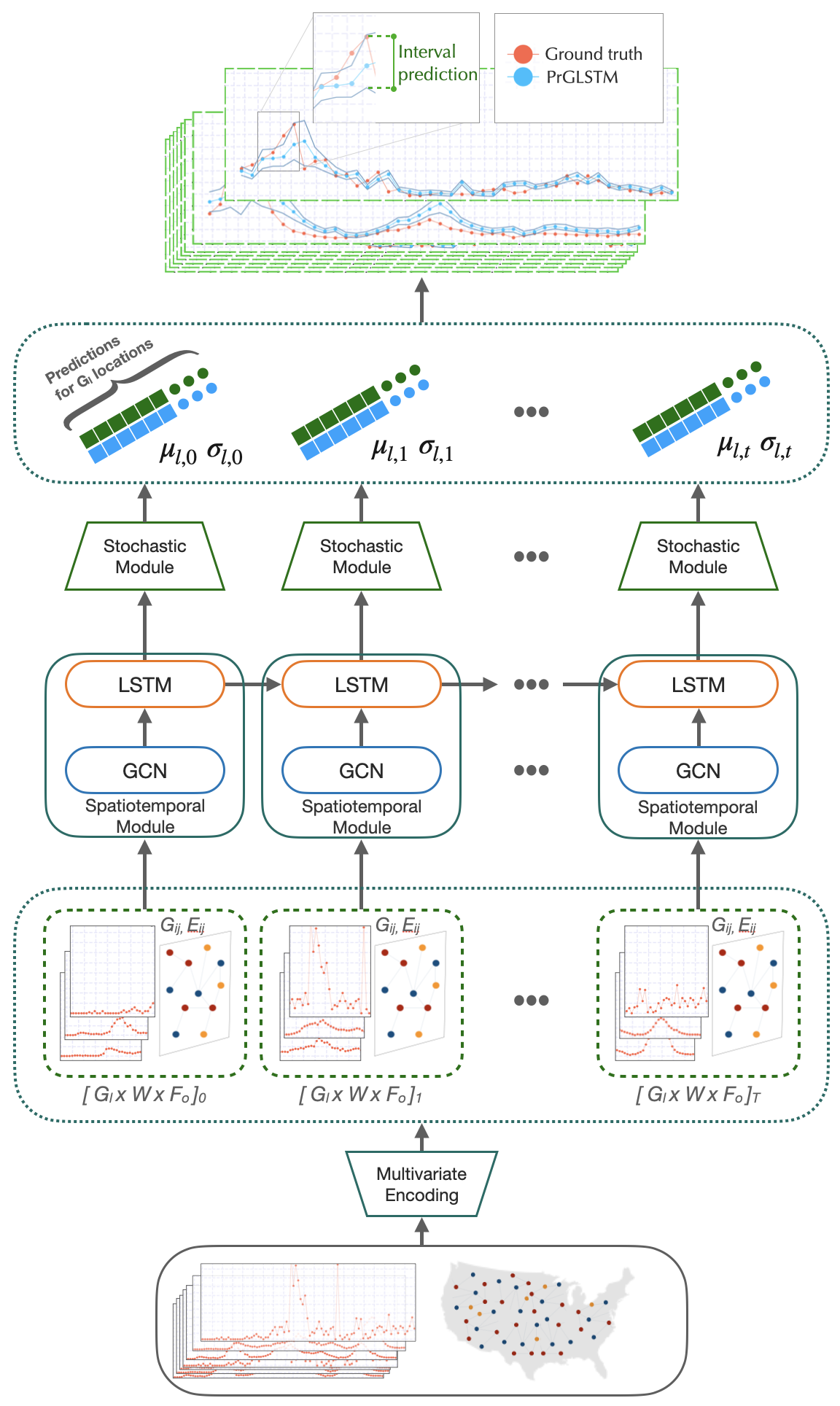}

    \caption{Overview of the proposed architecture (PrGLSTM${_{rmsle}}$ version), illustrating the interaction between Multivariate Sequence Encoding, Spatio-temporal Processing, and Stochastic modules, and the associated information flow. The input features consist of disease incidence and geographic network, using the United States (US) as an example.}
    \label{fig4:framework}
\end{figure}

The proposed architecture consists of three modules in sequential order, where the output of a previous layer serves as the input of the next. These components collectively enable modelling and analysing system relationships without involving numerous layers and activation functions. Figure \ref{fig4:framework} summarises the main modules where the first module is a multivariate sequence encoding, which uses normalisation and data iterator functions to standardise the information of geographical connections and cases of disease spread. This module organises relevant features of given temporal multi-inputs in a static spatial graph domain. The second module is a spatio-temporal mechanism to extract and correlate the latent features from the data. The third module implements a stochastic method, which consists of a fully connected layer with normal distribution functions. Randomness manifests itself as a diversity of outcomes from internal distributional projections. The final output of these modules is a set of predictions, where each estimate corresponds to a given time and location. The proposed architecture provides a model $\mathfrak{M}$ that learns a complex dynamical representation based on spatial and time-series data. Further methodological details are provided in the Appendix \ref{app4:methods}.

\subsection{Multivariate sequence encoding}
This module plays a significant role in supporting global comparability between locations and building the temporal and spatial dimensions of a complex system, where disease cases are correlated with the geographical topology of the region under study. Initially, the temporal data are preprocessed to mitigate large negative values and missing data. Raw time-series data of disease incidence are scaled to the range $[0,1]$. The scaling standardises the temporal patterns and allows a stable multivariate learning process. Iteration functions create lagged sequence samples from the scaled data. Considering all locations simultaneously, $X^{(t)} \in \mathbb{R}^{G_{l} \times W \times 1}$ at a given time step $t$ is a single lagged sequence sample with dimensions $(G_{l} \times W \times 1)$, where $G_{l} \in \mathbb{N}$ denotes the number of locations under study (or nodes), $W \in \mathbb{N}$ is the temporal observation window size, and $1$ is the number of features per location. Each input sample contains $W$ data steps of observations for all $G_{l}$ locations, with $1$ feature recorded at each location. Elements of the lagged sequence sample are denoted as $X^{(t)}_{i,s,1} \in \mathbb{R}$, where $i \in \{ 1, \dots, G_l \}$ identifies a specific node or geographic location, $s \in \{ 1, \dots, W \}$ identifies a time step within the temporal observation window, and $1$ denotes the feature index of $X^{(t)}$.

Each temporal snapshot consists of a spatio-temporal tensor $X^{(t)} \in \mathbb{R}^{G_{l} \times W \times 1}$ and the corresponding target $Y^{(t)} \in \mathbb{R}^{1\times G_l}$, paired with the same edge index tensor $G \in \mathbb{N}^{2 \times c}$ and edge attribute vector $E \in \mathbb{R}^{c \times 1}$. The spatio-temporal data set consisting of input-output pairs is denoted by $\mathcal{T}$, where $\mathcal{T} = \{ S^{(t)}, Y^{(t)} \}^T_{t=1} $, with t denoting the sample index and $T$ the total number of snapshots. $G$ and $E$ together provide a connectivity representation of the edge set $\mathcal{E}$ for each sample (further details in Appendix \ref{app4:mod1}).

The set of batches is partitioned sequentially into training and testing sets using a fixed $70/30$ split. $T_{train}$ and $T_{test}$ denote the number of snapshots used for the training and testing periods, respectively. Details regarding the partition strategy for the training and testing sets are provided in Subsection \ref{sub4:43conf}. The batches are then given to the spatio-temporal module, where the model learns the dependencies from the data structure to feed the next module, as shown in Fig. \ref{fig4:blockd}. Rather than having an entire data set for the learning process, this module produces a set of batches with spatial and temporal data.

\clearpage

\newpage

\subsection{Spatio-temporal processing}
This module is designed to learn and extract meaningful features from graph-structured data. We use a graph convolutional network (GCN) structure \cite{kipf17} as a graph-based model fundament, since it admits a distributed representation of spatial features and could be implemented with deterministic and stochastic perturbations. Disease spreading networks are modelled with two representations: (1) a PrGCN${_{rmsle}}$ version based on a GCN layer; and (2) a PrGLSTM${_{rmsle}}$ version composed of GCN and long short-term memory (LSTM) \cite{hoch97} layers. In both versions, the information is processed in a layer-by-layer fashion. The system captures the latent features of the disease cases using the batched temporal snapshots from the multivariate sequence encoding module.

\clearpage

\newpage
\begin{figure*}%[!h]
\centering

\includegraphics[width=0.95\textwidth]{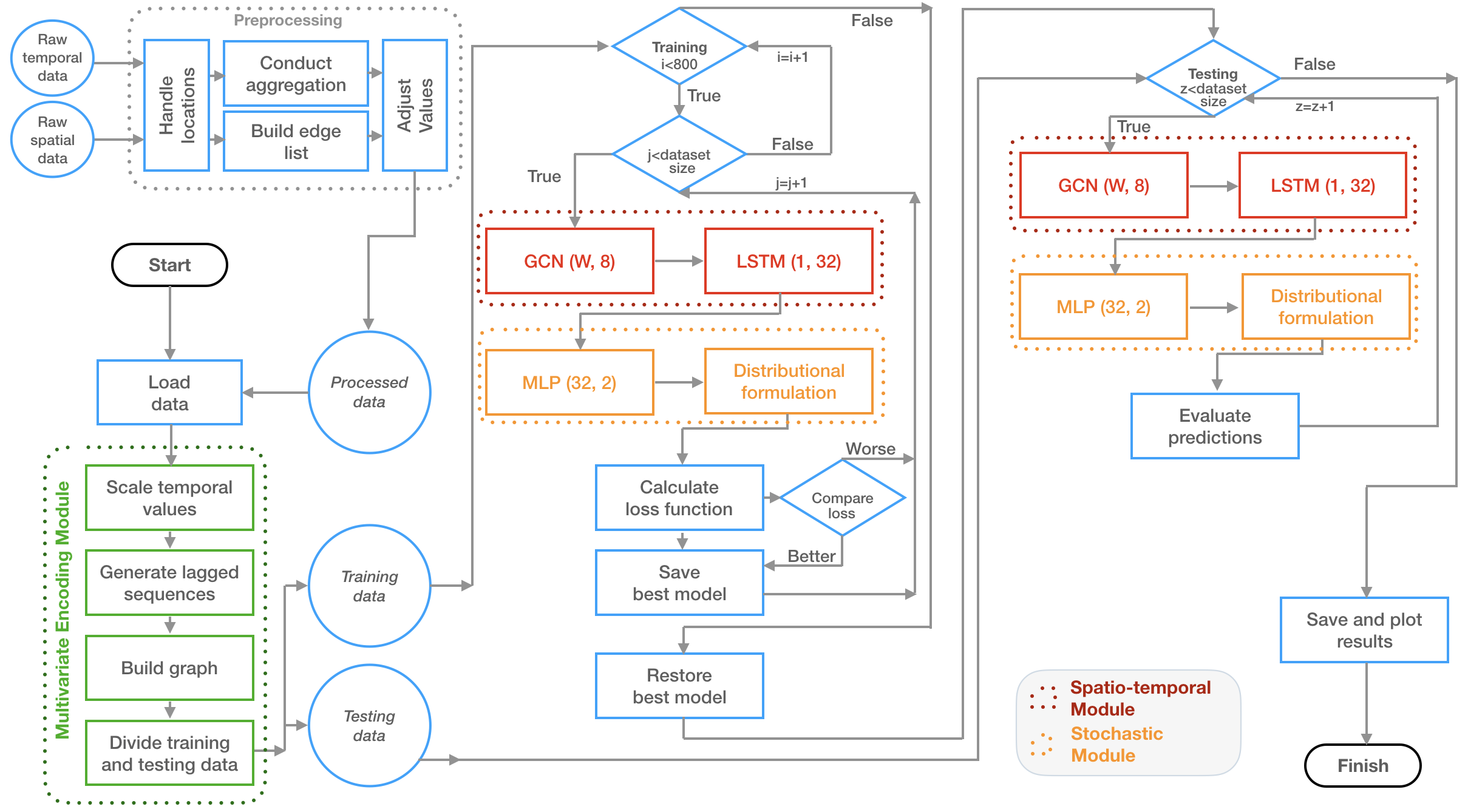}

\caption{Block diagram of the proposed architecture (PrGLSTM${_{rmsle}}$ version), showing the interaction between input data, preprocessing stages and the three modules for the training and testing phases.}
\label{fig4:blockd}
\end{figure*}

At each step $t$, a spatio-temporal snapshot $X^{(t)} \in \mathbb{R}^{G_{l} \times W}$ and the corresponding data are taken to map them into a high-dimensional representation. The new dense embeddings disentangle key factors of the observed system.

$\mathbf{PrGCN_{rmsle}}$ \textbf{version}. Given each spatio-temporal snapshop $X^{(t)} \in \mathbb{R}^{G_{l} \times W}$, the corresponding target $Y^{(t)} \in \mathbb{R}^{ 1 \times G_{l}  }$, the edge index tensor $G \in \mathbb{N}^{2 \times c}$ and edge attribute vector $E \in \mathbb{R}^{c \times 1}$, where $t= \{1, \dots,T_{train} \} \in \mathbb{N}$, a GCN layer generates $Q$-dimensional descriptors, yielding a tensor $\mathbf{X}^{\prime} \in \mathbb{R}^{G_{l} \times Q}$ at a given step $t$, by applying aggregation and readout methods. These operations support the exchange of information between neighbouring nodes in the graph structure. Consider an unweighted graph $\mathcal{G=(V, E)}$ that contains $G_{l}$ nodes, a vertex set $\mathcal{V} = \{ 1, ... , G_{l} \}$, and the edge set $G$ with $c$ undirected edges $( i,j ) = ( j,i) \in \mathcal{E} $. During the training process, the representations are learned through layer-wise propagation rules in a stage conditioned to spatio-temporal features of the data. The operations of the GCN are defined in Eq. (\ref{eqn4_GCN}), where $\mathbf{X}$ is the input feature matrix or the spatio-temporal snapshot $X^{(t)} \in \mathbb{R}^{G_{l} \times W}$. For each node $i$, the $Q$-dimensional temporal descriptor $\mathbf{x}^{\prime}_i \in \mathbb{R}^{1 \times Q}$ at a given step $t$ is calculated as a result of the symmetric normalisation in neighbourhood aggregation, illustrated in Eq. (\ref{eqn4_xedges}). The detailed description of these processes is provided in Appendix \ref{app4:PrGCN}.

\begin{equation}
    \label{eqn4_GCN}
        \mathbf{X}^{\prime} = \mathbf{\hat{D}}^{-1/2} \, \mathbf{\hat{A}} \, \mathbf{\hat{D}}^{-1/2} \, \mathbf{X} \, \Theta
\end{equation}

\begin{equation}\label{eqn4_xedges}
    \mathbf{x}^{\prime}_i = \sum_{j \in \mathcal{N}(i) \cup \{ i \}} \frac{e_{j,i}}{\sqrt{\hat{d}_j \hat{d}_i}} \mathbf{x}_j  {\theta}
\end{equation}

Setting $e_{j,i} = 1$ for adjacent counties ensures that the aggregation process treats all geographic neighbours as equally influential factors, and each node contribution is normalised relative to the local spatial density by the symmetric normalisation term. This process allows the model to learn different spatial correlations that are invariant to the local connectivity network, with a consistent influence scale, that are not clearly identified in the $W$-dimensional features. These steps ensure a better understanding of the contextual changes that facilitate the prediction process and prevent bias introduced by varying node degrees, where high-density geographic clusters might dominate the feature space. $\mathbf{X}^{\prime} \in \mathbb{R}^{G_{l} \times Q}$ is the resulting representation at a given step $t$.

$\mathbf{PrGLSTM_{rmsle}}$ \textbf{version}. This version accounts for a LSTM layer added sequentially after the GCN structure. Given each spatio-temporal snapshop $X^{(t)} \in \mathbb{R}^{G_{l} \times W}$, corresponding target $Y^{(t)} \in \mathbb{R}^{ 1 \times G_{l}  }$, edge index tensor $G \in \mathbb{N}^{2 \times c}$, edge attribute vector $E \in \mathbb{R}^{c \times 1}$, and output GCN tensor $\mathbf{X}^{\prime} \in \mathbb{R}^{G_{l} \times Q}$ at a given step $t$.

$t= \{1, \dots,T_{train} \}$ and $j= \{1, \dots,G_l \} \in \mathbb{N}$ denote the sample index of a spatio-temporal snapshot and the node or geographic location, respectively, a LSTM layer generates $8Q$-dimensional temporal descriptors, yielding a tensor $\mathbf{H}^{(t)} \in \mathbb{R}^{G_{l} \times Q \times 8Q}$ at a given step $t$. At each step $t$, the outputs $\mathbf{X}^{\prime} \in \mathbb{R}^{G_{l} \times Q}$ from the GCN layer are fed into the LSTM structure (Fig. \ref{fig4:blockd}). The LSTM processes each location independently, but in parallel across all $G_l$ locations. The operations of the LSTM cells at a given location $j$ and step $t$ are provided in Appendix \ref{app4:PrGLSTM}, where the outputs are conditioned to a context tensor and a previous hidden state. The output tensor $\mathbf{H}^{(t)}$ is reshaped into a tensor $\mathbf{H}_{G_l}^{(t)} \in \mathbb{R}^{G_{l} \times 8Q}$ to represent the last hidden state of the process. During model training, the GCN and LSTM layers sequentially process $T_{train}$ consecutive input sequences without randomly indexing in separate formats. After the GCN layer, the structured latent representation is a nonlinear space-state epidemic system and the LSTM approximates the learning of hidden state evolution with stable memory retention over multi-wave dynamics and non-stationary events, while filtering short-term noise.

\subsection{Stochastic method}
\label{sub4_stochfor}

This module induces minimal perturbations in the learning system to simulate internal variability during the training and testing processes. Given each tensor $\mathbf{Z}^{(t)} \in \mathbb{R}^{G_{l} \times \Gamma}$ at a step $t$, where $\mathbf{Z}^{(t)}$ represents the output of the spatio-temporal processing module, $t= \{1, \dots,T_{train} \}$ denotes the sample index of a spatio-temporal snapshot, $ \Gamma= Q $ and $ \Gamma= 8Q $ correspond to the $Q$- and $8Q$-dimensional descriptor output from PrGCN${_{rmsle}}$ and PrGLSTM${_{rmsle}}$ models, respectively, this module maps each embedding to a 1-dimensional descriptor, producing a tensor $\mathbf{Y}^{\prime} \in \mathbb{R}^{G_{l} \times 1}$ at a given step $t$ by using a fully connected layer and a random sampling formulation to induce stochasticity in the final stage of the prediction process.

Initially, a fully connected stochastic layer: a multi-layer perceptron (MLP) \cite{mlp91} processes the multiple input arguments $\mathbf{Z}^{(t)} \in \mathbb{R}^{G_{l} \times \Gamma}$ from the spatio-temporal module. After that, a representation of the latent features is decoded in a tensor $\mathbf{P}^{(t)} \in \mathbb{R}^ {G_{l} \times 2}$, where $t= \{1, \dots,T_{train} \}$ and $j= \{1, \dots,G_l \} \in \mathbb{N}$ denote the sample index of a spatio-temporal snapshot and the node or geographic location, respectively. The previous module structurally reduces variance, while the MLP provides expressive final mapping without reintroducing additional spatial and temporal structural priors that might risk over-smoothing the final representation.

At each location $j$, a mapping strategy is employed over the elements of the tensor $\mathbf{P}^{(t)}$, where $\mathbf{p}^{(t)}_{j,1}$ is equivalent to the form $ln \, \sigma^{(t)}_{j}$ to get $\sigma^{(t)}_{j} \in \mathbb{R}$, and $\mathbf{p}^{(t)}_{j,2} \in \mathbb{R}$ is represented by $\mu^{(t)}_{j} \in \mathbb{R}$. This strategy allows to generate normal distributions $\mathfrak{D}_{j}^{(t)}$ from the outputs, with $ \sigma^{(t)}_{j}$ and $\mu^{(t)}_{j}$ as the standard deviation and mean parameters. We process each $\mathfrak{D}_{j}^{(t)}$ via parameterised stochastic and deterministic operations that transform standard parameter-free distributions called a reparameterisation trick operation \cite{king13}. The model learns the distribution and pulls a random value from the fixed distribution. Let a final prediction $ \hat{y}_{j}^{(t)} \in \mathbb{R} $ at step $t$ and location $j$ be determined by $\hat{y}_{j}^{(t)} = \mu^{(t)}_{j}+\sigma^{(t)}_{j} \odot\epsilon_j$, with an auxiliary noise variable $\epsilon_j$. This randomness is external to the learnable components of the model and allows the model to represent a range of possibilities. This formulation ensures that the random element remains independent of the network weights, while enabling gradient-based optimisation when sampling from a probabilistic latent variable. The final predictions for all locations $G_l$ stack in a tensor $ \hat{Y}^{(t)} \in \mathbb{R}^{G_{l} \times 1} $ at a given step $t$.

\subsection{Loss function}

The proposed architecture is tailored to be trained in an end-to-end manner using a Root Mean Squared Logarithmic Error (RMSLE) as a loss function ($\mathcal{L}_{R\!M\!S\!L\!E}$). Given a target tensor $Y^{(t)} \in \mathbb{R}^{G_l}$ and a model prediction tensor $ \hat{Y}^{(t)} \in \mathbb{R}^{G_{l} \times 1} $, $\mathcal{L}_{R\!M\!S\!L\!E}$ calculates the error using the squared difference of logarithms, as described in Appendix \ref{app4:loss}.

$\mathcal{L}_{R\!M\!S\!L\!E}$ supports the model learning process by assigning larger penalties for the underestimation of disease cases than for overestimation. A global loss estimation guides the model to focus on the main global dynamical features. For example, $\mathcal{L}_{R\!M\!S\!L\!E}$ helped the model learn the dominant trends of COVID-19 spread across the US neighbourhood topology at a given time, since a variety of local intervention measures were handled during the COVID-19 outbreaks \cite{hallas21} and each location exhibited a disease transmission trend in a widely different fashion.

A total loss value per epoch $\mathcal{L}_{epoch} \in \mathbb{R}$ is also empirically computed to evaluate its convergence over multiple runs of the model training. This process enables to analyse optimisation stability and model parameters.

\subsection{Uncertainties}
\label{4.1.uncertain}

The proposed architecture accounts for the capture of epistemic uncertainties using an ensemble method to approximate empirical variance and maintain the structural integrity of the architecture and the same values of hyperparameters and data input. The proposed architecture is trained multiple times using different initialisations (seed values), yielding an ensemble of $R \in \mathbb{N}$ independent simulations (more details in Appendix \ref{app4:unc}). 

After training, the learnable parameters of the model are optimised, producing a fixed mapping $\mathfrak{M}_r$ with $r= \{1, \dots,R \} \in \mathbb{N}$. Each mapping is applied to the unseen data during the testing phase to obtain the predictions $ \hat{Y}_r^{(t)} \in \mathbb{R}^{G_{l} \times 1} $ at a given step $t= \{1, \dots,T_{test} \} \in \mathbb{N}$. After $R$ runs, a tensor $ \hat{Y}_E \in \mathbb{R}^{G_{l} \times T_{test} \times R} $ with $G_l, T_{test}, R \in \mathbb{N}$ represents the set of predictions for several individual mappings $[\mathfrak{M}_1, \mathfrak{M}_2,... , \mathfrak{M}_{R-1}, \mathfrak{M}_{R}]$. We assumed that the ensemble of predictions $ \hat{Y}_E \in \mathbb{R}^{G_{l} \times T_{test} \times R} $ with $j= \{1, \dots,G_l \} \in \mathbb{N}$, $t= \{1, \dots,T_{test} \} \in \mathbb{N}$, and $r= \{1, \dots,R \} \in \mathbb{N}$ follows a normal distribution along the dimension $R$. At each step $t$ and location $j$, a 95\% variability interval is estimated as a function of a sample mean $(\mu_E)$, a sample standard deviation $(\sigma_E)$, and number of observations $R$  from $\hat{Y}_E \in \mathbb{R}^{G_{l} \times T_{test} \times R}$. Finally, a more robust disease prediction is represented by $\mu_E$ with a 95\% variability interval at each step $t$ and location $j$, where high variability suggests that the final predictions are sensitive to a specific model initialisation.

\subsection{Experiments}
\label{sec:4exp}

\subsubsection{Datasets}

We conducted extensive experiments to examine how sensitive the architecture is to handle a variety of trends in disease spread. The experimental testing was conducted on public datasets for COVID-19 incidence in the US and chickenpox incidence in Hungary. Each data set represents a particularity of the complex nature of disease spread, as described in Appendix \ref{app4:data}. The 3-year analysis of COVID-19 spread involves highly complex spread patterns across $3,218$ counties in the US, while the 10-year analysis of Chickenpox spread accounts for a more consistent disease presence across $20$ counties in Hungary.

\textbf{COVID-19 data}\label{sub4:41covusa}. Disease incidence data for the US regions were obtained from the COVID-19 confirmed cases repository, compiled by Johns Hopkins University (JHU) \cite{jhu20}, at the county level and on a daily basis. We preprocessed a subset of these surveillance data from 2020 to 2022, which includes disease information for 3,218 counties, related to 50 states, 1 federal district, and 1 unincorporated territory. Therefore, the analysis considered data from 3,218 county-level locations in the US. The disease data cover a timeline from January 22, 2020, to October 4, 2022 (987 consecutive daily cases), with the same lengths of time series data for all counties. The corresponding topology data are collected from the United States Census Bureau source\footnote{US Census Bureau. County Adjacency File (2010). Accessed: 2026-05-18. https://www.census.gov/geographies/reference-files/2010/geo/county-adjacency.html}, as the county adjacency data.

In this study, daily disease data were smoothed using a 7-day aggregation process that resulted in a temporal data set $\, X_{raw,7} \,$ with 3,218 locations and 141 consecutive temporal steps. Each time step represents new weekly cases. The corresponding spatial data $\, G_{raw} \,$ were set with county-to-county edge connections equivalent to 22,154 geographic edges. Each location is represented by a node, which is connected with edges if the corresponding neighbour node shares a common border, and also with self-loops. More information is provided in Appendix \ref{apd4:second}.

\textbf{Chickenpox data}\label{sub4:41chick}. Disease incidence data and topology data for Hungary were collected from the UC Irvine Machine Learning Repository\footnote{Hungarian Chickenpox Cases (2021). UCI Machine Learning Repository. Accessed: 2026-05-18. https://doi.org/10.24432/C5103B}, at the county level and weekly basis. We preprocessed the total of this surveillance data from 2005 to 2014, which include disease weekly cases for 19 counties and the capital city Budapest in Hungary. The disease data cover a timeline from January 3, 2005, to December 29, 2014. $T_{s_r}$ is 522, which denotes a temporal sequence of 522 consecutive temporal steps. Each time step corresponds to a weekly total of confirmed chickenpox cases. This data set is represented by $X_{raw,7}$, and consists of static and non-overlapping observations, with the same lengths of time series data and no missing values for all counties. $G_l$ is 20, which reflects 20 locations in Hungary.

Each location represents an equivalent county entity in Hungary. The spatial structure $G_{raw}$ corresponds to the topology information with neighbouring edges and self-loops, equivalent to 102 geographic edges.

In this study, we analysed the temporal data set $X_{raw,7}$ with 20 locations and 522 consecutive time steps, and the corresponding spatial data $G_{raw}$ with 102 geographic edges. Each location is represented by a node, which is connected with edges if the corresponding neighbour node shares a common border, and also with self-loops.

\subsubsection{Single-scale analysis setting}

During our preliminary analysis, we experimentally evaluated multiple values for the temporal observation window $W \in \{2,4,6,8\}$. We observed that lower values ($W=2$) were insufficient for the model learning, while higher values ($W \geq 6$) led to increased computational overhead that did not improve the predictions. We found that $W = 4$ was the optimal value that provided the most consistent convergence of the predictions across the geographic locations. Furthermore, a temporal observation window of $W = 4$ represents four weeks of incidence data, providing the model with sufficient temporal context to identify emerging trends. With $W = 4$, the node feature matrix $X^{(t)} \in \mathbb{R}^{G_{l} \times W}$ becomes $X^{(t)} \in \mathbb{R}^{G_{l} \times 4}$.

\subsubsection{System configuration}
\label{sub4:43conf}
The proposed architecture employs operations to determine the best performing $\mathfrak{M}$ representation during the training process. The data are partitioned sequentially into a 70/30 ratio. The final evaluation is performed exclusively in the testing set to ensure that the predicted results reflect the out-of-sample performance of the model and prevent training-test overlap. This procedure supports the selection of a stable configuration to then verify whether it actually generalises during the testing phase. The training configuration includes 800 epochs, a input layer with 4 neurons, a hidden layer with 8 neurons for the PrGCN${_{rmsle}}$ and 8-64 neurons for the PrGLSTM${_{rmsle}}$, and a MLP layer with 2 neurons. Models are trained to minimise the loss $\mathcal{L}_{m}$ using back-propagation over time with a learning rate ($n$) of 0.01.

The model parameters are initialised randomly using the Adam optimiser \cite{adam14}. The uncertainty estimates are approximated by 100 runs of the model with different seed values. All experiments are conducted on a computer with NVIDIA RTX3070 GPU (8 GB memory), Ubuntu 20.04, CPU Intel i7 11th Gen 2.30GHz x 16, RAM memory 16 GB, with PyTorch Geometric \cite{pytorch19}, PyTorch Geometric Temporal \cite{pgt21}, PyTorch 1.10.0, PyTorch Geometric 2.0.4, PyTorch Geometric Temporal 0.52.0, and Python 3.8.10 libraries.

\subsubsection{Model complexity}

The computational demands of the proposed architecture are influenced by the operations performed in each stage of the model, such as processes in the modules, implementation details (e.g., batch size, number of epochs), data size (e.g., number of nodes, length of time series) and hardware resources (e.g., GPU). Therefore, the design of the proposed model focuses on keeping the internal trainable parameters limited to small values. The overall time complexity is empirically estimated by combining the time required to execute each component for the proposed and baseline models; more details are provided in Section \ref{sub4:5time}.

\subsubsection{Evaluation metrics}
We evaluate the performance of a model by estimating the mean squared error (MSE), mean absolute error (MAE), Huber error, root mean squared error (RMSE) and root mean squared logarithmic error (RMSLE) metrics, as listed in Table \ref{table4:errordef} (Appendix \ref{apd4:third}).

\subsubsection{Baselines}

We compared the proposed models with four spatio-temporal recurrent graph-based models as a benchmark study (Appendix \ref{app4:base}). The benchmark models are EvolveGCNH \cite{par20}, GCLSTM \cite{chen2022gc}, GConvLSTM \cite{youngjoo18}, and MPNNLSTM \cite{pan21}. Their architectures consist of different configurations of GCN and RNN layers, and have been developed in the PyTorch Geometric Temporal library. All benchmark models are trained under identical conditions and the configurations of their implementation settings are similar to the proposed architecture (Section \ref{sub4:43conf}) in terms of epochs and learning rate. Their hidden and output layer configurations depend on each architecture setting.

Similarly, ablation variants are trained under identical conditions to evaluate the effect of the architecture components during the single-scale analysis (Appendix \ref{app4:base}). First, we consider two baseline variants (PrLSTM${_{rmsle}}$ and PrLSTM${_{mse}}$) that replace the spatio-temporal module with a LSTM structure to assess whether the performance improves due to the proposed spatio-temporal module or can be explained by a standard temporal sequence structure alone. Second, we replace the RMSLE loss function with an MSE loss (PrGCN${_{mse}}$ and PrGLSTM${_{mse}}$), to evaluate whether the loss function contributes to better performance or adds complexity without clear benefit. Finally, we remove the random features of the stochastic module and the loss function, and retain minimal configurations which serve as baselines to analyse the impact of the architecture components compared to conventional models.

\section[Results]{Results}
\label{sec:4find}

This section presents the main finding of the COVID-19 and chickenpox studies. The analysis focuses on the predictive performance and temporal behaviour of the models across different locations. All figures show the 95\% variability intervals as shaded regions along the prediction trends, and the right-hand panels, highlighted in green, provide a magnified view of the prediction phase to facilitate detailed comparison.

\subsection{COVID-19 incidence predictions}
\label{sub:51cov}

\subsubsection{Overall predictability}
\label{sub:511over}
We report the predictions of new weekly cases. The data include counties with the highest weekly incidence: around \textit{300,000} confirmed cases in 2022; and also counties with few or no cases. The time-series data in the prediction phase cover two types of outbreak wave: a winter wave with larger peaks and a longer spring-summer wave with lower peaks. Figure \ref{fig4:errorstep_usa} shows the global weekly performance of PrGCN$_{rmsle}$ and benchmark models. In general, RMSLEs illustrate detailed comparative performance of the models for the spring-summer wave timing, and MSEs have larger values that help to visualise the enhanced performance of PrGCN$_{rmsle}$ for the winter wave timing.

Tables \ref{tab:benchmark_rmsle} and \ref{tab:benchmark_mse} (Appendix \ref{app4:glob}) show the global and averaged evaluation metrics for PrGCN$_{rmsle}$ and the baseline models. PrGCN$_{rmsle}$ exhibits a comparable global behaviour with lower variability intervals, resulting in lower error values during the prediction of winter peaks, and higher errors during the corresponding winter decreasing phase. Comparative analysis of the results reveals that PrGCN$_{rmsle}$ can approximate the overall temporal evolution of COVID-19 in most of the US counties, suggesting that the model might capture the dominant underlying disease dynamics.

Figures \ref{fig4:trend_best_us}, \ref{fig4:USsingle} and \ref{fig4:bakeraround_1} illustrate the PrGCN$_{rmsle}$ ability to represent the dominant temporal latent features of disease dynamics. However, predictions exhibit a consistent 1-step delay relative to observations. This temporal shift suggests that the model has reduced sensitivity to abrupt variations despite its ability to represent the wave trend direction. The ability of PrGCN$_{rmsle}$ to approximate the temporal alignment of the evolution of disease incidence may provide useful information to monitor disease trends that have similar particularities to COVID-19. However, the interpretation of these results in a public health policy context is limited. Its application to policy or decision-making strategies would require epidemiological expertise and methodological improvements in model performance to better respond to rapid changes in disease incidence trends.

\subsubsection{Winter predictability}
\label{sub:512win}
The substantially larger size of the 2022 winter wave is a unique feature of the spread of COVID-19 in the US regions, which lasted more than two months in most counties and exhibited periods of rapid rise and decline, irregularly timed peaks, and pre-outbreak intervals with distinct temporal patterns across locations. We focus on the prediction of these waves peaks because they reflect the larger proportion of people infected.

\begin{figure}%[!h]
\centering

\begin{minipage}{0.45\columnwidth}
\raggedleft

\includegraphics[width=0.97\textwidth]{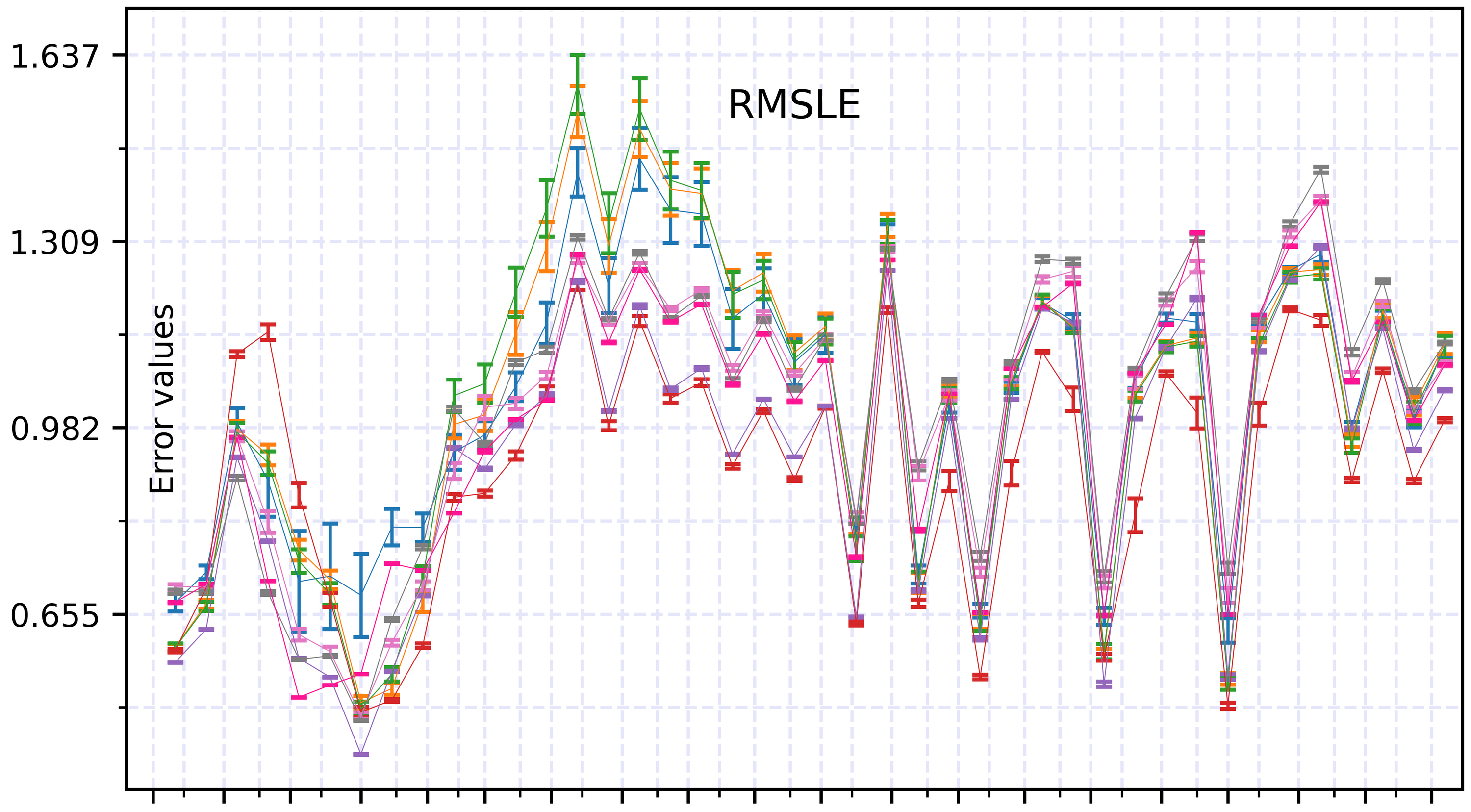}\par
\includegraphics[width=0.95\textwidth]{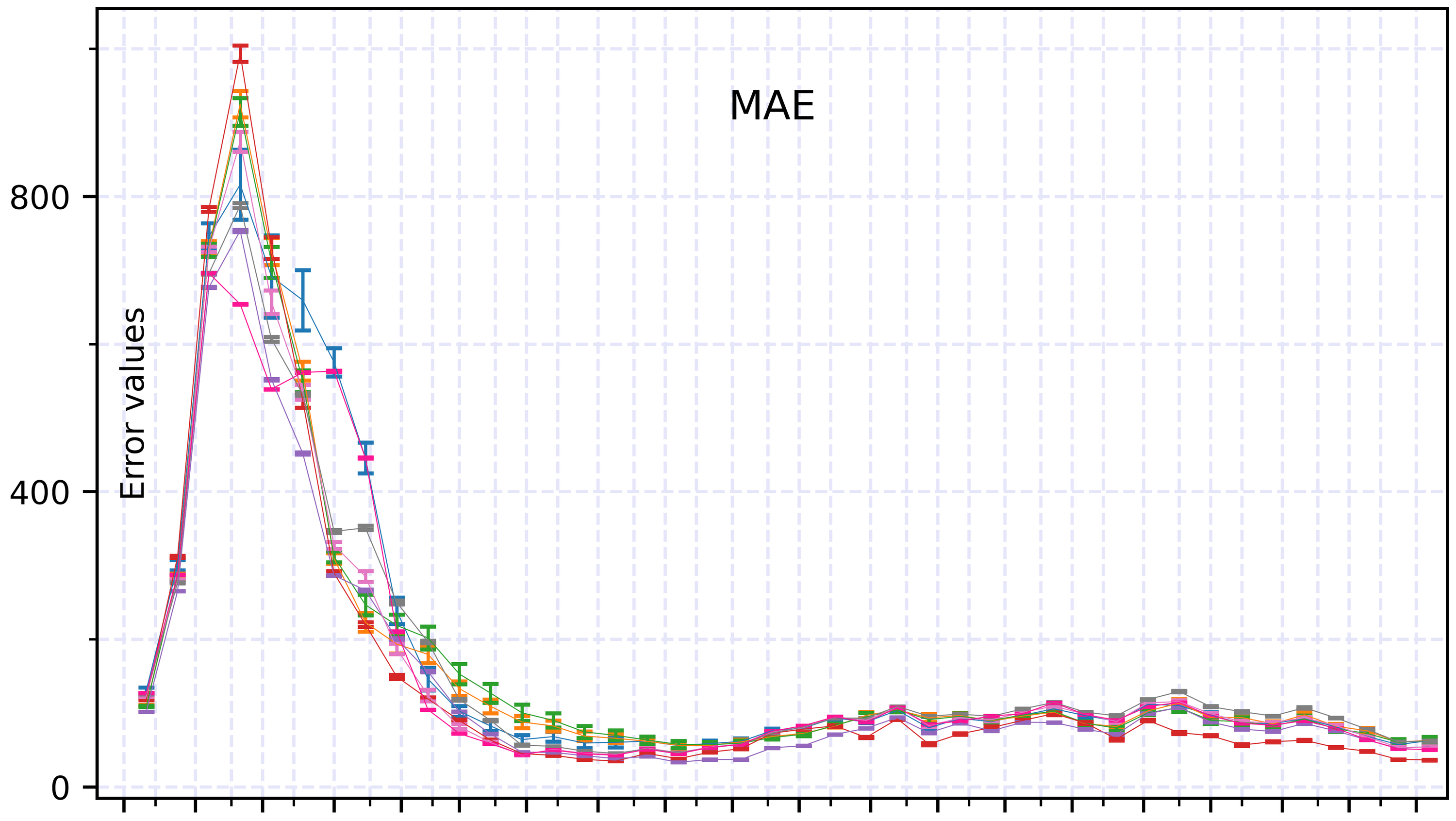}\par
\includegraphics[width=0.95\textwidth]{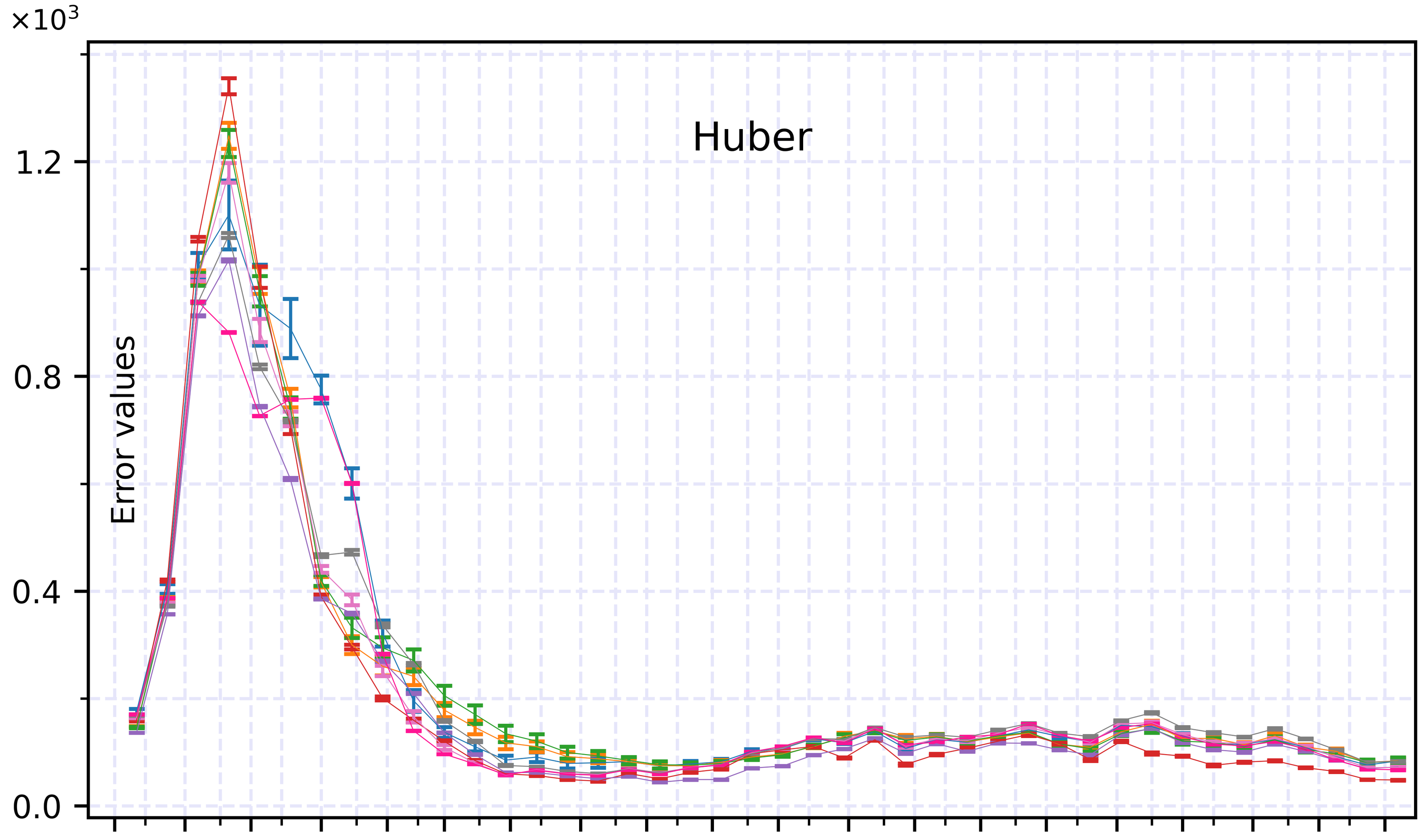}\par
\includegraphics[width=0.95\textwidth]{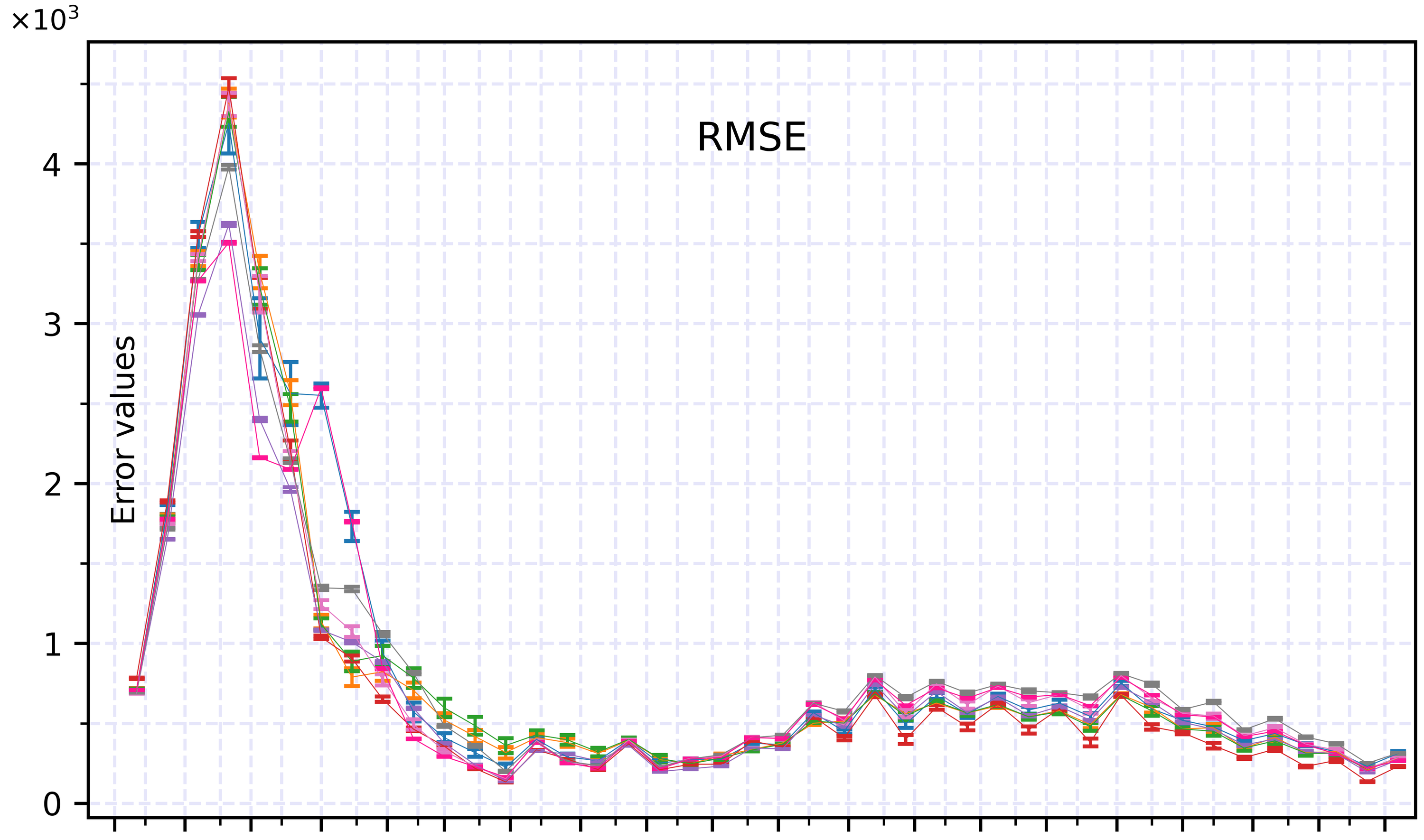}\par
\includegraphics[width=0.95\textwidth]{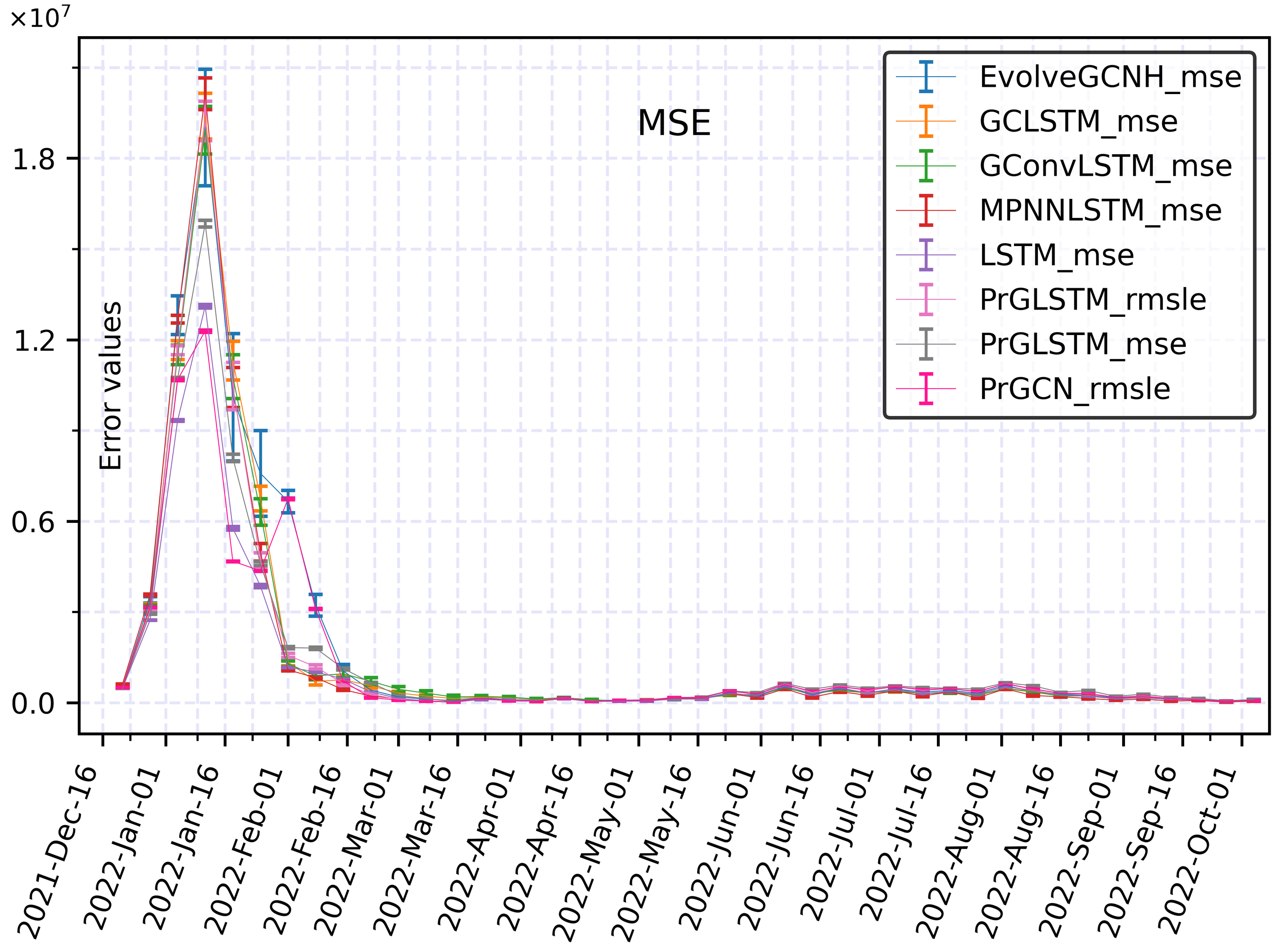}

\end{minipage}

\caption{
Weekly evolution of global performance metrics across the proposed and benchmark models for COVID-19 predictions in the US (3,218 counties). Dates are in year-month-day format.
}
\label{fig4:errorstep_usa}
\end{figure}

\begin{figure*}%[!h]
\centering

\begin{minipage}{1\textwidth}
\raggedleft

\includegraphics[width=1\linewidth]{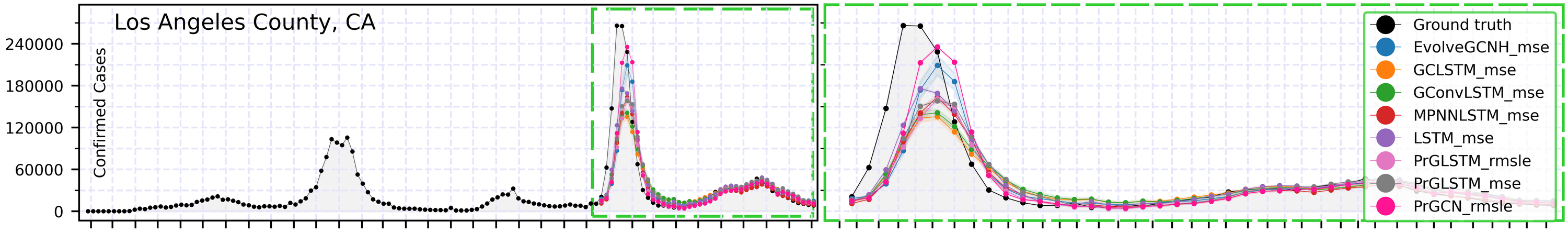}
\includegraphics[width=1\linewidth]{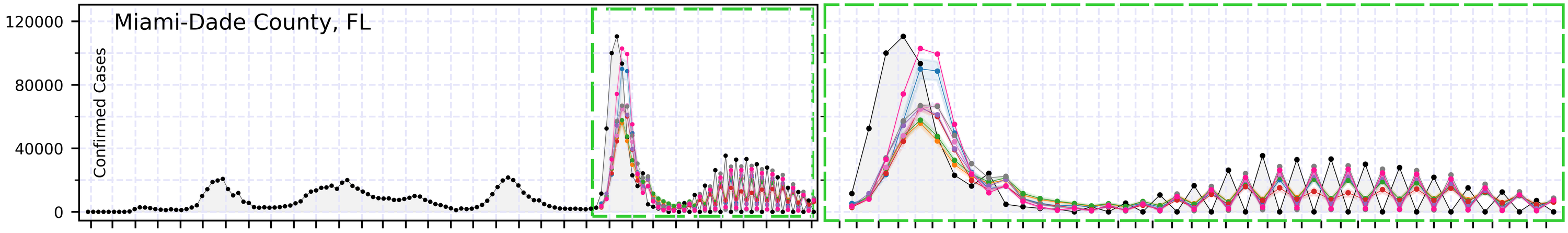}
\includegraphics[width=1\linewidth]{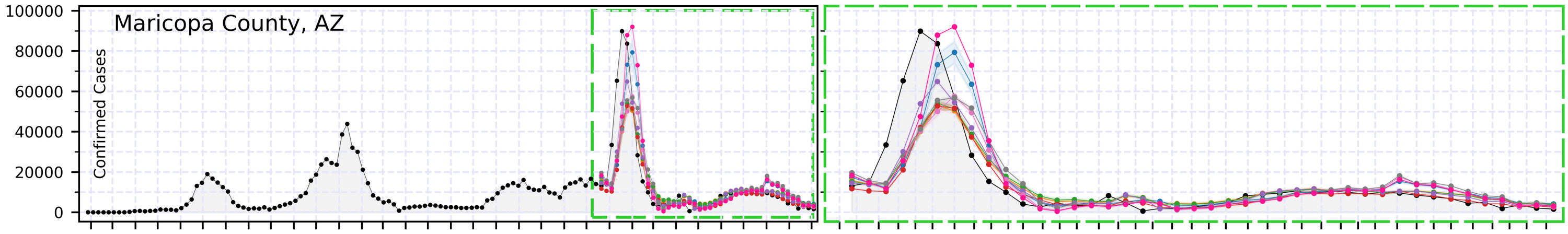}
\includegraphics[width=1\linewidth]
{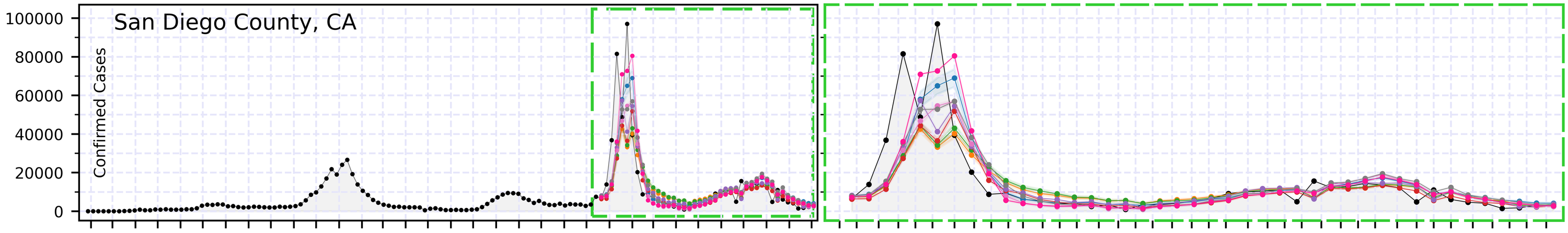}
\includegraphics[width=0.994\linewidth]
{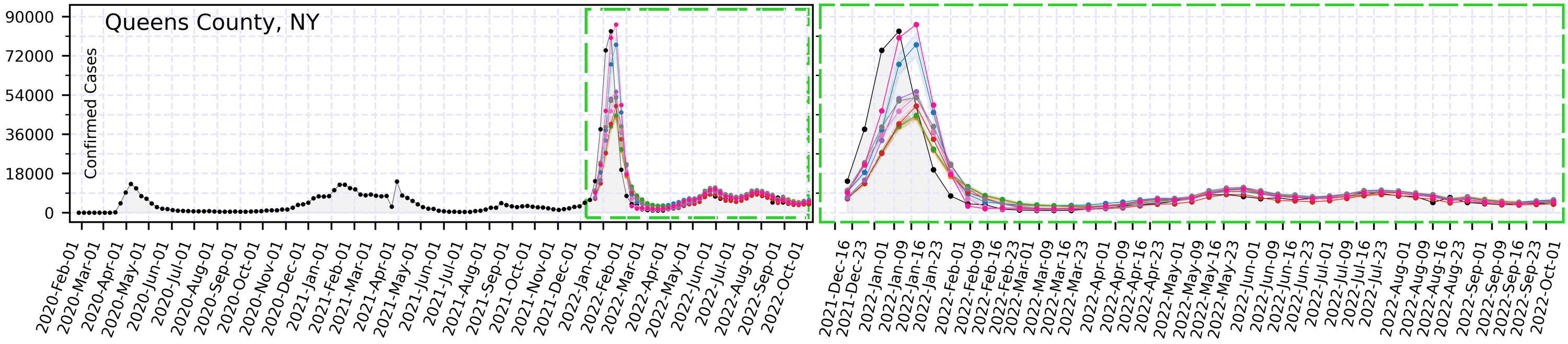}

\end{minipage}
\caption{Observed time-series data and model predictions for COVID-19 in US counties with the largest weekly incidence; right panel highlights a zoomed section of predictions with 95\% variability intervals.}
\label{fig4:trend_best_us}
\end{figure*}

\textbf{$\mathbf{PrGCN_{rmsle}}$}. Figure \ref{fig4:trend_best_us} shows that PrGCN$_{rmsle}$ is particularly good at following the shape of the winter peak, while maintaining a consistent 1-step delay pattern. This feature leads to an increased error estimation during the rising and decreasing phases of the winter wave, while showing lower average weekly error values during peak timing (Fig. \ref{fig4:errorstep_usa}). During the training process, the model encounters incidence waves with slopes that change steadily and slowly during the rising and decreasing phases, as a result, the prediction behaviour tends to follow such patterns during the testing phase, resulting in the delayed prediction timing of 1-step (equivalent to one week).

In contrast, few counties exhibit waves with slope values that steadily change during the testing phase, so PrGCN$_{rmsle}$ predictions change accordingly, as observed for Baker county in Figure \ref{fig4:bakeraround_1}. The more accurate estimation is linked to an improved model performance during the winter wave timing for such settings. In addition, we also observe that PrGCN$_{rmsle}$ overestimates predictions in locations with few cases or no cases at all, linked to a slight decrease in global model performance.

\begin{figure}%[!h]
\centering

\begin{minipage}{1\textwidth}
\raggedleft

\includegraphics[width=1\linewidth]{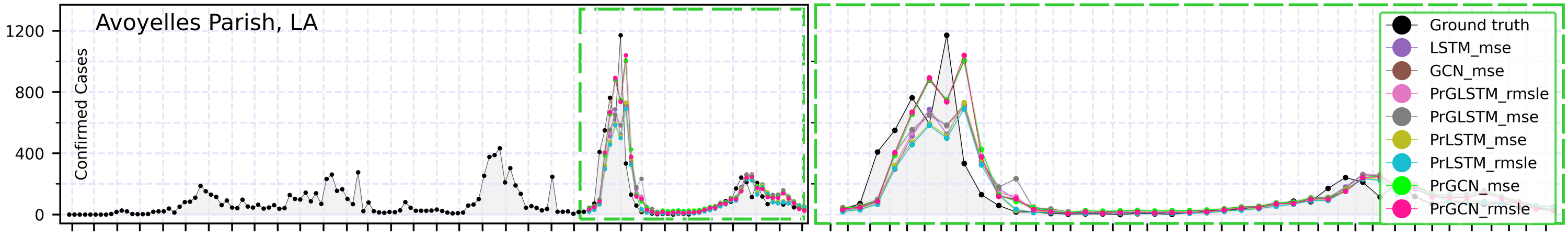}
\includegraphics[width=1\linewidth]{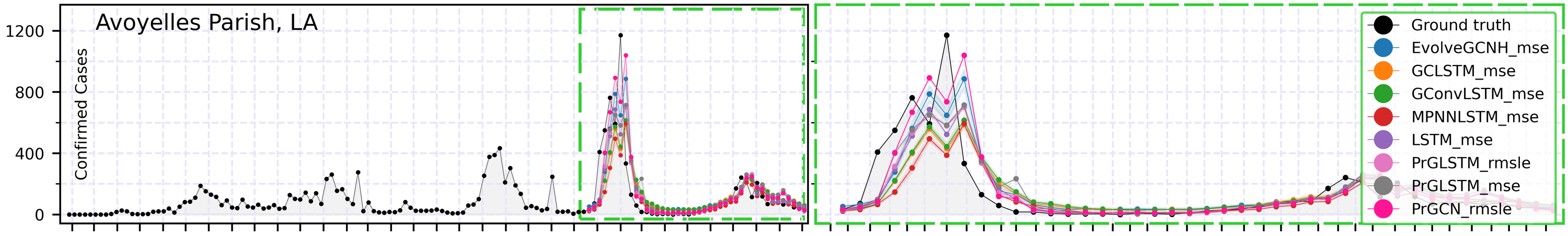}
\includegraphics[width=1\linewidth]{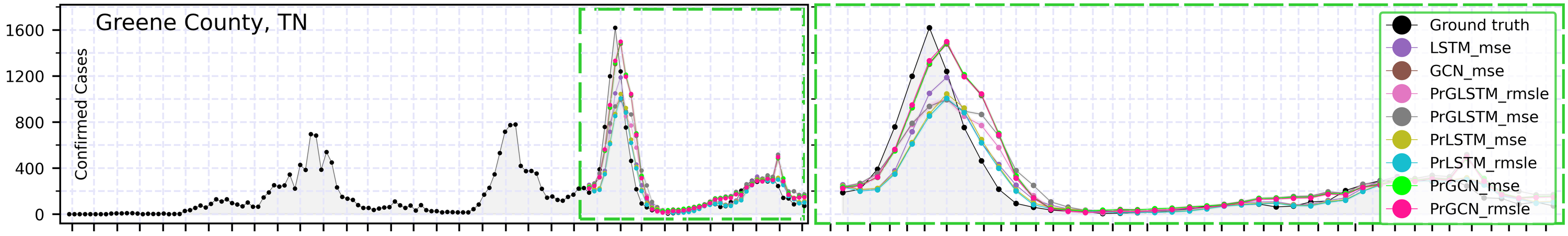}
\includegraphics[width=1\linewidth]{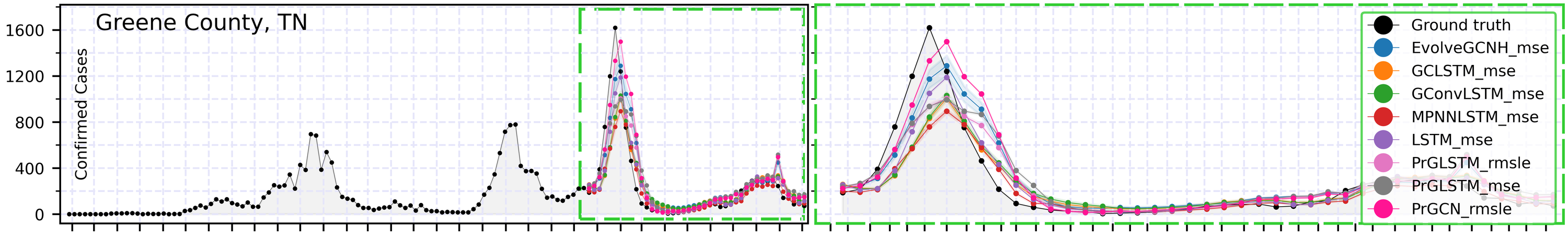}
\includegraphics[width=0.989\linewidth]{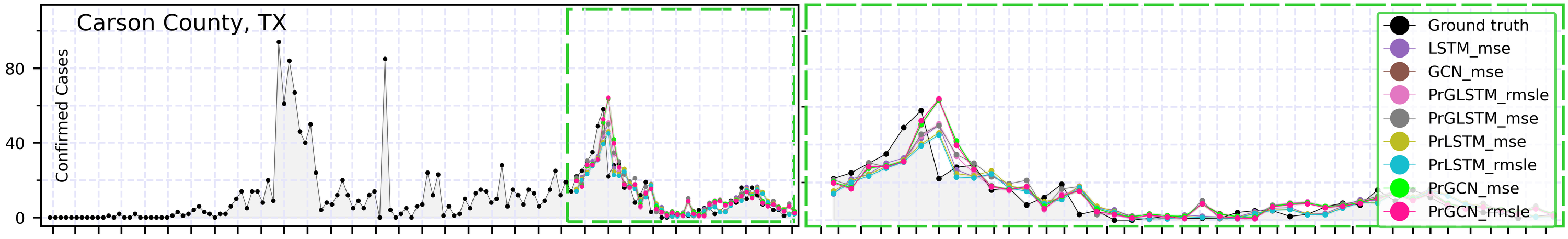}
\includegraphics[width=0.989\linewidth]{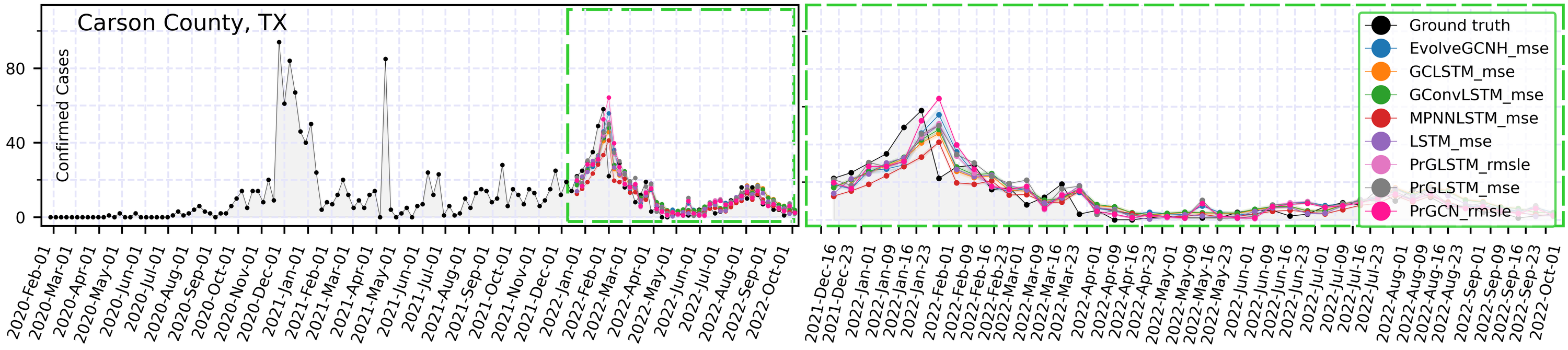}

\end{minipage}

\caption{
Observed time-series data and model predictions for COVID-19 in US counties using the proposed and baseline models during a heterogeneous training period.
}

\label{fig4:USsingle}
\end{figure}

\begin{figure}%[!h]
\centering

\begin{minipage}{1\textwidth}
\raggedleft

\includegraphics[width=0.989\linewidth]{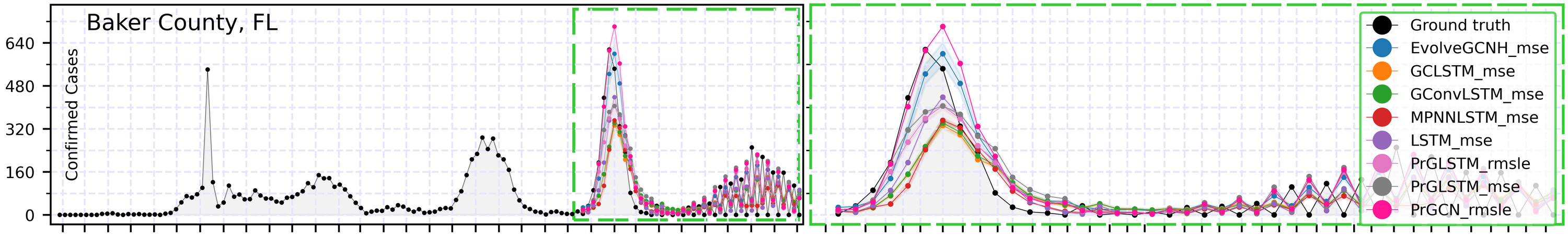}
\includegraphics[width=1.001\linewidth]{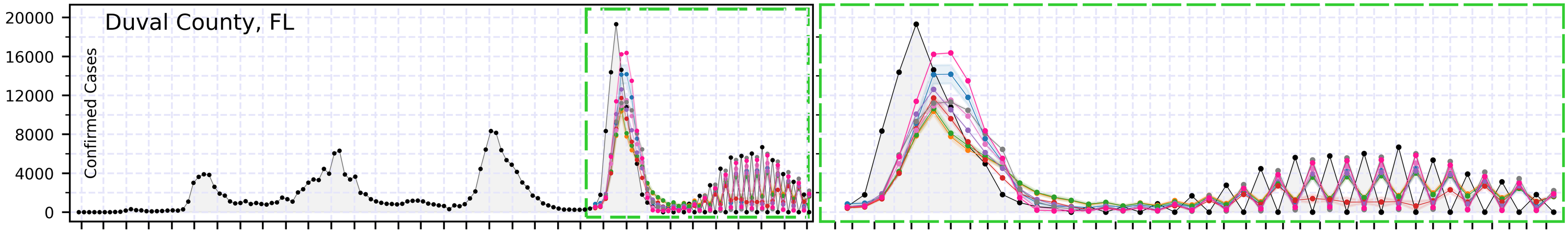}
\includegraphics[width=0.999\linewidth]{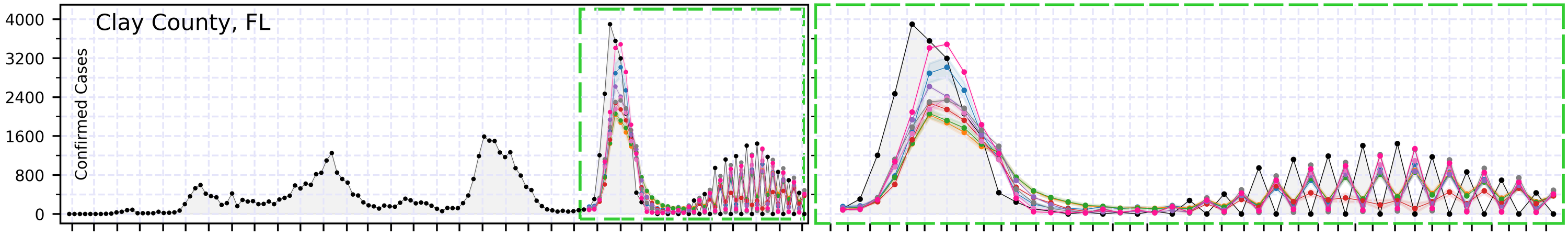}
\includegraphics[width=0.999\linewidth]{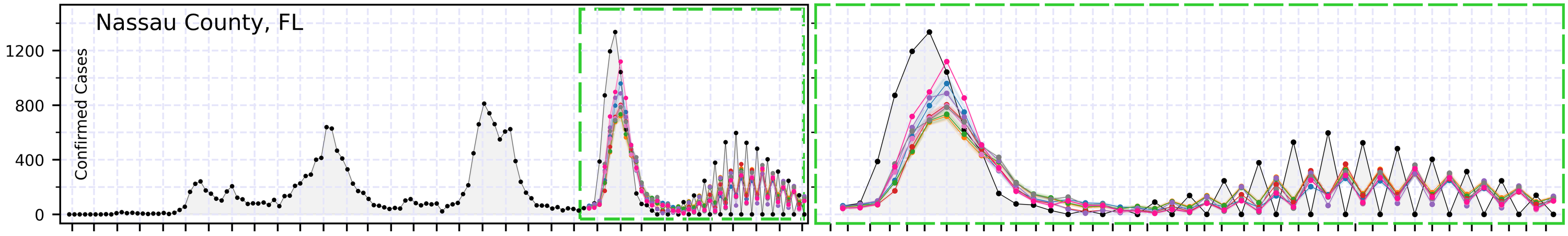}
\includegraphics[width=0.989\linewidth]{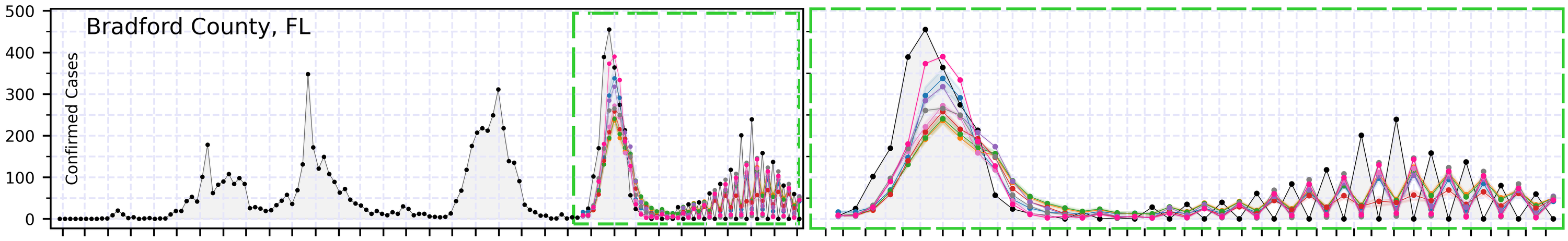}
\includegraphics[width=0.999\linewidth]{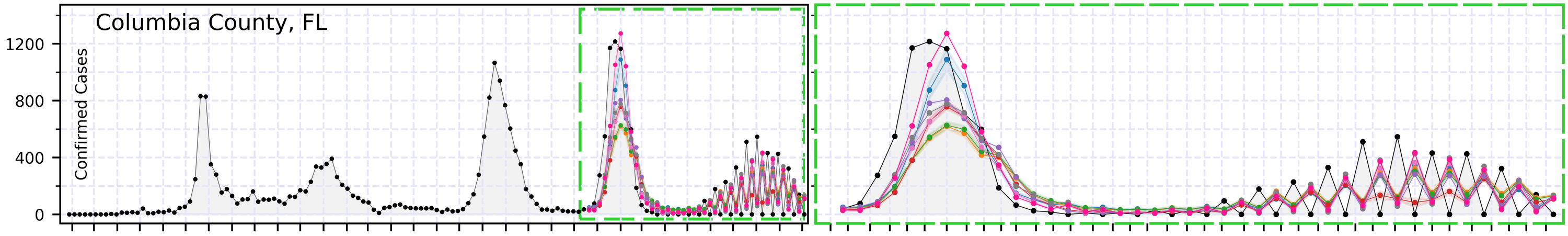}
\includegraphics[width=0.988\linewidth]{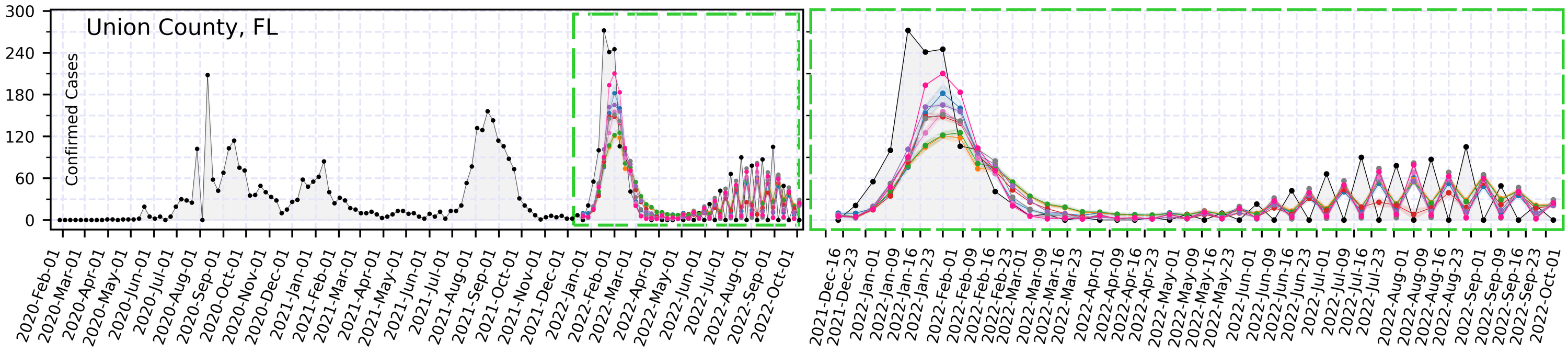}

\end{minipage}

\caption{
Observed time-series data and model predictions for COVID-19 in Baker and neighbouring counties in Florida (FL), US.
}

\label{fig4:bakeraround_1}
\end{figure}

$\mathbf{PrGLSTM_{rmsle}}$. Figures \ref{fig4:trend_best_us} and \ref{fig4:USsingle} illustrate that the proposed PrGLSTM$_{rmsle}$ misses the winter peaks, highlighted by increased weekly errors (Figs. \ref{fig4:errorstep_usa} and \ref{fig4:errorablat_us} in Appendix \ref{app4:pred}). As PrGLSTM$_{rmsle}$ and PrGCN$_{rmsle}$ account for the aggregation of the local neighbourhood in the learning process, the lower peak predictions of PrGLSTM$_{rmsle}$ during the winter season might be induced by the LSTM layer and its stronger temporal learning patterns, linked to the learning of smaller waves in the training data. This increases its errors (+60\% MSE of PrGCN$_{rmsle}$) during peak time.

PrGLSTM$_{rmsle}$ and PrGCN$_{rmsle}$ also maintain a similar performance during the rising phase of the winter wave. However, PrGLSTM$_{rmsle}$ predictions relatively approach the observed incidence during the decreasing phase (Fig. \ref{fig4:trend_best_us}).

\clearpage

\newpage

\clearpage

\newpage

This pattern is associated with smaller values of the peak predictions and delayed alignment, which yields lower errors during the decreasing phase of the winter wave. In contrast, PrGCN$_{rmsle}$ predictions are better aligned with real wave shapes while showing a delayed pattern, leading to larger errors (Fig. \ref{fig4:errorstep_usa}). PrGLSTM$_{rmsle}$ exhibits an MSE magnitude approximately 25\% that of PrGCN$_{rmsle}$ during the decreasing timing around February 1, 2022. This indicates that lower global weekly errors are not necessarily associated with better prediction capabilities for the winter peak values.

\textbf{Ablation models}. PrGCN$_{rmsle}$ performs better than its ablations in following the trend shape of winter peaks, as observed by lower weekly MSE values (Fig. \ref{fig4:errorablat_us} in Appendix \ref{app4:pred}). However, the proposed and ablation models show a consistent 1-step delay in their predictions (Figs. \ref{fig4:trend_FLc} and \ref{fig4:trend_best_us_ablation} in Appendix \ref{app4:pred}). The ablation models for PrGLSTM$_{rmsle}$ (with LSTM structures) maintain a constant prediction performance for counties with no cases at all, maintaining lower errors than the proposed models for such locations.

The ablation models for PrGCN$_{rmsle}$ show similar performances during the decreasing timing of the winter wave, resulting in higher errors than the LSTM-based ablations. In addition, replacing the PrGCN$_{rmsle}$ loss with MSE involves a subtle decrease in performance (+6\% MSE of PrGCN$_{rmsle}$). PrGCN$_{rmsle}$ estimates the peaks slightly better than GCN-based ablations, highlighting that the proposed architecture is capable of handling stochastic complexity without affecting performance.

\textbf{Benchmark models}. The proposed and benchmark models show predictions with a 1-step delayed pattern (Figs. \ref{fig4:trend_best_us}, \ref{fig4:USsingle}, \ref{fig4:bakeraround_1} and \ref{fig4:bakeraround_2} in Appendix \ref{app4:pred}). All models perform similarly during the rising timing (Figure \ref{fig4:errorstep_usa}), while PrGCN$_{rmsle}$ has a visible improvement in peak prediction.

\clearpage

\newpage

PrGCN$_{rmsle}$ and EvolveGCNH have the highest global weekly error values during decreasing timing, as both model predictions exhibit a delay but still align with the shape of the peak. PrGCN$_{rmsle}$ also shows better accuracy and lower variance during peak timing, highlighting its better stability for predictions.

\subsubsection{Spring-summer predictability}
\label{sub:513spri}
An interesting feature of the 2022 spring-summer wave is a wide distribution with a small peak value similar to those of the features in the training data. This surge can be attributed to the reinstatement of face-mask requirements around mid-2022 that aimed to minimise the disease spread, as new subvariants of Omicron were gaining dominance during these seasons. Some counties also display highly fluctuating behaviour (Fig. \ref{fig4:trend_FLc} in Appendix \ref{app4:pred}), correlated with the variable reporting discussed in Section \ref{apd4:first}. Since the training data do not account for waves with such fluctuating spikes, this behaviour has a significant impact on the prediction capabilities of the models.

\clearpage

\newpage

\clearpage

\newpage

During training, model mechanisms are able to capture features associated with rising patterns at low- and high-amplitude scales, which enables the model to distinguish upward trends regardless of the wave size. This learned pattern is also applied during the fluctuating period, where each spike reflects a rising point of a smaller wave, so the model predicts the relative change with a 1-step delay. As a result, the global weekly errors of the models are largely influenced by the predictions of such waves.

\textbf{$\mathbf{PrGCN_{rmsle}}$}. Figure \ref{fig4:trend_best_us} depicts the PrGCN$_{rmsle}$ predictions, characterised by estimation and even an overestimation of the peaks of spring-summer waves. We attribute this behaviour to the robust learning transference of the model. PrGCN$_{rmsle}$ predictions also oscillate with a slightly delayed onset for waves with fluctuating spikes of 1-step delay (Fig. \ref{fig4:trend_best_us}), which leads to inevitable error measures (Fig. \ref{fig4:errorstep_usa}). Despite this, PrGCN$_{rmsle}$ estimates approximate the global trend of the observed wave.

\textbf{$\mathbf{PrGLSTM_{rmsle}}$}. This model follows a similar behaviour to PrGCN$_{rmsle}$ in capturing waves with smaller peaks and no fluctuating patterns, as shown by comparable weekly RMSLEs (Fig. \ref{fig4:errorstep_usa}). These indicate that the sequence learning of the PrGCN$_{rmsle}$ offers similar capabilities to the LSTM layer in PrGLSTM$_{rmsle}$ for such waves (Fig. \ref{fig4:trend_best_us}).

\textbf{Ablation models}. PrGCN$_{rmsle}$ also shows a performance comparable to its ablations (Fig. \ref{fig4:errorablat_us} in Appendix \ref{app4:pred}). In case of waves with fluctuating patterns, predictions of GCN-based ablations tend to better align with the trending behaviour of the wave. However, LSTM-based ablations predict waves with slightly smoother features, resulting in missing global trending peaks.

\textbf{Benchmark models}. The predictions of all models also show a delayed pattern of 1-step (Fig. \ref{fig4:errorstep_usa}). In particular, LSTM and MPNNLSTM smooth the fluctuating behaviour underestimating the peak of the global trend, which tends to reduce global errors during the spring-summer season.

\subsection{Incidence predictions of chickenpox spread}
\label{sub:52cov}

\subsubsection{Overall predictability}
\label{sub:521overchick}

We provide predictions of new weekly cases. This period covers three seasonal waves during the 3-year testing period and few or no cases around September. The data account for counties with the highest weekly incidence (500+ cases), with a total number of incidence cases greater than zero. Figure \ref{fig4:errorstep_hun} compares the global weekly performance of PrGCN$_{rmsle}$ with its benchmark models, where all models exhibit comparable behaviour in terms of error metrics. Note that RMSLE values highlight the under- and overestimation of predictions by rescaling error differences for better visualisation.

Tables \ref{tab:benchmark_rmsle} and \ref{tab:benchmark_mse} in the Appendix report the global and averaged values of the evaluation metrics for PrGCN$_{rmsle}$ and baseline models. The underestimation and overestimation of the wave peaks by PrGCN$_{rmsle}$ result in varying errors, which leads PrGCN$_{rmsle}$ to have a comparable global performance. We also observed narrow variability intervals for PrGCN$_{rmsle}$ in terms of global performance and its weekly evolution, highlighting the stability of the model.

We found that PrGCN$_{rmsle}$ can predict the overall seasonal progression of chickenpox incidence in many Hungarian counties, as illustrated in Figs. \ref{fig4:trend_best_hun} and \ref{fig4:tolnaaround_bench} (Appendix \ref{app4:pred}). However, the model shows reduced sensitivity to some peak events, resulting in an underestimation or overestimation of incidence peaks (Fig. \ref{fig4:tolnaaround_bench}). In addition, the model shows reduced sensitivity to intermittent high-frequency and low-amplitude spikes present in the waves. Although this additional variability is not clearly resolved by the graph- and LSTM-based models, the LSTM-based models demonstrate improved alignment with the general trend of chickenpox incidence.

These results highlight that PrGCN$_{rmsle}$ performance is influenced by the specific characteristics of the disease and the spatial-scale of the network (20 nodes for Hungary). In this context, the proposed model might be more appropriate for a trend-level examination of disease incidence to understand broad disease patterns across the spatial network. However, this analysis is primarily descriptive and further methodological improvements are needed to translate it into actionable public health policy guidance.

\subsubsection{Time-series predictability}
\label{sub:522timchick}

The multiple seasonal outbreaks of chickenpox include fluctuating patterns during the pre-outbreak and peak timing of the waves. Each wave lasts an 11-month period and is shaped with a relatively wider peak, fluctuating spikes, and varying peak distributions. This analysis focuses on evaluating the model capabilities in capturing the disease spread behaviour where the training data have a large set of sequential weekly cases during a 7-year period, and the testing data include a 3-year period. The testing data have fewer waves with slightly higher peak magnitudes than the training data, and the graph structures of the model are shaped by the neighbourhood relationships between 20 counties in Hungary, which is considerably smaller than the one in the COVID-19 study (Section \ref{sub:51cov}).

\clearpage

\newpage
\begin{figure}[!h]
\centering

\begin{minipage}{0.45\columnwidth}
\raggedleft
    
\includegraphics[width=0.972\textwidth]{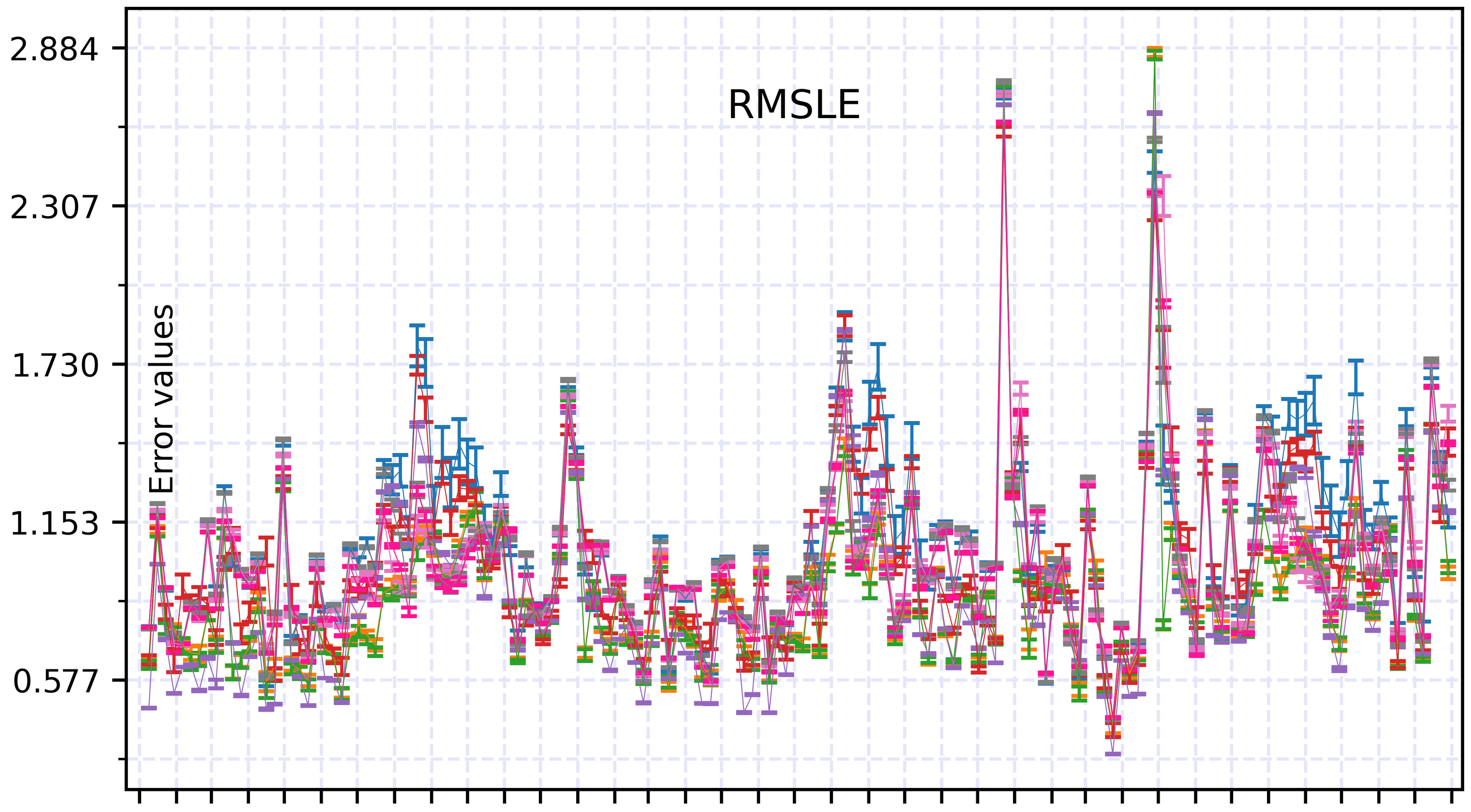}\par
\includegraphics[width=0.95\textwidth]{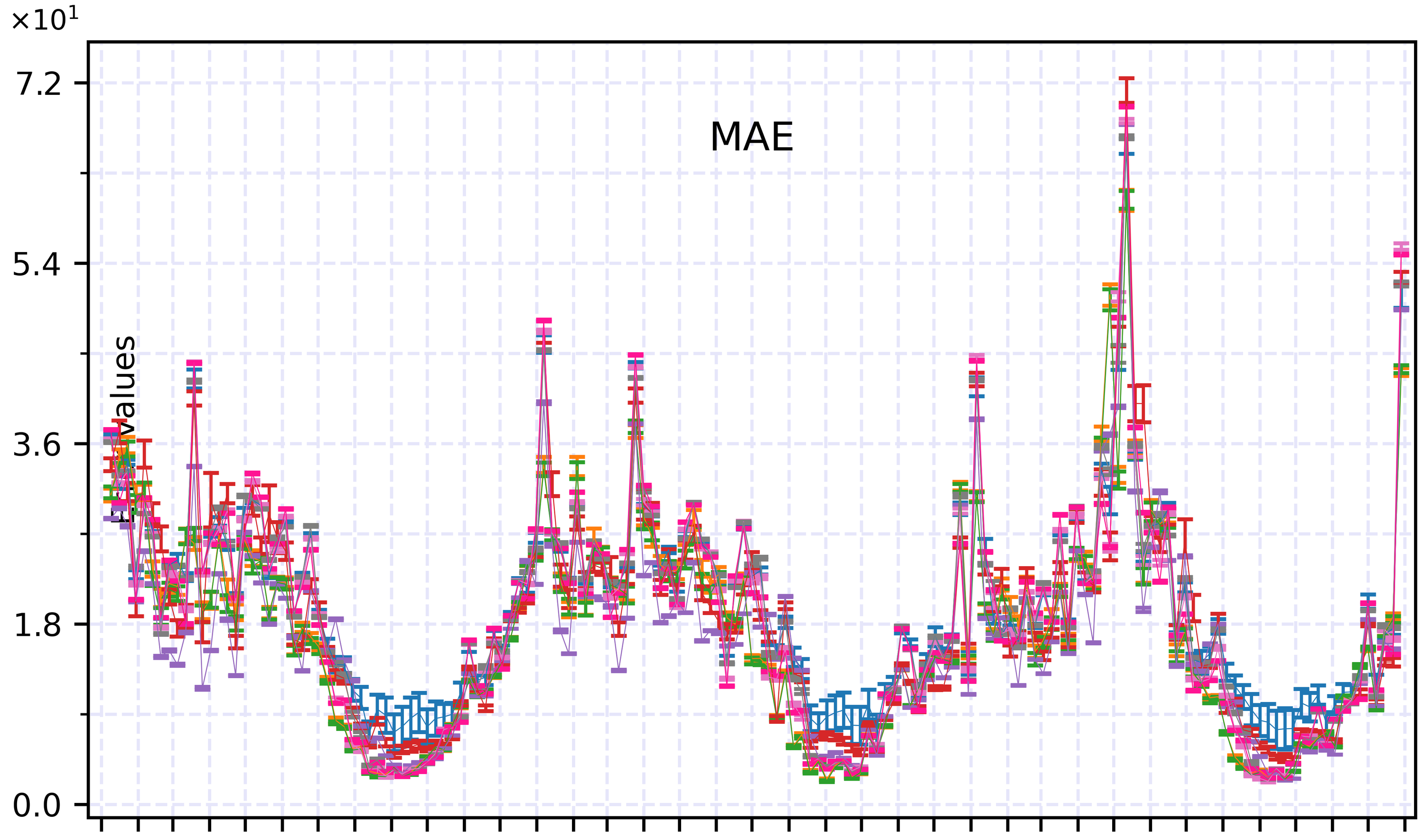}\par
\includegraphics[width=0.95\textwidth]{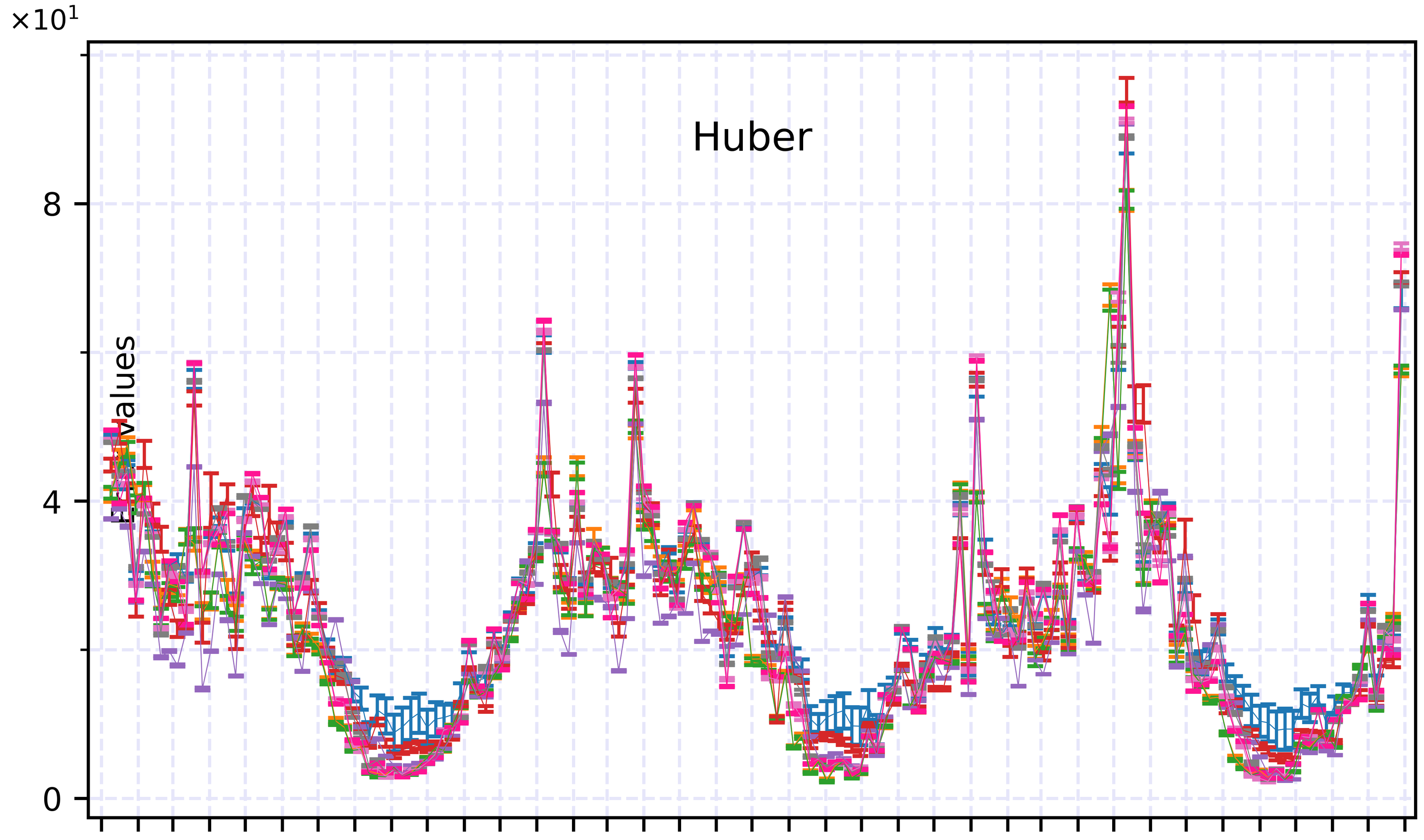}\par
\includegraphics[width=0.95\textwidth]{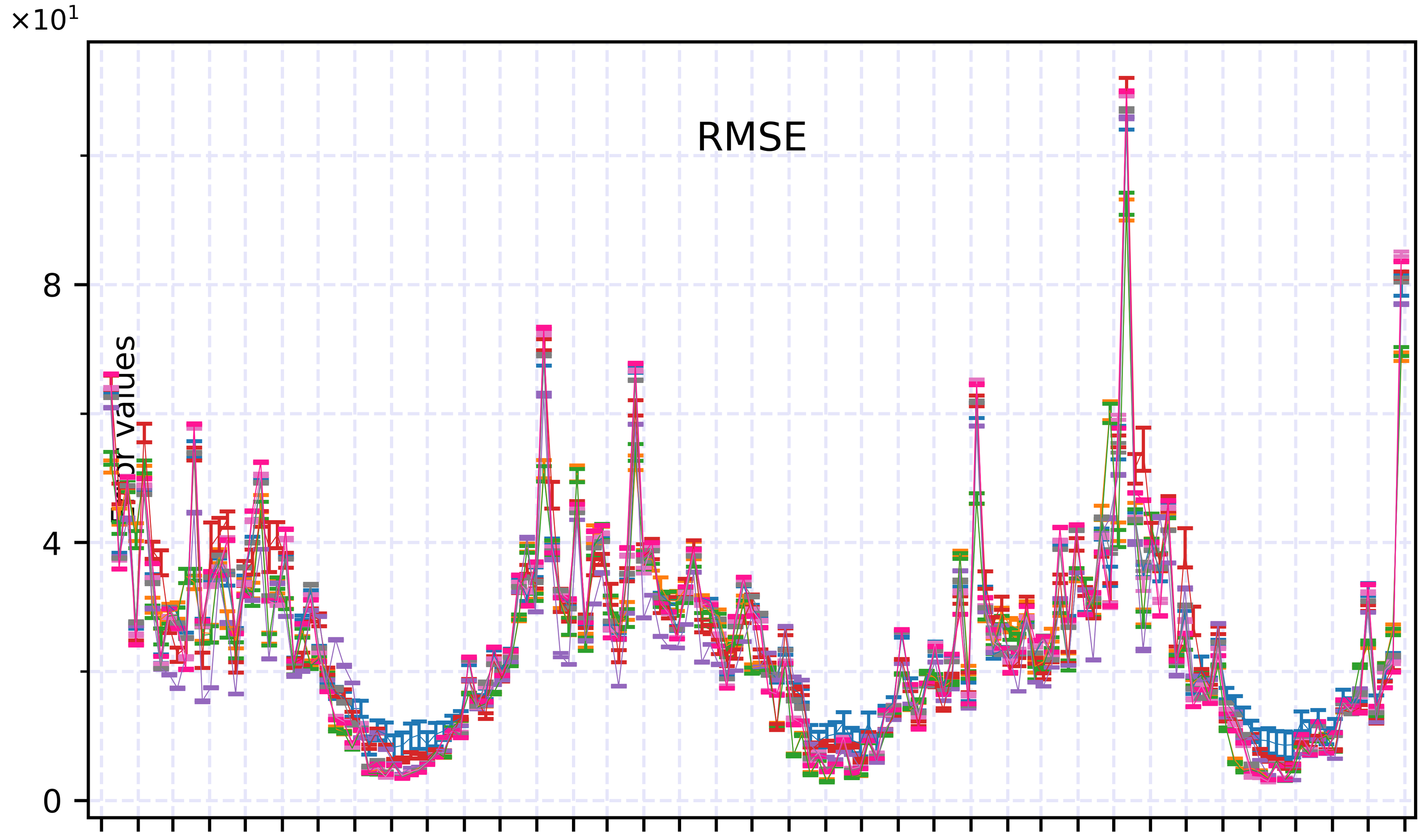}\par
\includegraphics[width=0.95\textwidth]{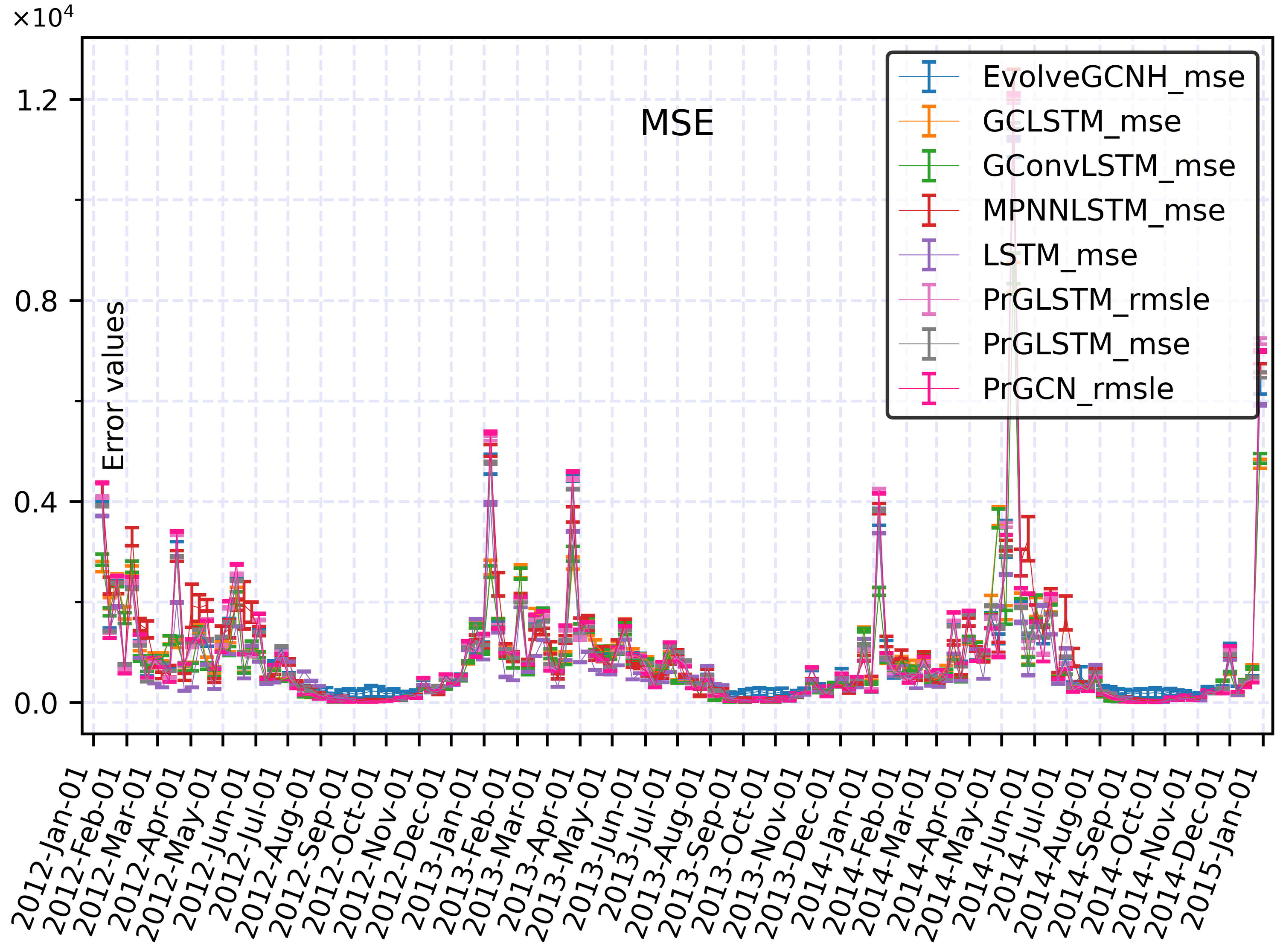}\par

\end{minipage}

\caption{Weekly evolution of global performance metrics across the proposed and benchmark models for chickenpox predictions in Hungary (20 counties).}
\label{fig4:errorstep_hun}
\end{figure}

\clearpage

\newpage
\begin{figure}%[!h]
\centering

\begin{minipage}{1\textwidth}
\raggedright

\includegraphics[width=0.99\textwidth]{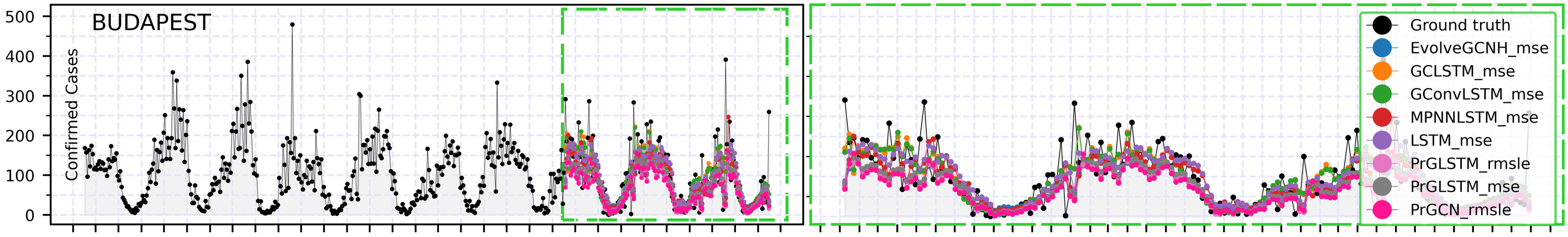}
\includegraphics[width=0.99\textwidth]{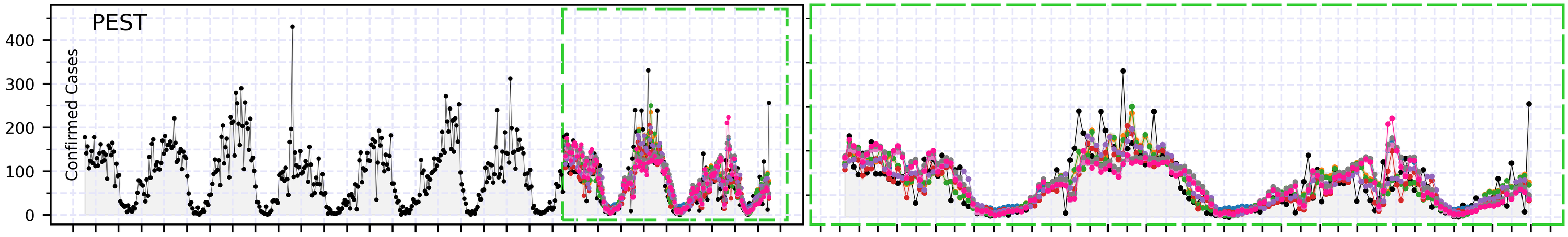}
\includegraphics[width=0.99\textwidth]{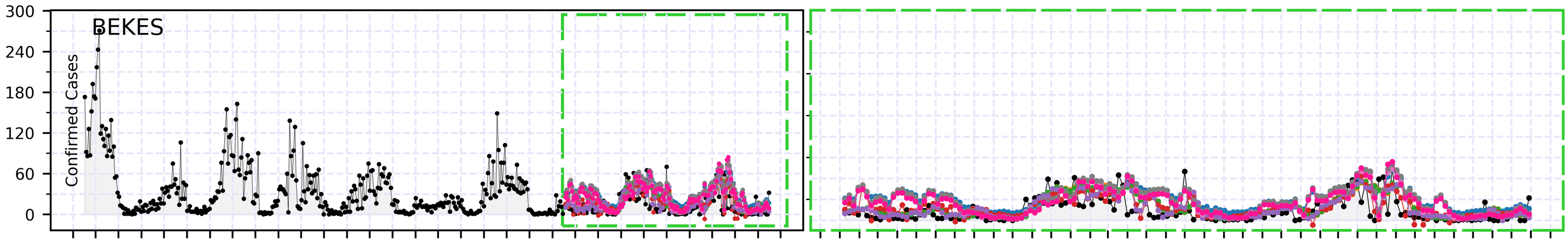}
\includegraphics[width=0.992\textwidth]{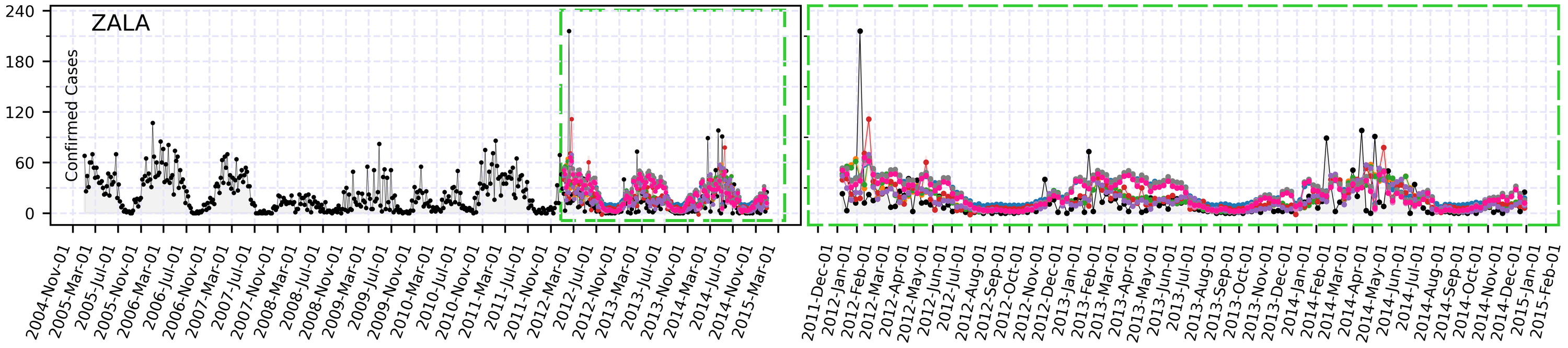}

\end{minipage}

\caption{
Observed time-series data and model predictions for chickenpox in Hungarian counties with the largest weekly incidence using the proposed and benchmark models.
}
\label{fig4:trend_best_hun}
\end{figure}

\textbf{$\mathbf{PrGCN_{rmsle}}$}. Figures \ref{fig4:trend_best_hun} and \ref{fig4:trend_best_hun_ablation} (Appendix \ref{app4:pred})  illustrate the results of the proposed and baseline models. All models often exhibit reduced sensitivity to incidence waves with fluctuating spikes characterised by high-frequency and low-amplitude features. Despite the model capabilities to capture rising slopes of consistent wave trends, it might struggle with rapid local changes, and sharper spikes might be harder to capture. This strong local heterogeneity in smaller networks challenges the smoothing effect of the aggregation process, leading to under- and overestimation of predictions. PrGCN$_{rmsle}$ is able to predict the trending behaviour of most waves in the counties with the highest weekly incidence (Fig. \ref{fig4:trend_best_hun}). There is good performance in predicting waves with average dimensions (Fig. \ref{fig4:tolnaaround_ablation} in Appendix \ref{app4:pred}) while PrGCN$_{rmsle}$ tends to overestimate the peak values of smaller waves. We also observed that PrGCN$_{rmsle}$ smooths out the peak values of waves with an irregular spiking behaviour, resulting in underestimations of wave trends and increasing global weekly errors (Fig. \ref{fig4:errorstep_hun}).

$\mathbf{PrGLSTM_{rmsle}}$. The results reflect a similar performance between the GCN-LSTM layers of PrGLSTM$_{rmsle}$ and the sequence learning in PrGCN$_{rmsle}$ (Fig. \ref{fig4:trend_best_hun}). However, the PrGLSTM$_{rmsle}$ training process requires significantly more time than PrGCN$_{rmsle}$ ($\sim 4.5$ times) (Table \ref{tab4:overhead} in Appendix \ref{app4:pred}). The additional required time can be attributed to the complex processes within the LSTM structure.

\textbf{Ablation models}. The predictions of LSTM-based ablation models slightly align better to the fluctuating spikes in the wave peaks (Fig. \ref{fig4:trend_best_hun_ablation} in Appendix \ref{app4:pred}). This subtle improvement can be partially attributed to the processing of short- and long-term dependencies in the LSTM only for the temporal features, without the influence of the aggregation process. However, the training process time is increased: PrLSTM$_{rmsle}$ requires 2.4 times the PrGCN$_{rmsle}$ timing (Table \ref{tab4:overhead}). Despite stochastic features, PrGCN$_{rmsle}$ produces epidemic trajectories largely similar with ablation models (Fig. \ref{fig4:trend_best_hun_ablation} in Appendix \ref{app4:pred}), suggesting stochasticity has minimal benefit for predictions of overall epidemic trends, supported by closer global weekly errors (Fig. \ref{fig4:errorablat_hun} in Appendix \ref{app4:pred}).

\textbf{Benchmark models}. All models show similar performance (Figure \ref{fig4:errorstep_hun}). In particular, PrGCN$_{rmsle}$ and LSTM$_{mse}$ have the narrowest variability intervals for their global error values, while EvolveGCNH shows the highest variability intervals in its errors.

This indicates PrGCN$_{rmsle}$ is consistent and stable across runs, although a systematic over- and underestimation of incidence magnitudes is observed. However, LSTM$_{mse}$ captures the fluctuating spikes of the wave peaks slightly better with a more consistent alignment to the trending behaviour of the observed waves. In terms of time complexity, the benchmarks require an increased training time, approximately three to five times the PrGCN$_{rmsle}$ timing (Table \ref{tab4:overhead}).

\subsection{Overhead time}
\label{sub4:5time}
Table \ref{tab4:overhead} reports the average time required to run 800 epochs for each model and the corresponding number of iterations per second. PrGCN$_{rmsle}$ displays faster convergence to a solution. This can be explained by its more light-weight and streamlined model structure. PrGCN$_{rmsle}$ is capable of integrating the stochastic component without increasing the training time required, while RNN-based baselines require approximately $\sim$1.5-5 the computational time of PrGCN$_{rmsle}$. In particular, PrGCN$_{rmsle}$ exhibits fewer trainable parameters than the benchmark models that use LSTM layers: GCONVLSTM (5121), GCLSTM (4897) and MPNNLSTM (22405). The benchmark models require longer training times ($\approx$ 3-5 times) as they involve multiple regularisation techniques and aggregation stages, including the many processes related to the RNNs structures.

\section[Discussion]{Discussion}
\label{sec:4dis}

\subsection{Related work}
\label{sec:4back}

\textbf{Spatio-temporal learning.} GNN-based models mostly use spatial and sequential historical data to forecast the incidence of infectious disease in different settings (Table \ref{tab4:modelcomparison} in Appendix \ref{apd4:prework}). These models offer great adaptability and flexibility to incorporate a diverse set of features and integrate other mechanisms for hybrid modelling. Some MPNN-based models have predicted the spread of COVID-19 during low-infection seasons. These periods have been characterised by zero observations and a significant number of cases \cite{pan21,fritz22}. These MPNN models have also been combined with recurrent neural networks (RNNs) and transfer learning methods to analyse the disease spread across small geographical networks in certain European cities \cite{pan21}, and with the zero-inflated Poisson (ZIP) distribution method to improve model accuracy in a medium size network of 401 districts in Germany \cite{fritz22}.

Works based on GCNs use static spatial graphs to represent geographic adjacency \cite{kap20,yu23,dav21,tom22, rodri2021}. GCN models have been integrated with attention mechanisms and RNNs, and also aggregate topological information into the graph structure to forecast COVID-19 cases in 51 US states \cite{yu23}. Follow-up studies incorporate graph theory using the effective reproduction number (Rt) \cite{dav21} and generative models to estimate the cumulative evolution of the disease spread in the counties of Boston and Cambridge, in the US \cite{tom22}. Other approaches involve the study of skipping connections in a static graph based on the 20 most populated counties in the US, where each node is connected to past instances of itself with temporal edges, generating a supergraph \cite{kap20}.

Studies based on GATs have been combined with linear and Elman RNN models in a network of 52 Spanish provinces \cite{mur21}, and a gated recurrent unit (GRU) in 193 counties in the US \cite{gao21}. An attention mechanism similar to GAT has been incorporated in the GNN structure to learn cross-location dependencies for epidemic predictions across 47 regions in Japan and 49 states in the US \cite{den20}. Despite their effectiveness, most of the previous methods model epidemics on a timeline that showcases a relatively limited variability of disease spread dynamics during testing. These studies only exploit disease spread patterns within a small extent of geographical networks, where the addition of more mechanisms can lead to increased model complexity.

\textbf{Long-term epidemiological analysis.} Multiple studies have used extensive data sets that cover years of disease outbreaks and spread trends. An MPNN-based model has been combined with convolutional neural networks (CNNs), RNNs, and attention mechanisms to predict long-term influenza-like illness (ILI) for Japan and the US regions (10-49 location nodes) using 7-15 years of epidemiological data \cite{den20}. Other analyses cover several MPNNs, GCNs, and GATs models that incorporate RNNs structures to examine model performance using 10-year data from Chickenpox cases within the Hungarian population (20 location nodes) \cite{roze21,agg22}. Despite the significant amount of data available for the analysis of disease spread, these works trained their epidemiological models on smaller geographical networks.

\textbf{Large-scale geographical graphs.} Some GNN-based studies have analysed the 1-year spread of COVID-19 in the US. GCN-based models were used for the analysis of counties in 50 states \cite{guo21}. GAT-based models incorporated demographic similarity and geographic proximity of 45 states and 193 counties, including only counties with more than 1,000 confirmed cases \cite{gao21}. This approach was expanded to analyse 52 states and 1,351 counties, and to process cumulative cases and counties with more than 3,000 confirmed cases \cite{wang22}. These approaches used a variety of strategies to handle heterogeneous large networks while preserving its geographical characteristics. However, their analyses excluded portions of the spatio-temporal data set, which may not reflect the real-world dynamics of epidemics, and limit model generalisation.

\textbf{Stochastic learning.} Learning stochasticity has rarely been studied in graph-based learning epidemiological applications. Few works incorporate probabilities into the model formulation to account for the variability of the underlying disease spread process \cite{mur21,fritz22}. Although some probability estimation methods are automated to fit regression modelling in graph-based learning architectures, manual parameter tuning is often necessary based on the data features and complexity of the analysis. This may limit hyperparameter space exploration, which can include subjectivity and scalability issues.

By integrating insights from these four research topics, we consider an extensive time range of high- and low-complexity disease spread trends for the analysis of COVID-19, where the graphs are constructed based on large geographical networks (+3000 nodes) and include settings characterised by zero observations during the course of the disease spread. These features make neighbourhood learning rely on some neighbours with no disease cases at all, which endows the learned parameters with the ability to capture the widest range of underlying spread patterns from the data. This structure makes the learning process robust to disease spread fluctuations across the neighbourhood while incorporating a random formulation in the learning architecture.

\subsection{Performance analysis}
\label{sec:4dis:analysis}
In this study, we introduced a spatio-temporal stochastic graph-based learning architecture to forecast new weekly disease cases, with a focus on understanding the effect of the architecture components over the predictions, e.g. spatial, temporal, and stochastic modules. Our findings elucidate the complex characteristics the proposed architecture can deal with and shed light on its potential epidemiological application to process highly variable disease spread patterns over relatively big and small geographical networks.

The proposed architecture involves the construction of a fundamental and trainable GCNs-based structure that includes an inherent stochasticity to simulate and forecast a disease spread system. Moreover, this data-driven architecture has a light-weight design that globally extends spatial and temporal feature encoding across its network to learn contagion dynamics. Since an observed disease evolution is shaped by a variety of factors that typically differ between regions, e.g., disease attributes, population susceptibility, intervention measures, and level of compliance; their inaccuracies and incompleteness may worsen their implementation in data-driven modelling. As a result, we employed fully and minimally processed data sets of new cases of infectious disease spread to preserve the integrity of the real-world system. The data include geographical and sequential information to predict new cases of COVID-19 in the US and chickenpox in Hungary.

After an extensive experimental setting of the proposed architecture and its baselines, we found that the proposed PrGCN$_{rmsle}$ improves the peak estimation of highly complex and larger waves that do not match the training samples, without compromising training time. This forecasting capability is demonstrated across diverse geographical regions and disease outbreaks. In some cases, the model tends to overestimate some peak predictions for smaller waves, which emphasises the influence of the global minimisation process of the loss function and the network aggregation process over the predictions. These results highlight the potential of spatial learning to be used in future outbreak scenarios, as the model can capture the diverse disease dynamics of the neighbourhood to approach most of the unseen epidemic waves. However, the proposed PrGCN$_{rmsle}$ and its baselines also have a delayed prediction pattern of 1-step (1 week) that affects most of the larger predicted waves.

Although the proposed architecture can approach trending behaviours with 1-step delayed onset for smaller waves in large-scale networks, it does require additional exploration to predict fluctuating spikes of wave peaks across small-scale networks, in settings characterised by more seasonal disease spread patterns and high-frequency, low-amplitude spikes within incidence waves. This is a particular constraint, despite having large extended real-world incidence data for training.

Comparative analysis with the baselines reveal that some of the LSTM-based models tend to slightly improve the estimation -albeit still with 1-step delayed onset- of such fluctuating spikes in the peaks, essentially because such models hold stronger long-, and short-term dependencies between the training epochs. This feature indicates that LSTM-based models are capable of handling particularities of outbreak waves that are similar to the training data, but struggle to get the peaks of larger unseen waves.

In the COVID-19 study with larger networks, some testing data also include smaller waves with fluctuating patterns due to the variable reporting frequency, with values that can oscillate to zero at each step. These patterns were not observed during the training phase. Although the proposed PrGCN$_{rmsle}$ generates 1-step delayed predictions for such waves, the results tend to follow the global trending behaviour of the waves, with a slightly greater visible improvement than its benchmarks. Despite this better predictive approach of PrGCN$_{rmsle}$, the 1-step delayed pattern magnifies its error estimation at each step, which affects its global weekly performance.

We attribute the limitation in predicting the onset of most of the disease waves to the training patterns and the diverse wave onsets, which are also affected by the averaging loss function minimisation process.

Further analysis shows that PrGCN$_{rmsle}$ tends to be statistically stable as its predictions and evaluation metrics have lower variability intervals than its baselines after 100 runs. Such characteristics indicate that PrGCN$_{rmsle}$ performance is enhanced by the stochastic formulation and the model components. In addition, this improvement becomes more significant for highly heterogeneous disease networks with a large-scale spatial structure, since the geographical neighbourhood impacts disease transmission patterns. This suggests the potential applicability of PrGCN$_{rmsle}$ in other complex systems with similar or less heterogeneity in their time-series patterns, and also opens the possibility of integrating more graph randomness strategies and techniques to isolate the learning of high variable disease spread trajectories in large or small network structures for a better outbreak prediction process.

The analysis of COVID-19 and chickenpox as case studies reflects the potential of the proposed PrGCN$_{rmsle}$ to predict the overall temporal progression of incidence in settings with similar particularities. High heterogeneity in transmission dynamics, reporting structures, and spatio-temporal scales might affect the model generalisability, highlighting the need for future work to evaluate the proposed architecture across a wider range of infectious diseases. In terms of processing large-scale temporal and spatial structures, the deployment of graph-based models might face scalability challenges. The construction of graphs and operations with high-resolution data can introduce additional computational cost related to training time and memory usage. These considerations are relevant for the application of graph-base models in real-world systems, suggesting further examination on strategies to improve the architecture design for a broader applicability on large-scale data.

\section[Conclusions]{Conclusions}
\label{sec:4fut}

This study provides an estimate of the weekly disease trajectories of COVID-19 transmission across 3,218 locations in the US during 2022, and Chickenpox spread across 20 regions in Hungary during 2012-2014. Prediction estimates with variability intervals are produced at a given time and space for the proposed model and its baselines after 100 runs. This allowed us to examine the evolution of the observed and predicted transmission patterns to confirm the importance of a more extended network design to accurately predict the complete shape of the unseen outbreak waves in heterogeneous complex systems. This is an important finding, as PrGCN$_{rmsle}$ cannot capture all the intricacies involved with large and smaller outbreaks in highly variable and more constant disease spread scenarios. Overall, the analysis showed that the model captures general trends for disease transmission and provides insight into the influence of various factors.

We evaluated neighbourhood network scales, temporal correlations, and stochastic features, finding that the aggregation mechanism of graph-based models relies on spatial-scale structure and heterogeneous epidemic wave signals. Despite the fact that the proposed architecture exhibits global performance comparable to its benchmarks, most of the visible improvements are not correlated to the global error estimation values.

There are some limitations to our analysis. The real-world incidence and geographic data of COVID-19 and chickenpox used in the US and Hungary settings were provided by public data sources, and limited to the accuracy of their reporting process. The proposed model is entirely dependent on the epidemiological and geographic processed data. For instance, the epidemiological COVID-19 data exhibited negative values which were imputed to be 0 for convenient analysis and therefore may not represent the complete heterogeneity of disease trajectories. The geographic data did not include the complete multi-scale structure for all the unincorporated territories in the US; hence, only one unincorporated territory was considered in the COVID-19 analysis. In addition, the proposed model assumes that the data are complete, so does not adjust for under-reporting data, while the delays observed in reporting COVID-19 cases were handled in part by the aggregation data process in the model.

Since the model may need adaptations to capture the different particularities of complex disease trajectories, we plan to address this by improving the modular architecture and the learning mechanism of the model. Implementing an automated clustering classification module and reconfiguring the model-tuning procedure might improve the estimation of wave peaks and reduce delays. This could also allow a more deeper understanding of the transmission dynamics within regions that share similar characteristics. Furthermore, as we aim to improve the model capabilities for probabilistic prediction estimates, it is important to consider a more extensive stochastic analysis that incorporates more randomness in the graph structure. We are also planning to investigate probabilistic learning architectures that incorporate computationally feasible processes.

\bibliographystyle{unsrt}  
\bibliography{references}  %%% Remove comment to use the external .bib file (using bibtex).

\clearpage

\newpage

\appendix
\renewcommand\thefigure{\thesection.\arabic{figure}}
\renewcommand\thetable{\thesection.\arabic{table}}
\renewcommand\theequation{\thesection.\arabic{equation}}

\setcounter{table}{0}
\setcounter{figure}{0}
\setcounter{equation}{0}

\section{Supplementary methods}
\label{app4:methods}

We provide the mathematical formulations and additional details of the model.

\subsection{Multivariate sequence encoding specification}
\label{app4:mod1}

This module plays a significant role in supporting global comparability between locations and building the temporal and spatial dimensions of a complex system, where disease cases are correlated with the geographical topology of the region under study. Initially, the temporal data are preprocessed to mitigate large negative values and missing data. A normalisation procedure is then defined for the disease cases to scale them in a range from zero to one. The scaling standardises the temporal patterns for the entire data set and allows a stable multivariate learning process for the model. Iteration functions create lagged sequence samples from the scaled data. Considering all locations simultaneously, $S^{(t)} \in \mathbb{R}^{G_{l} \times W \times F_{o}}$ at a given time step $t$ is a single lagged sequence sample with dimensions $(G_{l} \times W \times F_{o})$, where $G_{l} \in \mathbb{N}$ denotes the number of locations under study (or nodes), $W \in \mathbb{N}$ is the temporal observation window size, and $F_{o} \in \mathbb{N}$ is the number of features per location. Each input sample contains $W$ data steps of observations for all $G_{l}$ locations, with $F_{o}$ features recorded at each location. Each element of the lagged sequence sample is denoted as $S^{(t)}_{i,s,d} \in \mathbb{R}$, where $i \in \{ 1, \dots, G_l \}$ identifies a specific node or geographic location, $s \in \{ 1, \dots, W \}$ identifies a time step within the temporal observation window, and $d \in \{ 1, \dots, F_o \}$ denotes the feature index of $S^{(t)}$.

The target is represented as $Y^{(t)} \in \mathbb{R}^{G_l}$, where each element corresponds to a scalar value associated with a node. This is equivalent to a row-vector representation $ \mathbb{R}^{1\times G_l}$ in tensor-based implementations. $Y^{(t)}$ captures the predicted value (target) at the time step $t$ for all locations. This procedure increases the size of the original data while preserving the temporal order, enabling a more detailed and deeper exploration of the underlying temporal patterns.

Each sample is associated with a static graph structure that encodes the spatial relationships between all locations, that is generated from the geographical data structure $G_{raw}$. The graph connectivity is represented by a new edge index tensor $G \in \mathbb{N}^{2 \times c}$, where each column encodes a pair of connected nodes, with node indices $( i,j ) \in \mathcal{E} $, and $c$ denotes the total number of edges. This tensor represents all relationships in the neighbourhood: a source location $i$ that has a connection with a location $j$, also known as edges. Edge attributes are stored in a structure $E \in \mathbb{R}^{c \times F_e}$, where $F_e \in \mathbb{N}$ is the number of features per edge. In this setting, $F_e =1$ denotes a single scalar edge weight per edge. $F_e$ was predefined based on geographic border sharing between neighbouring locations. Thus, the edge attributes $E$ correspond to a vector of scalar edge weights. Consequently, $G$ and $E$ together provide a representation of the edge set for each sample. Therefore, each static graph represents the neighbourhood structure associated with each sample and is used for the model to learn the spatial features of the region under study.

The process enables a more intricate examination of disease temporal features within the geographical network. As a result, the output of this module is a batch of spatio-temporal snapshots. Each temporal snapshot consists of a spatio-temporal tensor $S^{(t)} \in \mathbb{R}^{G_{l} \times W \times F_{o}}$ and the corresponding target $Y^{(t)} \in \mathbb{R}^{1\times G_l}$, paired with the same edge index tensor $G \in \mathbb{N}^{2 \times c}$ and edge attribute vector $E \in \mathbb{R}^{c \times 1}$. The spatio-temporal data set consisting of input-output pairs is denoted by $\mathcal{T}$, where $\mathcal{T} = \{ S^{(t)}, Y^{(t)} \}^T_{t=1} $, with t denoting the sample index and $T$ the total number of snapshots. $G$ and $E$ define the fixed connectivity pattern of the graph structure associated with each sample.

At each step $t$, each location (or node) is characterised by a single feature $F_o =1$ representing the incidence of confirmed cases. Therefore, the spatio-temporal tensor $S^{(t)} \in \mathbb{R}^{G_{l} \times W \times 1}$ is reshaped into a node feature matrix $X^{(t)} \in \mathbb{R}^{G_{l} \times W}$, where each node is associated with a sequence of previous observations $W$ at a given step $t$. The set of batches is partitioned sequentially into training and testing sets using a fixed $70/30$ split. Therefore, $T_{train}$ and $T_{test}$ denote the number of snapshots used for the training and testing periods, respectively. Details regarding the partition strategy for the training and testing sets are provided in Subsection \ref{sub4:43conf}. The batches are then given to the spatio-temporal module, where the model learns the dependencies from the data structure to feed the next module, as shown in Fig. \ref{fig4:blockd}. Rather than having an entire data set for the learning process, this module produces a set of batches with spatial and temporal data.

\subsection{Spatio-temporal processing specification}

\subsubsection{$\mathbf{PrGCN_{rmsle}}$}
\label{app4:PrGCN}

Given each spatio-temporal snapshop $X^{(t)} \in \mathbb{R}^{G_{l} \times W}$, the corresponding target $Y^{(t)} \in \mathbb{R}^{ 1 \times G_{l}  }$, the edge index tensor $G \in \mathbb{N}^{2 \times c}$ and edge attribute vector $E \in \mathbb{R}^{c \times 1}$, where $t= \{1, \dots,T_{train} \} \in \mathbb{N}$, a GCN layer generates $Q$-dimensional descriptors, yielding a tensor $\mathbf{X}^{\prime} \in \mathbb{R}^{G_{l} \times Q}$ at a given step $t$, by applying aggregation and readout methods. These operations support the exchange of information between neighbouring nodes in the graph structure.

Consider an unweighted graph $\mathcal{G=(V, E)}$ that contains $G_{l}$ nodes, a vertex set $\mathcal{V} = \{ 1, ... , G_{l} \}$, and the edge set $G$ with $c$ undirected edges $( i,j ) = ( j,i) \in \mathcal{E} $. During the training process, the representations are learned through layer-wise propagation rules in a stage conditioned to spatio-temporal features of the data.

The operations of the GCN are defined in Eq. (\ref{app4_eqn4_GCN}).

\begin{equation}
    \label{app4_eqn4_GCN}
        \mathbf{X}^{\prime} = \mathbf{\hat{D}}^{-1/2} \, \mathbf{\hat{A}} \, \mathbf{\hat{D}}^{-1/2} \, \mathbf{X} \, \Theta
\end{equation}

where $\mathbf{X}$ is the input feature matrix or the spatio-temporal snapshot $X^{(t)} \in \mathbb{R}^{G_{l} \times W}$ in this study. $\mathbf{\hat{D}}$ is the diagonal degree matrix  with entries $\hat{D}_{ii} = \sum_{j=0} \hat{A}_{ij}$.  $\mathbf{\hat{A}} = \mathbf{A} + \mathbf{I_{G_l}}$ denotes the adjacency matrix with self-loops, where $\mathbf{I_{G_l}}$ is the identity matrix. The adjacency matrix includes values of $1$ if node $i$ is connected to node $j$, and values of $0$ otherwise. The term $\mathbf{\hat{D}}^{-1/2} \, \mathbf{\hat{A}} \, \mathbf{\hat{D}}^{-1/2}$ represents the symmetric normalised adjacency matrix, which ensures stability by rescaling the eigenvalues of the graph Laplacian.  $\Theta$ is a tensor of learnable filter parameters, where $\Theta \in \mathbb{R}^{W \times Q}$ with $W$ and $Q$ as the temporal observation window and the number of filters in the convolutional layer, respectively. The linear transformation $\mathbf{X} \, \Theta$ projects the $W$-dimensional temporal features into a higher $Q$-dimensional latent space. This process allows the model to capture more complex spatial-temporal patterns that are not clearly identified in the $W$-dimensional features. $\mathbf{X}^{\prime} \in \mathbb{R}^{G_{l} \times Q}$ is the resulting representation at a given step $t$.

Symmetric normalisation in neighbourhood aggregation, denoted as the node-wise formulation \cite{kipf17}, describes the mathematical operation of localised message-passing process from the perspective of a single node and is given by Eq. (\ref{app4_eqn4_xedges}). For each node $i$, the $Q$-dimensional temporal descriptor $\mathbf{x}^{\prime}_i \in \mathbb{R}^{1 \times Q}$ at a given step $t$ is calculated as:

\begin{equation}\label{app4_eqn4_xedges}
    \mathbf{x}^{\prime}_i = \sum_{j \in \mathcal{N}(i) \cup \{ i \}} \frac{e_{j,i}}{\sqrt{\hat{d}_j \hat{d}_i}} \mathbf{x}_j  {\theta}
\end{equation}

where $\mathbf{x}_j \in \mathbb{R}^{1 \times W}$ is the sequence input of neighbour $j$, $\theta \in \mathbb{R}^{W \times Q}$ is the learnable weight projection matrix that maps the local node features to a $Q$-dimensional latent space, $e_{j,i}$ denotes the edge weight from the source node $j$ to the target node $i$, and $\sum e_{j,i}$ represents the total weights of the neighbours $j \in \mathcal{N}(i)$ of node $i$ and the self-loop $\cup \{ i \}$. The term $1/{\sqrt{\hat{d}_j \hat{d}_i}}$ describes the symmetric normalisation, which accounts for the degrees of the source node $j$ and the target node $i$, and scales the sum based on the number of neighbours. $\hat{d}_i$ is the augmented degree of node $i$, that is defined as $\hat{d}_i = 1 + \sum_{j \in \mathcal{N}(i)} e_{j,i}$, where the constant $1$ represents the added self-loop.

Edge weights are values that can be associated with independent parameters or might be obtained experimentally. This spatial feature encourages the model to focus on relevant connections from the local neighbourhood. Since the connectivity graph structure is constructed as a function of a geographic network in the region under study, the relevant connections between locations are mainly influenced by physical borders. Setting $e_{j,i} = 1$ for adjacent counties ensures that the aggregation process treats all geographic neighbours as equally influential factors. Therefore, the latent features of the disease are collected by aggregating the information through the geographic network, where each node contribution is normalised relative to the local spatial density by the symmetric normalisation term. This process allows the model to learn different spatial correlations that are invariant to the local connectivity network, with a consistent influence scale. These steps ensure a better understanding of the contextual changes that facilitate the prediction process and prevent bias introduced by varying node degrees, where high-density geographic clusters might dominate the feature space.

\subsubsection{$\mathbf{PrGLSTM_{rmsle}}$}
\label{app4:PrGLSTM}

This version accounts for a second layer added sequentially after the GCN structure: an LSTM. Given each spatio-temporal snapshop $X^{(t)} \in \mathbb{R}^{G_{l} \times W}$, corresponding target $Y^{(t)} \in \mathbb{R}^{ 1 \times G_{l}  }$, edge index tensor $G \in \mathbb{N}^{2 \times c}$, edge attribute vector $E \in \mathbb{R}^{c \times 1}$, and output GCN tensor $\mathbf{X}^{\prime} \in \mathbb{R}^{G_{l} \times Q}$ at a given step $t$, where $t= \{1, \dots,T_{train} \}$ and $j= \{1, \dots,G_l \} \in \mathbb{N}$ denote the sample index of a spatio-temporal snapshot and the node or geographic location, respectively, a LSTM layer generates $8Q$-dimensional temporal descriptors, yielding a tensor $\mathbf{H}^{(t)} \in \mathbb{R}^{G_{l} \times Q \times 8Q}$ at a given step $t$, by  sharing latent temporal features between training iterations, retaining or forgetting the temporal dependencies through mathematical formulations such as the sigmoid function. The different cells in the LSTM layer selectively enhance the learning of sequential relationships over time.

At each step $t$, the outputs $\mathbf{X}^{\prime} \in \mathbb{R}^{G_{l} \times Q}$ from the GCN layer are fed into the LSTM structure (Fig. \ref{fig4:blockd}). First, $\mathbf{X}^{\prime} \in \mathbb{R}^{G_{l} \times Q}$ at a given step $t$ is reshaped into a tensor $\mathbf{X}^{\prime} \in \mathbb{R}^{G_{l} \times Q \times 1} $, which treats each vector of incidence counts at location $j$ as a univariate sequence of length $Q$.

The LSTM processes each location independently, but in parallel across all $G_l$ locations, where each value of a sequence of length $Q$ at the location $j$ is denoted as $x^{\prime}_q \in \mathbb{R}^1$ with a sequence index $q= \{1, \dots,Q \} \in \mathbb{N}$. The operations of the LSTM cells at a given location $j$ and step $t$ are defined in Eq. (\ref{app4_eqn4_LSTM}), where the outputs are conditioned to a context tensor and a previous hidden state.

\begin{equation}\label{app4_eqn4_LSTM}
\begin{split}
    i_q = \Phi(W_{ii} x^{\prime}_q + b_{ii} + W_{hi} h_{q-1} + b_{hi}) \\
    f_q = \Phi(W_{if} x^{\prime}_q + b_{if} + W_{hf} h_{q-1} + b_{hf}) \\
    g_q = \Phi(W_{ig} x^{\prime}_q + b_{ig} + W_{hg} h_{q-1} + b_{hg}) \\
    o_q = \Phi(W_{io} x^{\prime}_q + b_{io} + W_{ho} h_{q-1} + b_{ho}) \\
    c_q = f_q \odot c_{q-1} + i_q \odot g_q \\
    h_q = o_q \odot \Phi(c_q) \\
\end{split}
\end{equation}

where $h_q$, $ g_q$, $c_q$, $x^{\prime}_q$ are hidden, candidate, cell, and input states at a given step $q$ with $h_q, g_q, c_q \in \mathbb{R}^{8Q}$ and $x^{\prime}_q \in \mathbb{R}^1$. The hidden state of the layer at step $(q-1)$ is represented as $h_{q-1} \in \mathbb{R}^{8Q}$, which can also represent the initial hidden state at step $q=0$. The input, forget, and output gates are denoted by $i_q, f_q, o_q \in \mathbb{R}^{8Q}$, respectively.

The input weight matrices are $W_{ii}, W_{if}, W_{ig}, W_{io} \in \mathbb{R}^{8Q \times 1}$ and transform the input state $x^{\prime}_q$ into a suitable representation for each gate. The hidden or recurrent weight matrices are $W_{hi}, W_{hf}, W_{hg}, W_{ho} \in \mathbb{R}^{8Q \times  8Q}$ and transform the previous hidden state $h_{q-1}$ to influence each gate decision. The bias terms associated with each gate are $b_{ii}, b_{if}, b_{ig}, b_{io} , b_{hi}, b_{hf}, b_{hg}, b_{ho} \in \mathbb{R}^{8Q}$. The activation function and Hadamard product are $\Phi(\cdot)$ and $\odot$, respectively. The activation functions are sigmoid for the input, forget, and output gates, and hyperbolic tangent for the candidate and cell states. The full sequence output for all locations $G_l$ stack in a tensor $\mathbf{H}^{(t)} \in \mathbb{R}^{G_{l} \times Q \times 8Q}$ at a given step $t$, which is reshaped into a tensor $\mathbf{H}_{G_l}^{(t)} \in \mathbb{R}^{G_{l} \times 8Q}$ to represent the last hidden state of the process.

During model training, the GCN and LSTM layers sequentially process $T_{train}$ consecutive input sequences without randomly indexing in separate formats, while the LSTM network saves each learning temporal state to enhance the learning of the next iteration. After the GCN layer, the structured latent representation is a nonlinear space-state epidemic system and the LSTM approximates the learning of hidden state evolution with stable memory retention over multi-wave dynamics and non-stationary events. The LSTM component explicitly enables stable propagation of long-term epidemic memory while filtering short-term noise, whereas vanilla RNN or GRU models might provide limited or compressed memory representations.

\subsection{Stochastic method specification}
\label{app4:stoch}

This module addresses internal variability during the training and testing processes of the models PrGCN${_{rmsle}}$ and PrGLSTM${_{rmsle}}$. Our aim is to induce minimal perturbations over the spatio-temporal learning architecture. Given each tensor $\mathbf{Z}^{(t)} \in \mathbb{R}^{G_{l} \times \Gamma}$ at a step $t$, where $\mathbf{Z}^{(t)}$ represents the output of the spatio-temporal processing module, $t= \{1, \dots,T_{train} \}$ denotes the sample index of a spatio-temporal snapshot, $ \Gamma= Q $ and $ \Gamma= 8Q $ correspond to the $Q$- and $8Q$-dimensional descriptor output from PrGCN${_{rmsle}}$ and PrGLSTM${_{rmsle}}$ models, respectively, this module maps each embedding to a 1-dimensional descriptor, producing a tensor $\mathbf{Y}^{\prime} \in \mathbb{R}^{G_{l} \times 1}$ at a given step $t$ by using a fully connected layer and a random sampling formulation to induce stochasticity in the final stage of the prediction process.

Initially, a fully connected stochastic layer, also known as a multi-layer perceptron (MLP) \cite{mlp91}, processes the multiple input arguments $\mathbf{Z}^{(t)} \in \mathbb{R}^{G_{l} \times \Gamma}$ from the spatio-temporal module. At each step $t$, the MLP processes the data from each location independently while sharing weight parameters across all locations $G_l$, with computations performed in parallel. After that, a representation of the latent features is decoded in a tensor $\mathbf{P}^{(t)} \in \mathbb{R}^ {G_{l} \times 2}$, where $t= \{1, \dots,T_{train} \}$ and $j= \{1, \dots,G_l \} \in \mathbb{N}$ denote the sample index of a spatio-temporal snapshot and the node or geographic location, respectively. The universal function approximation of the MLP allows flexible nonlinear decoding from the learned latent representation of the spatio-temporal module to a 2-dimensional descriptor tensor. The previous module structurally reduces variance, while the MLP provides expressive final mapping without reintroducing additional spatial and temporal structural priors that might risk over-smoothing the final representation.

At each location $j$, the elements of the tensor $\mathbf{P}^{(t)}$ are indicated as $[ \mathbf{p}^{(t)}_{j,1} , \mathbf{p}^{(t)}_{j,2} ] $ at a given step $t$, with $\mathbf{p}^{(t)}_{j,1} , \mathbf{p}^{(t)}_{j,2} \in \mathbb{R}$. Then, a mapping strategy is employed over the elements of the tensor, where $\mathbf{p}^{(t)}_{j,1}$ is equivalent to the form $ln \, \sigma^{(t)}_{j}$ to get $\sigma^{(t)}_{j} \in \mathbb{R}$, and $\mathbf{p}^{(t)}_{j,2} \in \mathbb{R}$ is represented by $\mu^{(t)}_{j} \in \mathbb{R}$. This strategy allows to generate normal distributions $\mathfrak{D}_{j}^{(t)}$ from the outputs, where $ \sigma^{(t)}_{j}$ and $\mu^{(t)}_{j}$ are considered the standard deviation and mean parameters of $\mathfrak{D}_{j}^{(t)}$.

We process each $\mathfrak{D}_{j}^{(t)}$ via parameterised stochastic and deterministic operations that transform standard parameter-free distributions called a reparameterisation trick operation \cite{king13}. The model learns the distribution and pulls a random value from the fixed distribution. Let a final prediction $ \hat{y}_{j}^{(t)} \in \mathbb{R} $ at step $t$ and location $j$ be determined by $\hat{y}_{j}^{(t)} = \mu^{(t)}_{j}+\sigma^{(t)}_{j} \odot\epsilon_j$, with an auxiliary noise variable $\epsilon_j$. This randomness is external to the learnable components of the model and allows the model to represent a range of possibilities. This formulation ensures that the random element remains independent of the network weights, while enabling gradient-based optimisation when sampling from a probabilistic latent variable.

The final predictions for all locations $G_l$ stack in a tensor $ \hat{Y}^{(t)} \in \mathbb{R}^{G_{l} \times 1} $ at a given step $t$. These values have well-defined gradients to guide the model optimisation process and are used to tune the model and evaluate its performance. The internal interactions of the stochastic module involve multiple operations to grant stochastic freedom to the model, allowing the learning of the latent features from the complex system and introducing randomness into the forward pass of the network.

\subsection{Loss function specification}
\label{app4:loss}

The proposed architecture is tailored to be trained in an end-to-end manner using a Root Mean Squared Logarithmic Error (RMSLE) as a loss function ($\mathcal{L}_{R\!M\!S\!L\!E}$), defined in Eq. (\ref{app4:eqn4_mNLLloss}). Given a target tensor $Y^{(t)} \in \mathbb{R}^{G_l}$ and a model prediction tensor $ \hat{Y}^{(t)} \in \mathbb{R}^{G_{l} \times 1} $, where the elements of each tensor at a given location $j= \{1, \dots,G_l \} \in \mathbb{N}$ and step $t= \{1, \dots,T_{train} \} \in \mathbb{N}$ are denoted as $ {y}_{j}^{(t)} , \hat{y}_{j}^{(t)} \in \mathbb{R} $, respectively, $\mathcal{L}_{R\!M\!S\!L\!E}$ calculates the error using the squared difference of logarithms. By minimising this difference, the loss function ensures that predictions for small- and large-scale locations are weighted with equal relative impact on the geographic network.

\begin{equation}\label{app4:eqn4_mNLLloss}
\begin{aligned}
\mathcal{L}_{R\!M\!S\!L\!E} = \sqrt{ \sum_{j=1}^{|G_l|} \Bigl(  \, log (\hat{y}_{j}^{(t)} \, + \, 1) \, - \,  log ({y}_{j}^{(t)} \, + \, 1)    \Bigr)^2}
\end{aligned}
\end{equation}

The $\mathcal{L}_{R\!M\!S\!L\!E}$ penalises the inaccuracies of the difference of logarithmic values. Since predictions of disease cases account for the estimation and even an overestimation of wave peaks, $\mathcal{L}_{R\!M\!S\!L\!E}$ supports the PrGCN and PrGLSTM learning process by assigning larger penalties for the underestimation of disease cases than for overestimation.

$\mathcal{L}_{R\!M\!S\!L\!E}$ also considers a relative penalty between the predicted and real values of disease cases. In contrast, other metrics, such as the MSE, increase the penalty if the scale of the estimation and real values increase. 

Although $\mathcal{L}_{R\!M\!S\!L\!E}$ is an extension of the Mean Squared Error (MSE) metric, $\mathcal{L}_{R\!M\!S\!L\!E}$ is a robust measure in terms of outliers, allowing the model to learn a more stable trend across the geographic network. The loss is calculated globally at each time step using the root of an arithmetic sum of the squared logarithmic difference for all nodes. By using a global loss function, we assume that the input features at a given time are uniformly distributed and equally important across all node locations, which helps the model account for variability in disease transmission across neighbourhood connections. Consequently, $\mathcal{L}_{R\!M\!S\!L\!E}$ guides the model to focus on the main global dynamical features instead of local fluctuations, to determine the best fit $\mathfrak{M}$.

This consideration is critical because it allows for capturing a global spread pattern of the course of disease outbreaks, which, in turn, yields more information to the locations with low reach and spread of the disease. For example, the US states typically handled a variety of intervention measures during the COVID-19 outbreaks, which differed between locations  \cite{hallas21}. As a result, each location exhibits a disease transmission trend in a widely different fashion. In this case, $\mathcal{L}_{R\!M\!S\!L\!E}$ helps the proposed model learn the dominant trends of COVID-19 spread across the neighbourhood topology over time.

The proposed model is trained by directly minimising an empirical loss $\mathcal{L}_{R\!M\!S\!L\!E}$ as a function of $ {y}_{j}^{(t)} , \hat{y}_{j}^{(t)} \in \mathbb{R} $ at a given step $t$. In addition, a total loss value per epoch $\mathcal{L}_{epoch} \in \mathbb{R}$ is also empirically computed as the average over all nodes and temporal steps in the training data. During training, gradients are computed via backpropagation to iteratively update the model parameters using stochastic optimisation. Multiple runs are performed to evaluate the convergence of the loss trajectories. The parameters are considered optimised by observing stabilisation of $\mathcal{L}_{epoch}$ over the epochs. In this process, $\mathcal{L}_{epoch}$ is computed exclusively on the training set without the need for an external validation process, while the testing set is used for the out-of-sample evaluation and is not involved in parameter updates.

\subsection{Uncertainty specification}
\label{app4:unc}

The proposed architecture accounts for the capture of epistemic uncertainties using an ensemble method. This approach allows us to maintain the structural integrity of the architecture and the same values of hyperparameters and data input. The process involves varying only the sources of stochasticity that affect the initialisation of model parameters. Specifically, the proposed architecture is trained multiple times using different initialisations (different seed values), yielding distinct learned parameters. As a result, the disease predictions of several individual independent simulations of the model are processed to obtain a more robust and accurate prediction with an estimate variability interval that captures the epistemic uncertainty.

After training, the learnable parameters of the model is optimised, yielding a fixed mapping $\mathfrak{M}$. To ensure the reproducibility and robustness of the findings, we train the model multiple times with different seed values and generate a ensemble of $R \in \mathbb{N}$ independent simulations. Each $\mathfrak{M}_r$ with $r= \{1, \dots,R \} \in \mathbb{N}$ is applied to the unseen data during the testing phase. This process allows to obtain the model predictions $ \hat{Y}_r^{(t)} \in \mathbb{R}^{G_{l} \times 1} $ at a given step $t= \{1, \dots,T_{test} \} \in \mathbb{N}$ for several individual mappings $[\mathfrak{M}_1, \mathfrak{M}_2,... , \mathfrak{M}_{R-1}, \mathfrak{M}_{R}]$. After $R$ runs, a tensor $ \hat{Y}_E \in \mathbb{R}^{G_{l} \times T_{test} \times R} $ with $G_l, T_{test}, R \in \mathbb{N}$ represents the set of predictions from the testing phase.

Variability in the ensemble predictions reflects sensitivity to parameter estimation and provides an estimate of epistemic uncertainty through empirical variance, which is used to construct approximate variability intervals around the mean prediction. This diversity provides an alternative to estimate the epistemic uncertainty and supports a more comprehensive analysis of the different aspects of the model learning process as a function of the predictions from the testing phase.

We assumed that the ensemble of predictions $ \hat{Y}_E \in \mathbb{R}^{G_{l} \times T_{test} \times R} $ with $j= \{1, \dots,G_l \} \in \mathbb{N}$, $t= \{1, \dots,T_{test} \} \in \mathbb{N}$, and $r= \{1, \dots,R \} \in \mathbb{N}$ follows a normal distribution along the dimension $R$. At each step $t$ and location $j$, a 95\% variability interval is estimated as a function of a sample mean $(\mu_E)$, a sample standard deviation $(\sigma_E)$, and number of observations $R$  from $\hat{Y}_E \in \mathbb{R}^{G_{l} \times T_{test} \times R}$. As a final result, a more robust disease prediction is represented by $\mu_E$ with a 95\% variability interval. These intervals represent the estimate of the epistemic uncertainty at each step $t$ and location $j$, where high variability suggests that the final predictions are sensitive to a specific model initialisation.

\section{Supplementary experiments}

\subsection{Dataset specification}
\label{app4:data}

We conducted extensive experiments to evaluate the capability of the proposed architecture in predicting new cases of COVID-19 in the US and new cases of Chickenpox in Hungary. Two types of data sets are used for each type of disease: new weekly cases and geographical connections. Since the multivariate sequence encoding module is designed to handle the same time series lengths, the data are preprocessed to account for missing values and negative values for the regions under study (these values were found in the original data).

To examine how sensitive the architecture is to handle a variety of trends in disease spread, each data set represents a particularity of the complex nature of disease spread. The analysis of COVID-19 spread involves highly complex spread patterns across the US from 2020 to 2022, while the analysis of Chickenpox spread accounts for a more consistent disease presence in Hungary from 2005 to 2014. We conducted a single-scale analysis at a county spatial level to examine the neighbourhood influence over the disease spread at a specific node location and to explore the impact of the stochastic formulation, spatio-temporal components, and loss function over the prediction values.

\subsubsection{COVID-19 data} \label{apd4:second}

COVID-19 is a highly contagious viral illness. This experiment covers COVID-19 spread in the US during 2020 to 2022, which is characterised by multiple waves of different shapes and peak sizes, where newer peaks are usually larger than the previous ones, except for the smaller Spring waves towards the end of 2022. Since COVID-19 transmission presents as rapidly spreading during an outbreak, the goal is for a model to learn through neighbouring features a representation $\mathfrak{M}$ to generate better estimates of disease cases. In particular, the aim of the model is to focus on diverse trajectories with abrupt growth, the bigger waves, and the prediction of their peaks. In this context, the underestimation of peaks represents an issue as the model is required to handle nonlinear escalation and capture fast and rising waves in heterogeneous settings.

COVID-19 county-level surveillance data were obtained from a Data Repository managed by the Center for Systems Science and Engineering (CSSE) at Johns Hopkins University \cite{jhu20}. This repository included data of COVID-19 spread for all countries and was publicly accessible.

Information was collected entirely from open-source data sources and then verified in reports from governments and health authorities when they become available. Raw county detailed data are only available for the US territory.

The raw data for COVID-19 mainly cover confirmed reports of new cases and deaths, dates, country name, state dependency, and only for US counties: county name and Federal Information Processing Standards code (FIPS). Despite the JHU team's efforts to curate the data, some issues arose when daily counts were analysed, such as errors in reporting that generate subsequent updates to correct them.

\begin{table}%[ht!]
\centering

\caption{\label{tab4:reporting} Irregular reporting frequencies observed in daily incidence data of COVID-19 for territories and corresponding counties in the US.}

%\resizebox{\columnwidth}{!}{%
{\fontsize{8pt}{10pt}\selectfont
\begin{tabular}{lcc}
\\[-0.75em]
\hline
\\[-0.75em]
\emph{Location} & \emph{Reporting} & \emph{Year}  \\
\\[-0.75em]
\hline
\\[-0.75em]
\rule{0pt}{12pt} Florida (FL) \& 67 counties & 4-14 days & 2021-2022   \\  

\rule{0pt}{12pt} Georgia (GA) \& 159 counties   & 4-7 days & 2022  \\

\rule{0pt}{12pt} Iowa (IA) \& 99 counties  & 4-7 days & 2021-2022  \\  

\rule{0pt}{12pt} Kansas (KS) \& 105 counties  & 4-7 days & 2022  \\  

\rule{0pt}{12pt} Kentucky (KY) \& 120 counties  & 4-7 days  & 2022 \\  

\rule{0pt}{12pt} Massachusetts (MA) \& 14 counties   & 4-7 days  & 2022  \\  

\rule{0pt}{12pt} Michigan (MI) \& 83 counties  & 4-7 days & 2022 \\  

\rule{0pt}{12pt} Minnesota (MN) \& 87 counties   & 4-7 days & 2022  \\  

\rule{0pt}{12pt} Mississippi (MS) \& 82 counties  & 4-7 days & 2022  \\  

\rule{0pt}{12pt} Missouri (MO) \& 114 counties  & 4-7 days  & 2022 \\  

\rule{0pt}{12pt} Montana (MT) \& 56 counties  & 4-7 days  & 2022  \\  

\rule{0pt}{12pt} Nevada (NV) \& 17 counties   & 4-7 days  & 2022  \\  

\rule{0pt}{12pt} North Carolina (NC) \& 100 counties   & 4-7 days  & 2022   \\  

\rule{0pt}{12pt} Ohio (OH) \& 88 counties   & 4-7 days  & 2022  \\  

\rule{0pt}{12pt} Oklahoma (OK) \& 77 counties   & 4-7 days  & 2021-2022  \\  

\rule{0pt}{12pt} Pennsylvania (PA) \& 67 counties   & 4-7 days  & 2022   \\  

\rule{0pt}{12pt}  Rhode Island (RI) \& 5 counties   & 4-7 days  & 2020-2022   \\  

\rule{0pt}{12pt} South Carolina (SC) \& 46 counties   & 4-14 days  & 2022  \\  

\rule{0pt}{12pt} South Dakota (SD) \& 66 counties   & 4-7 days  & 2022   \\  

\rule{0pt}{12pt} Utah (UT) \& 29 counties   & 4-7 days  & 2022  \\  

\rule{0pt}{12pt} Vermont (VT) \& 14 counties   & 4-7 days  & 2022   \\  

\rule{0pt}{12pt} Wyoming (WY) \& 23 counties   & 4-7 days  & 2022   \\  
\\[-0.75em]
\hline

\end{tabular}%
}
\footnotesize
    {
    \parbox{1\linewidth}{
    \textit{Note:} Frequency of reporting is detailed with referential values (column \textit{Reporting}), and reporting is in place during specific time periods (column \textit{Year}).
    }
    }
\end{table}

In this study, we analysed the confirmed daily counts and corresponding spatial and descriptor information. To account for the temporal variance, we used the raw data aggregated over a temporal window of different resolutions, denoted $X_{raw,s_r}$, where $s_r \in \mathbb{N}$ defines the temporal aggregation window, with $s_r=1$ for daily counts, while $s_r=3$ and $s_r=7$ correspond to 3-day and weekly aggregates.

The resulting data sets consisted of static and non-overlapping observations, where a larger $s_r$ yields a smaller set of time-series observations for each location. We first derived a data set composed of a tensor $ X_{raw,1} \in \mathbb{R}^{G_l \times T_{s_r}} $, a location descriptor $N_l$, and a temporal index set $D_{l}$, where $G_l$ indicates the number of locations. In this case, locations refer to counties in the US, and $T_{s_r}$ represents the total number of observations for a given resolution $s_r$. $N_l$ provides the county names and their FIPS codes, while $D_{l}$ specifies the dates in chronological sequence associated with each time step. Therefore, $X_{raw,1}$ represents the daily confirmed cases of COVID-19 in the set of $G_l$ counties during a period $t_l$.

As the data in the study correspond to the period from January 2020 to October 2022, $T_{s_r}$ is 987, which denotes a sequence of 987 consecutive days for confirmed cases of COVID-19 per location, while $G_l$ is 3,218, which reflects the number of county locations in the US under study.

Geographical county adjacency data for the US were obtained from the US Census Bureau source. The data cover county names with their FIPS codes and state territory abbreviations. Data were organised into columns where each county is associated with itself and its adjacent neighbours. Some issues arose with some mixed county names and FIPS codes, which required additional processing. From these geographical data, a raw connectivity structure $G_{raw}$ and a location descriptor $N_G$ were derived. The structure $G_{raw}$ encodes the connections between the source and destination locations across $G_l$ locations, while $N_G$ contains the corresponding county names and their FIPS codes.

Several filtering stages were necessary to extract complete information on case counts per location from $ X_{raw,1} $ that was linked to a structure of connectivity relationships from $G_{raw}$. Therefore, in this study, 3,218 county locations and 22,154 geographic connections were analysed. Later, additional matrices were derived to correlate confirmed cases with those associated with the territories and national scale.

Similar processes were developed with the relationship matrices of the connectivity structure. After a merging stage, a data set of confirmed cases involved 3,218 locations with COVID-19 confirmed cases over 987 days. The 3,218 locations correspond to 52 state equivalent entities in the US. The data were again processed with the aim of eliminating noise and, as a result, we derived new temporal data sets $\, X_{raw,3} \,$ and $\, X_{raw,7} \,$ by aggregating the initial daily counts in $\, X_{raw,1} \,$. These new data correspond to 3-day and weekly aggregation periods, and each data set consists of static and non-overlapping observations.

\textbf{COVID-19 variable reporting} \label{apd4:first}. Identification and management of fluctuating patterns in the data are important, since the oscillating trends might affect the training and testing phases of the model. During the COVID-19 outbreaks, the time-series data were affected by different reporting frequencies, and the trends of daily confirmed cases were characterised by highly oscillating periods. Such confirmed case values oscillate between zero and the accumulated count value for the delayed reporting period. Table \ref{tab4:reporting} illustrates that the reporting frequency varied between regions, which depended on policies in place at each location. For example, reporting timing might normally vary from 4 to 7 days, while in some locations, reporting counts can be available every 14 days (2 weeks).

To minimise the rate of fluctuations in the incidence trajectories, the daily reporting data $\, X_{raw,1} \,$ were temporally aggregated to produce the data sets $\, X_{raw,3} \,$ and $\, X_{raw,7} \,$. As a result, confirmed counts in $\, X_{raw,7} \,$ were represented by time-series trends without oscillations for the majority of locations within the US topology network. However, some time series still showed oscillations in the 2022 spring-summer wave due to a different set of reporting frequency. This pattern was observed in Florida, Mississippi, and South Carolina states and their corresponding counties due to a variable reporting frequency placed between 2021 and 2022, as shown in Fig. \ref{fig4:trend_FLc}. These fluctuating patterns can have a significant impact on the testing phase of the model, because this chaotic behaviour is hugely different from the former learning parameters of the training phase. Each start of these oscillations might be processed as a rising wave, which might be erroneously associated with a small representation of the increase phase of high-amplitude incidence waves.

On the other hand, the temporal data set $\, X_{raw,3} \,$ tended to not improve the fluctuating patterns. For example, Rhode Island and its five counties showed oscillations in their time-series trends due to irregular reporting frequencies since 2020. These oscillations affected the shape of small and large waves. Oscillations as training data can be helpful to the learning process of the model, however, the small sample size might represent a significant issue. Another notorious feature was the variable starting point of the outbreak waves across the topology network. The 99 counties in Iowa (IA) state are examples where the rise of the waves occurred at different points through the summer of 2021 to winter of 2022. There were also significant differences in the wave features between these counties.

Finally, COVID-19 data were structured in batches where $G_l = 3,218$, $c = 22,154$. With a 70/30 split for the train/test data, the training data are from January 22, 2020 to December 11, 2021; and the testing data are from December 12, 2021 to October 4, 2022. In the training period, outbreak patterns are characterised by multiple waves with rapid increases and decreases in the number of cases during 2020-2021. During the testing period, high variability is exhibited during the first months of 2022, and the waves have larger peaks compared to previous wave patterns in the training period. The data also include counties with few or no cases during 2020-2022, in training and testing periods.

\subsubsection{Chickenpox data} \label{app4:chick}

Chickenpox is an illness that spreads easily through airborne droplets or direct contact. After having chickenpox, most people have lifelong immunity. This experiment covers the spread of Chickenpox disease in Hungary during 2005 to 2014, which has seasonal patterns with a constant presence of waves with similar shapes and peak sizes. The goal of this experiment is to learn a representation $\mathfrak{M}$ that predicts wave patterns with some fluctuating spikes and avoids underestimating the cases.

Chickenpox data were structured in batches where $G_l = 20$, $c = 102$. With a 70/30 split for the train/test data, the training data are from January 3, 2005 to January 2, 2012; and the testing data are from January 3, 2012 to December 29, 2014. Disease spread patterns are characterised by multiple waves with cyclical occurrence and similar peak sizes from 2005 to 2014, for training and testing periods.

\subsection{Evaluation metrics specification}
\label{apd4:third}

Table \ref{table4:errordef} presents the mathematical equations of the error metrics used to evaluate the performance of the models. Each metric is defined in terms of the observed and predicted values at a given step $t$, where $y(x)$ denotes the ground true value, $\hat{y}(x)$ corresponds to the prediction, and $\Omega$ is the total number of observations.

\begin{table}%[hb!]
    \centering
    \caption{Error metrics with a mean reduction ($1/{|\Omega|}$), with $\Omega$ as the number of samples.}
    \label{table4:errordef}

    {\fontsize{8pt}{10pt}\selectfont
    \begin{tabular}{ll}
        \hline
        \\[-0.75em]
        Metric & Definition\\
        \\[-1.00em]
        \hline
        \\[-0.65em]
         MSE & $\frac{1}{|\Omega|} \sum_{x \in \Omega} |\, \hat{y}(x)-y(x)\,|^2 $\\
         \\[-0.65em]
         MAE & $\frac{1}{|\Omega|} \sum_{x \in \Omega} |\, \hat{y}(x)-y(x)\,| $\\
         \\[-0.65em]
         Huber & $\frac{1}{|\Omega|} \sum_{x \in \Omega} \big($ $\frac{1}{2}|\hat{y}(x)-y(x)|^2$ for $|\hat{y}(x)-y(x)|\leq\delta$,\\
         &  $\,\delta . (|\hat{y}(x)-y(x)|-\frac{1}{2}\delta)$, otherwise $\big)$, with $\delta = 1.35$\\
         \\[-0.75em]
         RMSE & $\big( \frac{1}{|\Omega|} \sum_{x \in \Omega} |  \,  \hat{y}(x) \, - \,  {y}(x)  \,   |^2\big)^{1/2}$\\
         \\[-0.75em]
         RMSLE & $\big( \frac{1}{|\Omega|} \sum_{x \in \Omega} |  \, log (\hat{y}(x) \, + \, 1) \, - \,  log ({y}(x) \, + \, 1)    |^2\big)^{1/2}$\\
         \\[-0.75em]
         \hline
    \end{tabular}
    }

\end{table}

\subsection{Baselines specification}
\label{app4:base}

To evaluate the effect of the architecture components during the single-scale analysis, we provide ablation studies for the experiments.

All ablation variants are trained under identical conditions, following the setup detailed in Section \ref{sub4:43conf}, with modifications as follows:

\begin{itemize}
    \item {PrLSTM${_{rmsle}}$} is an ablation variant that allows us to assess the contribution of the spatio-temporal module while keeping the rest of the architecture unchanged. The module is replaced with a single LSTM layer (64 neurons). Since the LSTM accounts for temporal patterns, this variant processes sequences of the multi-locality data independently, without the spatial influence of the connectivity network.
    \item {PrLSTM${_{mse}}$} is a combined baseline variant to evaluate the effect of simplifying both the spatio-temporal module and the loss function, while keeping all other components unchanged. The spatio-temporal module is replaced with a single LSTM layer (64 neurons) and the loss function is replaced with mean squared error (LSTM+MSE). The results of this variant are interpreted alongside other ablations that isolate individual contributions of the architecture.
    \item {PrGCN${_{mse}}$} is a variant that replaces the loss function of the proposed PrGCN$_{rmsle}$ to determine whether the proposed loss provides tangible benefits. The use of a standard training objective, such as a mean squared error (MSE), is motivated by practical considerations due to simplicity and stability. 
    \item {PrGLSTM${_{mse}}$} is a variant that replaces the loss function of the proposed PrGLSTM$_{rmsle}$ with a MSE loss.
    \item {GCN${_{mse}}$ \cite{kipf17}} is a variant that removes the stochastic formulation and loss function of the proposed PrGCN$_{rmsle}$, and retains only a GCN with a MSE loss to offer insight into the impact of the proposed model compared to a standard setup.
    \item {LSTM${_{mse}}$ \cite{hoch97}} is a basic baseline. It removes the stochastic formulation and loss function of the proposed PrGLSTM$_{rmsle}$, and considers a minimal configuration of the spatio-temporal module with a LSTM and MSE loss. Since this configuration does not isolate individual contributions, it is used as a minimal baseline.
\end{itemize}

We further compare the proposed models, PrGCN${_{rmsle}}$ and PrGLSTM${_{rmsle}}$, against four spatio-temporal recurrent graph-based models as a benchmark study. The benchmark models are EvolveGCNH \cite{par20}, GCLSTM \cite{chen2022gc}, GConvLSTM \cite{youngjoo18}, and MPNNLSTM \cite{pan21}. Their architectures consist of different configurations of GCN and RNN layers, and have been developed in the PyTorch Geometric Temporal library. All benchmark models are trained under identical conditions.

The configurations of their implementation settings are similar to the proposed architecture (Section \ref{sub4:43conf}) in terms of epochs and learning rate. Their hidden and output layer configurations depend on each architecture setting; and the loss function is the MSE metric with a sum as a reduction for the four models.

\section{Supplementary results}

\subsection{Global performance metrics}
\label{app4:glob}

Tables \ref{tab:benchmark_rmsle} and \ref{tab:benchmark_mse} show the global estimates of performance metrics for the proposed models and baselines. Model predictions were obtained from the ensemble structure after 100 runs, as described in Section \ref{sub4:43conf}.

\begin{table}[h!]
    \caption{\label{tab:benchmark_rmsle} Global estimation of performance metrics (RMSLE, MAE and Huber) with 95\% variability intervals for all models.}
    %\tiny
    \centering
    %\resizebox{\columnwidth}{!}{%
    {\fontsize{8pt}{10pt}\selectfont
    \begin{tabular}{llcr}
        %\toprule
        \\[-1.25em]
        \multicolumn{1}{l}{Metric} & \multicolumn{1}{l}{Method} & \multicolumn{1}{c}{COVID-19} & \multicolumn{1}{c}{chickenpox}\\ 
        
        \\[-1.00em]
        
        \hline
        \hline
        \\[-0.75em]
        RMSLE & EvolveGCNH & 1.0644 \textit{$\pm 0.0257$} & 1.1511 \textit{$\pm 0.0266$} \\
        \\[-1.00em]
        & GCLSTM & 1.0668 \textit{$\pm 0.0121$} & 0.9607 \textit{$\pm 0.0057$} \\
        \\[-1.00em]
         & GConvLSTM & 1.0744 \textit{$\pm 0.0145$} & \underline{0.9514} \textit{$\pm 0.0055$} \\
        \\[-1.00em]
         & MPNNLSTM & \underline{0.9289} \textit{$\pm 0.0037$} & 1.0795 \textit{$\pm 0.0070$} \\
        \\[-1.00em]
         & LSTM$_{mse}$ & 0.9569 \textit{$\pm 0.0009$} & 0.9809 \textit{$\pm 0.0012$}  \\
        \\[-1.00em]
        & GCN$_{mse}$ & 1.0047 \textit{$\pm 0.0000$} & 1.1033 \textit{$\pm 0.0000$}  \\
        \\[-1.00em]
        \hline
        \\[-0.75em]
         & $\mathbf{PrGLSTM_{rmsle}}$ & 1.0368 \textit{$\pm 0.0033$} & 1.0640 \textit{$\pm 0.0034$} \\
        \\[-1.00em]
        & PrGLSTM$_{mse}$ & 1.0495 \textit{$\pm 0.0017$} & 1.0777 \textit{$\pm 0.0023$} \\
        \\[-1.00em]
        & PrLSTM$_{mse}$ & 0.9738 \textit{$\pm 0.0022$} & 0.9915 \textit{$\pm 0.0029$} \\
        \\[-1.00em]
        & PrLSTM$_{rmsle}$ & 0.9774 \textit{$\pm 0.0015$} & 0.9615 \textit{$\pm 0.0014$} \\
        \\[-1.00em]
        & PrGCN$_{mse}$ & 1.0390 \textit{$\pm 0.0024$} & 1.1249 \textit{$\pm 0.0027$} \\
        \\[-1.00em]
        & $\mathbf{PrGCN_{rmsle}}$ & 1.0150 \textit{$\pm 0.0002$} & 1.0383 \textit{$\pm 0.0012$} \\
        \\[-0.75em]

        \hline
        \hline
        \\[-0.75em]
        MAE & EvolveGCNH & 175.30 \textit{$\pm 6.1418$} & 19.853 \textit{$\pm 0.5108$} \\
        \\[-1.00em]
         & GCLSTM & 165.74 \textit{$\pm 1.9623$} & 17.070 \textit{$\pm 0.2012$} \\
        \\[-1.00em]
         & GConvLSTM & 167.12 \textit{$\pm 2.8474$} & 16.823 \textit{$\pm 0.1805$} \\
        \\[-1.00em]
         & MPNNLSTM & 149.06 \textit{$\pm 0.9669$} & 18.571 \textit{$\pm 0.1846$} \\
        \\[-1.00em]
         & LSTM$_{mse}$ & \underline{140.34} \textit{$\pm 0.2430$} & 16.452 \textit{$\pm 0.0195$}  \\
        \\[-1.00em]
        & GCN$_{mse}$ & 158.87 \textit{$\pm 0.0030$} & 18.893 \textit{$\pm 0.0005$}  \\
        \\[-1.00em]
        \hline
        \\[-0.75em]
         & $\mathbf{PrGLSTM_{rmsle}}$ & 157.55 \textit{$\pm 1.3,218$} & 18.583 \textit{$\pm 0.0448$} \\
        \\[-1.00em]
         & PrGLSTM$_{mse}$ & 164.21 \textit{$\pm 0.3344$} & 18.961 \textit{$\pm 0.0265$} \\
        \\[-1.00em]
        & PrLSTM$_{mse}$ & 144.74 \textit{$\pm 0.4254$} & 16.502 \textit{$\pm 0.0325$} \\
        \\[-1.00em]
        & PrLSTM$_{rmsle}$ & 143.69 \textit{$\pm 0.4093$} & \underline{16.128} \textit{$\pm 0.0166$} \\
        \\[-1.00em]
        & PrGCN$_{mse}$ & 161.81 \textit{$\pm 0.2485$} & 19.168 \textit{$\pm 0.0308$} \\
        \\[-1.00em]
        & $\mathbf{PrGCN_{rmsle}}$ & 158.24 \textit{$\pm 0.0506$} & 18.456 \textit{$\pm 0.0121$} \\
        \\[-0.75em]
        
        \hline
        \hline
        \\[-0.75em]
        Huber & EvolveGCNH & 235.78 \textit{$\pm 8.2904$} & 25.907 \textit{$\pm 0.6891$} \\
        \\[-1.00em]
         & GCLSTM & 222.87 \textit{$\pm 2.6486$} & 22.164 \textit{$\pm 0.2713$} \\
        \\[-1.00em]
         & GConvLSTM & 224.74 \textit{$\pm 3.8435$} & 21.832 \textit{$\pm 0.2435$} \\
        \\[-1.00em]
         & MPNNLSTM & 200.37 \textit{$\pm 1.3053$} & 24.181 \textit{$\pm 0.2490$} \\
        \\[-1.00em]
         & LSTM$_{mse}$ & \underline{188.60} \textit{$\pm 0.3280$} & 21.328 \textit{$\pm 0.0262$}  \\
        \\[-1.00em]
        & GCN$_{mse}$ & 213.59 \textit{$\pm 0.0040$} & 24.613 \textit{$\pm 0.0007$}  \\
        \\[-1.00em]
        \hline
        \\[-0.75em]
         & $\mathbf{PrGLSTM_{rmsle}}$ & 211.83 \textit{$\pm 1.7843$} & 24.204 \textit{$\pm 0.0598$} \\
        \\[-1.00em]
         & PrGLSTM$_{mse}$ & 220.81 \textit{$\pm 0.4513$} & 24.711 \textit{$\pm 0.0355$} \\
        \\[-1.00em]
        & PrLSTM$_{mse}$ & 194.54 \textit{$\pm 0.5742$} & 21.392 \textit{$\pm 0.0434$} \\
        \\[-1.00em]
        & PrLSTM$_{rmsle}$ & 193.12 \textit{$\pm 0.5521$} & \underline{20.896} \textit{$\pm 0.0219$} \\
        \\[-1.00em]
        & PrGCN$_{mse}$ & 217.57 \textit{$\pm 0.3354$} & 24.986 \textit{$\pm 0.0417$} \\
        \\[-1.00em]
        & $\mathbf{PrGCN_{rmsle}}$ & 212.76 \textit{$\pm 0.0682$} & 24.029 \textit{$\pm 0.0161$} \\
        \\[-0.75em]
        
        \hline
        \hline
        
        % \bottomrule
    \end{tabular}%
    }

    \footnotesize
    {
    \parbox{1\columnwidth}{
    \textit{Note:} EvolveGCNH \cite{par20}, GCLSTM \cite{youngjoo18}, GConvLSTM \cite{chen2022gc}, and MPNNLSTM \cite{pan21} are benchmark models.
    }
    }
    %}
\end{table}

\begin{table}[h!]
    \caption{\label{tab:benchmark_mse} Global estimation of performance metrics (MSE and RMSE) with 95\% variability intervals for benchmark and ablation models.}
    \small
    \centering
    %\resizebox{\columnwidth}{!}{%
    {\fontsize{8pt}{10pt}\selectfont
    \begin{tabular}{llcr}
        %\toprule
        \\[-1.25em]
        \multicolumn{1}{l}{Metric} & \multicolumn{1}{l}{Method} & \multicolumn{1}{c}{COVID-19} & \multicolumn{1}{c}{chickenpox}\\ 
        
        \\[-1.00em]
        
        \hline
        \hline
        \\[-0.75em]
        MSE & EvolveGCNH & 17139 \textit{$\pm 1485.1$} & 9.2984 \textit{$\pm 0.2964$} \\
        \\[-1.00em]
        ($x10^2$) & GCLSTM & 15055 \textit{$\pm 464.59$} & 7.9855 \textit{$\pm 0.1964$} \\
        \\[-1.00em]
         & GConvLSTM & 14846 \textit{$\pm 521.06$} & 7.8199 \textit{$\pm 0.1819$} \\
        \\[-1.00em]
         & MPNNLSTM & 14339 \textit{$\pm 308.14$} & 9.5696 \textit{$\pm 0.2132$} \\
        \\[-1.00em]
         & LSTM$_{mse}$ & \underline{10753} \textit{$\pm 47.710$} & 7.4318 \textit{$\pm 0.0066$}  \\
        \\[-1.00em]
        & GCN$_{mse}$ & 13086 \textit{$\pm 0.2919$} & 8.7667 \textit{$\pm 0.0004$}  \\
        \\[-1.00em]
        \hline
        \\[-0.75em]
        & $\mathbf{PrGLSTM_{rmsle}}$ & 14575 \textit{$\pm 382.14$} & 9.0035 \textit{$\pm 0.0103$} \\
        \\[-1.00em]
         & PrGLSTM$_{mse}$ & 13633 \textit{$\pm 50.527$} & 8.8688 \textit{$\pm 0.0117$} \\
        \\[-1.00em]
        & PrLSTM$_{mse}$ & 11540 \textit{$\pm 62.317$} & \underline{7.3791} \textit{$\pm 0.0057$} \\
        \\[-1.00em]
        & PrLSTM$_{rmsle}$ & 12351 \textit{$\pm 92.855$} & 7.5784 \textit{$\pm 0.0088$} \\
        \\[-1.00em]
        & PrGCN$_{mse}$ & 13271 \textit{$\pm 21.937$} & 8.8991 \textit{$\pm 0.0147$} \\
        \\[-1.00em]
        & $\mathbf{PrGCN_{rmsle}}$ & 12922 \textit{$\pm 13.108$} & 9.1467 \textit{$\pm 0.0066$} \\
        \\[-0.75em]
        
        \hline
        \hline

        \\[-0.75em]
        RMSE & EvolveGCNH & 1285.0 \textit{$\pm 48.741$} & 30.410 \textit{$\pm 0.4395$} \\
        \\[-1.00em]
         & GCLSTM & 1222.9 \textit{$\pm 19.513$} & 28.202 \textit{$\pm 0.3462$} \\
        \\[-1.00em]
         & GConvLSTM & 1213.3 \textit{$\pm 21.861$} & 27.914 \textit{$\pm 0.3234$} \\
        \\[-1.00em]
         & MPNNLSTM & 1195.7 \textit{$\pm 12.653$} & 30.888 \textit{$\pm 0.3287$} \\
        \\[-1.00em]
         & LSTM$_{mse}$ & \underline{1036.9} \textit{$\pm 2.2992$} & 27.261 \textit{$\pm 0.0121$}  \\
        \\[-1.00em]
        & GCN$_{mse}$ & 1143.9 \textit{$\pm 0.0127$} & 29.608 \textit{$\pm 0.0006$}  \\
        \\[-1.00em]
        \hline
        \\[-0.75em]
        & $\mathbf{PrGLSTM_{rmsle}}$ & 1204.6 \textit{$\pm 15.486$} & 30.005 \textit{$\pm 0.0172$} \\
        \\[-1.00em]
         & PrGLSTM$_{mse}$ & 1167.5 \textit{$\pm 2.1673$} & 29.780 \textit{$\pm 0.0196$} \\
        \\[-1.00em]
        & PrLSTM$_{mse}$ & 1074.1 \textit{$\pm 2.8887$} & \underline{27.164} \textit{$\pm 0.0105$} \\
        \\[-1.00em]
        & PrLSTM$_{rmsle}$ & 1111.1 \textit{$\pm 4.1754$} & 27.528 \textit{$\pm 0.0160$} \\
        \\[-1.00em]
        & PrGCN$_{mse}$ & 1152.0 \textit{$\pm 0.9506$} & 29.831 \textit{$\pm 0.0246$} \\
        \\[-1.00em]
        & $\mathbf{PrGCN_{rmsle}}$ & 1136.7 \textit{$\pm 0.5765$} & 30.243 \textit{$\pm 0.0109$} \\
        \\[-0.75em]
        
        \hline
        \hline

    \end{tabular}%
    }
    %}

\end{table}

\begin{table}%[hb!]
    \caption{\label{tab4:overhead} Overhead time during model training for the proposed and baseline models; values are in minutes per 800 epochs (values in parentheses are the number of iterations per second).}
    \small
    \centering    
    %\resizebox{\columnwidth}{!}{%
    {\fontsize{8pt}{10pt}\selectfont
    \begin{tabular}{lrr}
        \\[-1.25em]
        & \multicolumn{2}{c}{$Minutes/800 \,epochs$}\\
        \multicolumn{1}{l}{Method} & \multicolumn{2}{c}{$(iterations/sec)$}\\
        & \multicolumn{1}{c}{COVID-19} & \multicolumn{1}{c}{chickenpox}\\
        \\[-1.00em]
        
        \hline
        \\[-0.75em]
        EvolveGCNH & 5:09 (\textit{$02.58$}) & 8:24 (\textit{$01.59$}) \\
        \\[-1.00em]
        GCLSTM & 3:36 (\textit{$03.70$}) & 10:35 (\textit{$01.26$}) \\
        \\[-1.00em]
        GConvLSTM & 5:32 (\textit{$02.41$}) & 9:59 (\textit{$01.33$}) \\
        \\[-1.00em]
        MPNNLSTM & 3:52 (\textit{$03.44$}) & 10:12 (\textit{$01.31$}) \\
        \\[-1.00em]
        LSTM$_{mse}$ & 2:38 (\textit{$05.08$}) & 6:18 (\textit{$02.12$}) \\
        \\[-1.00em]
        GCN$_{mse}$ & 1:20 (\textit{$10.04$}) & 1:18 (\textit{$10.25$}) \\
        \\[-1.00em]
        $\mathbf{PrGLSTM_{rmsle}}$ & 3:40 (\textit{$03.63$}) & 9:56 (\textit{$01.34$}) \\
        \\[-1.00em]
        \hline
        \\[-0.75em]
        PrGLSTM$_{mse}$ & 3:39 (\textit{$03.65$}) & 9:36 (\textit{$01.39$}) \\
        \\[-1.00em]
        PrLSTM$_{mse}$ & 2:17 (\textit{$05.86$}) & 5:01 (\textit{$02.65$}) \\
        \\[-1.00em]
        PrLSTM$_{rmsle}$ & 2:14 (\textit{$06.00$}) & 5:16 (\textit{$02.52$}) \\
        \\[-1.00em]
        PrGCN$_{mse}$ & 1:55 (\textit{$06.93$}) & 2:06 (\textit{$06.32$}) \\
        \\[-1.00em]        
        $\mathbf{PrGCN_{rmsle}}$ & 1:30 (\textit{$08.85$}) & 2:11 (\textit{$06.08$}) \\
        \\[-0.75em]

        \hline
    \end{tabular}%
    }
\end{table}

\subsection{Additional results}
\label{app4:pred}

Table \ref{tab4:overhead} shows the overhead time for the proposed models and baselines, where all model were training under similar configuration, as indicated in Section \ref{sub4:43conf}.

\begin{figure}%[!h]
\centering

\begin{minipage}{0.43\textwidth}
\raggedleft

\includegraphics[width=0.97\linewidth]{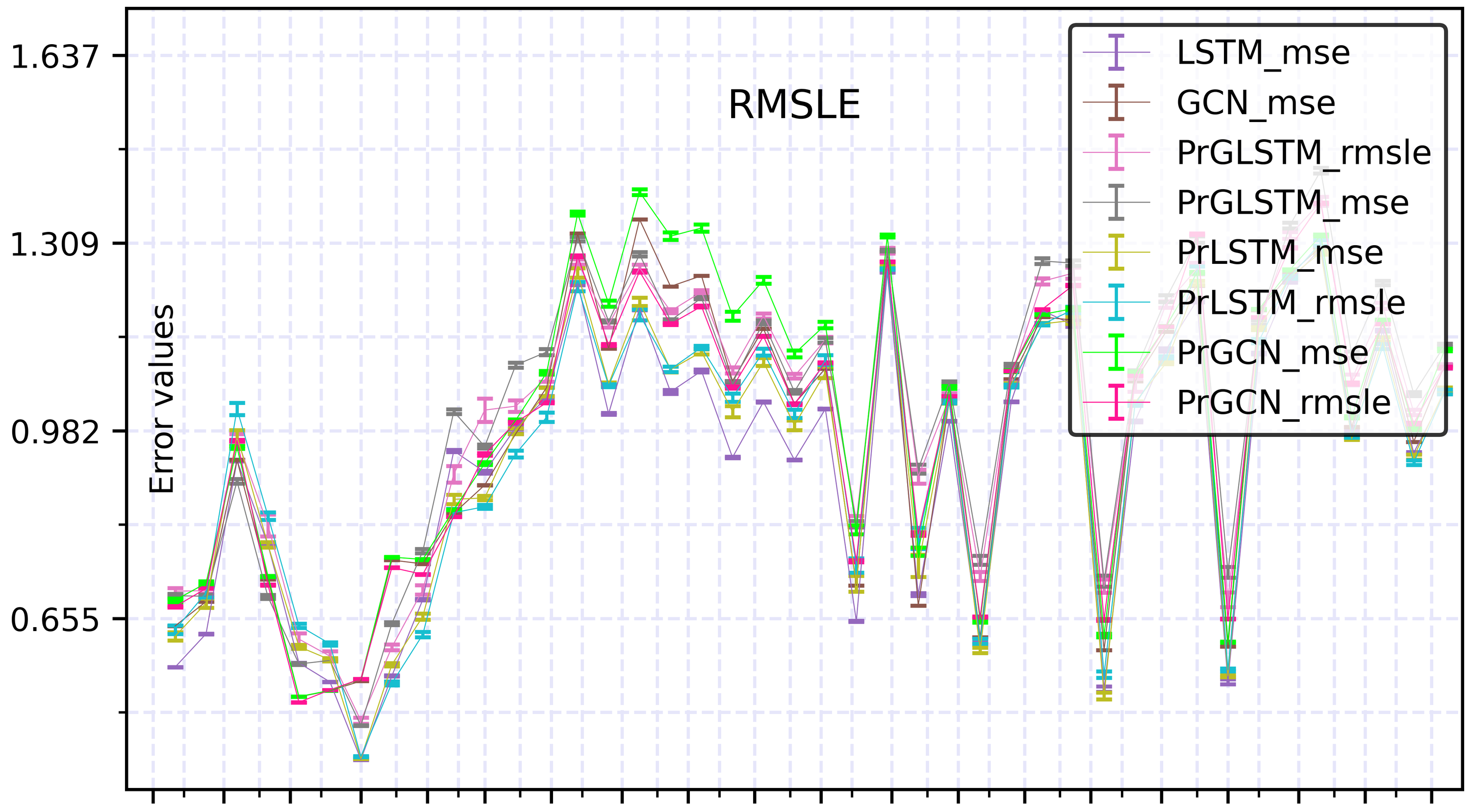}\par
\includegraphics[width=0.95\linewidth]{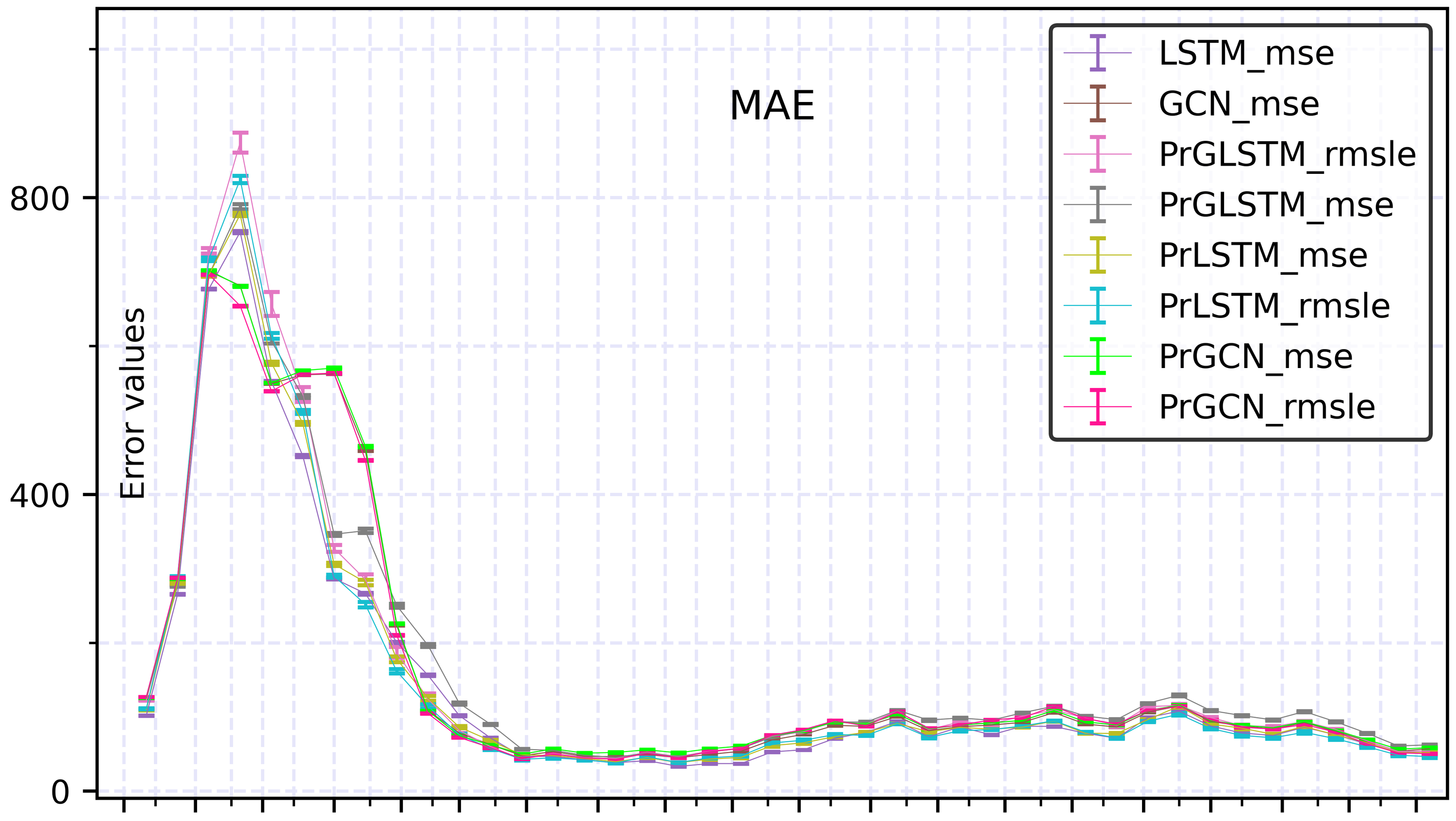}\par
\includegraphics[width=0.95\linewidth]{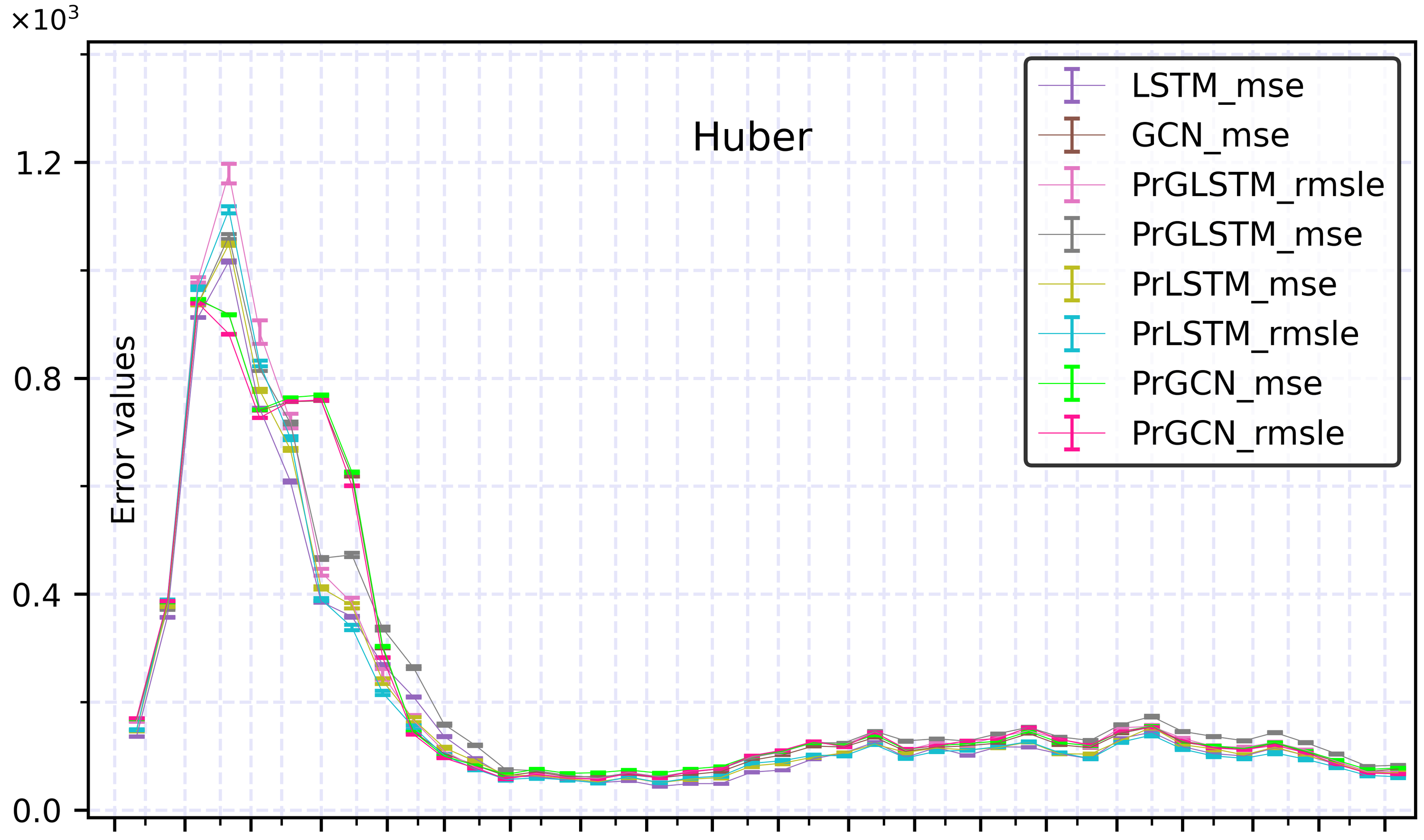}\par
\includegraphics[width=0.95\linewidth]{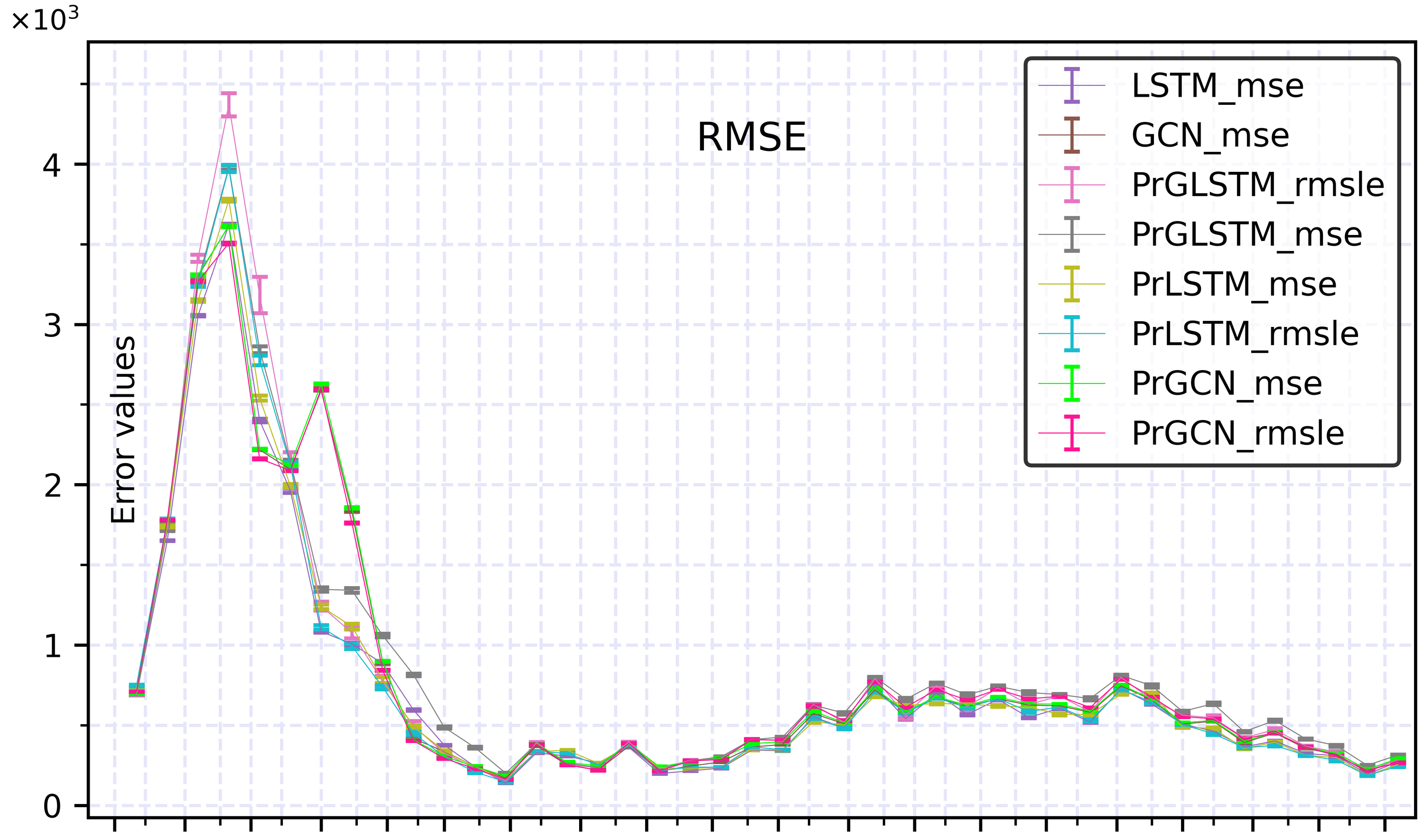}\par
\includegraphics[width=0.95\linewidth]{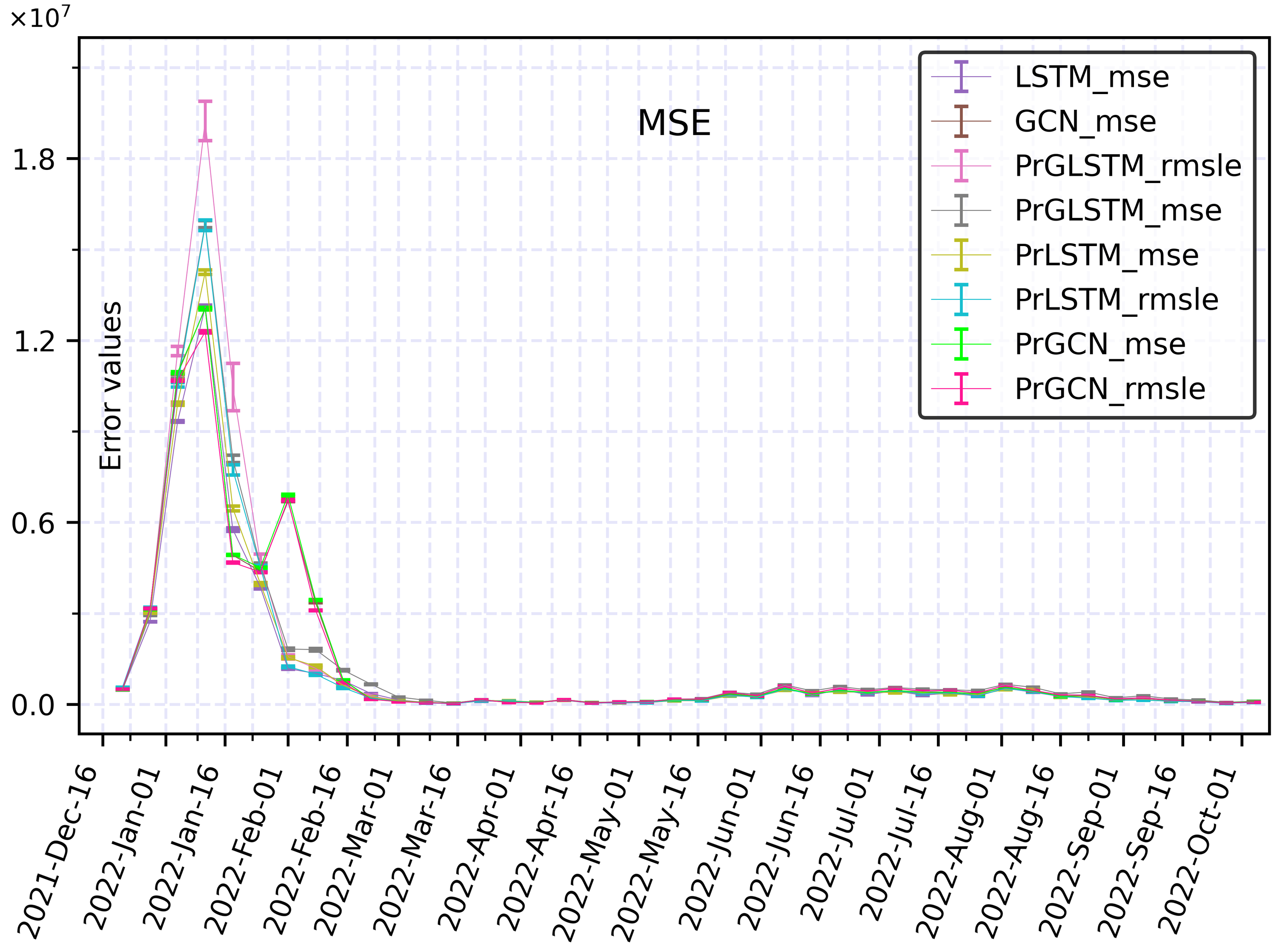}\par

\caption{Weekly evolution of global performance metrics across the proposed and ablation models for the COVID-19 predictions in the US (3,218 counties).}
\label{fig4:errorablat_us}
\end{minipage}
\hfill
\begin{minipage}{0.43\textwidth}
\raggedleft

\includegraphics[width=0.973\textwidth]{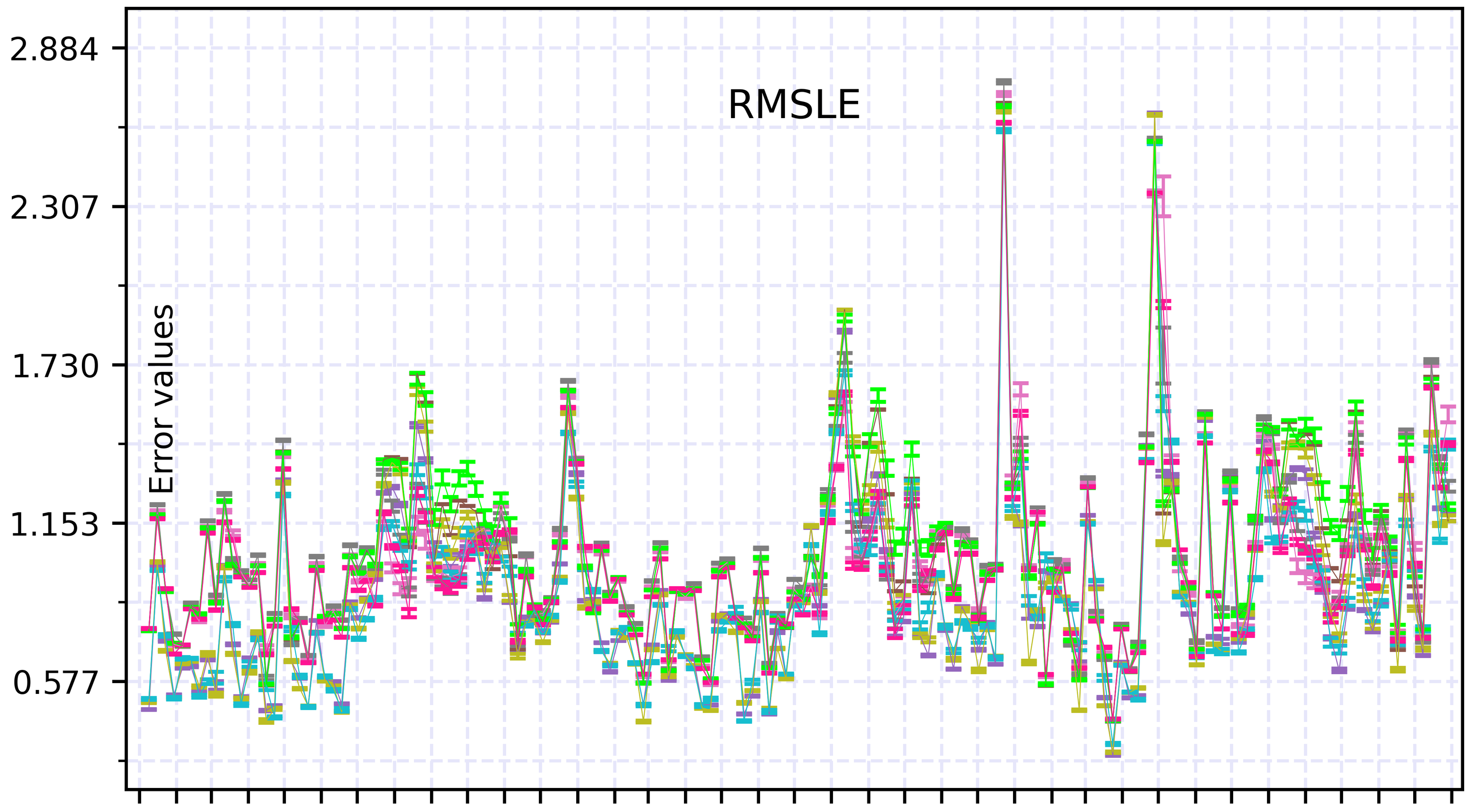}\par
\includegraphics[width=0.95\textwidth]{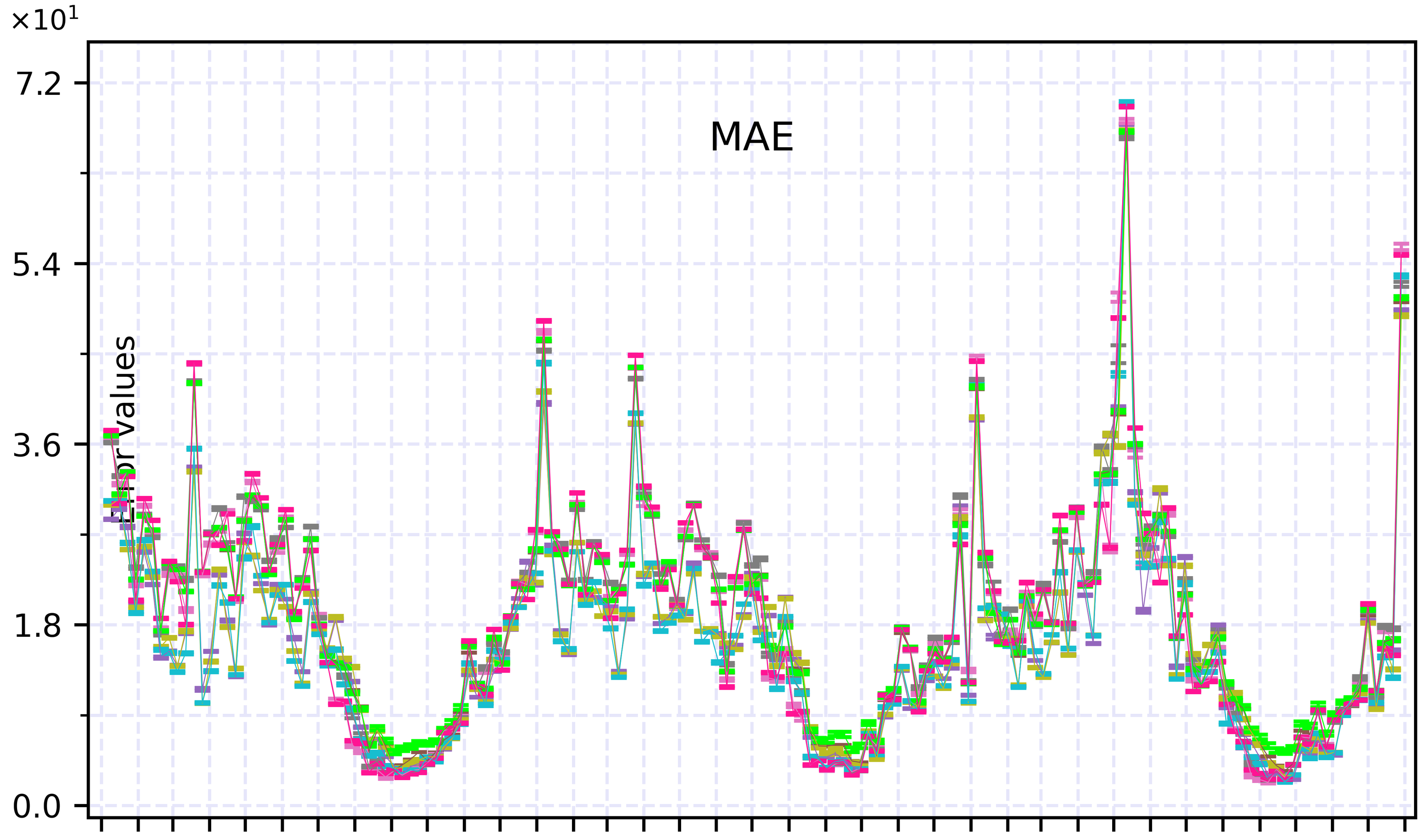}\par
\includegraphics[width=0.95\textwidth]{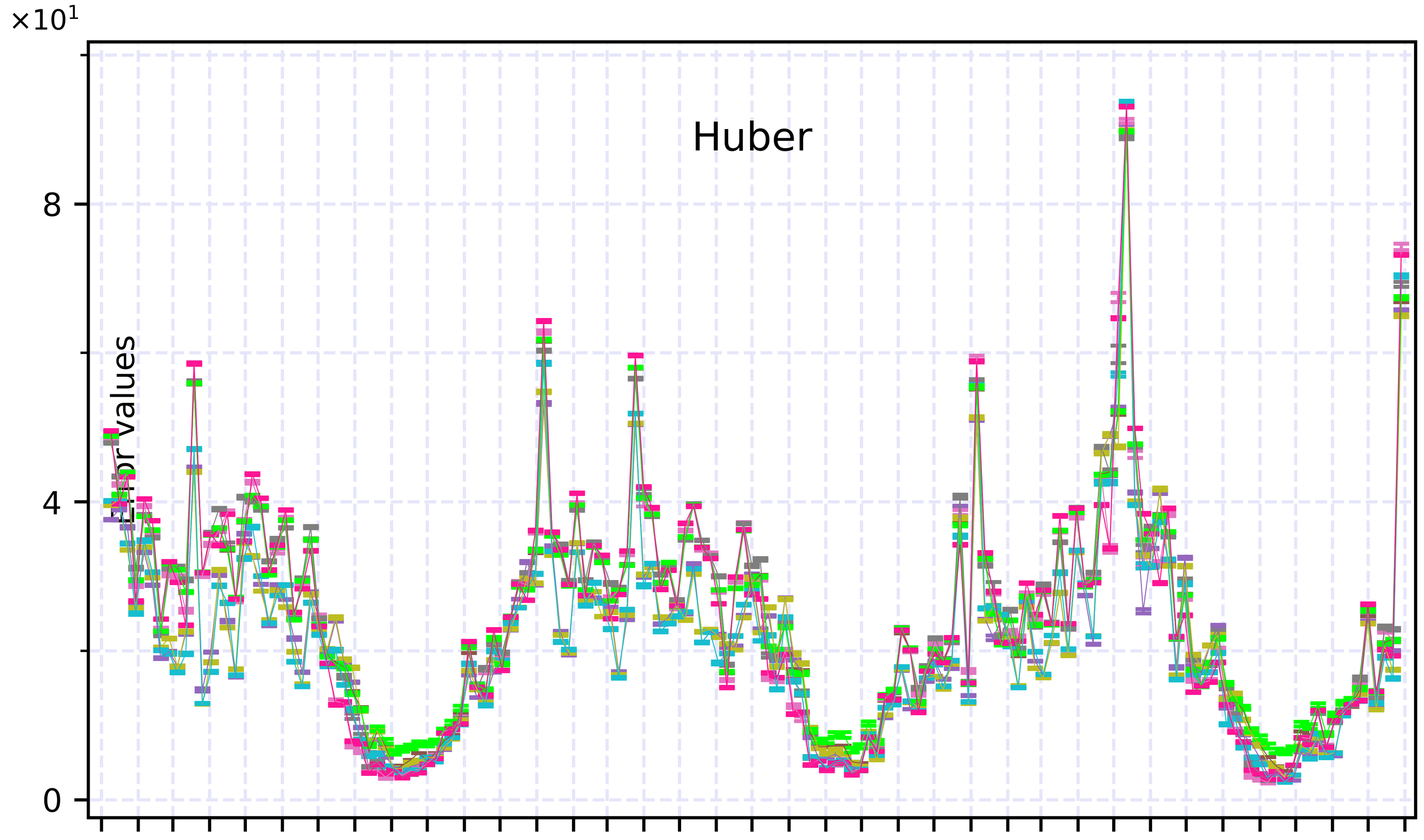}\par
\includegraphics[width=0.95\textwidth]{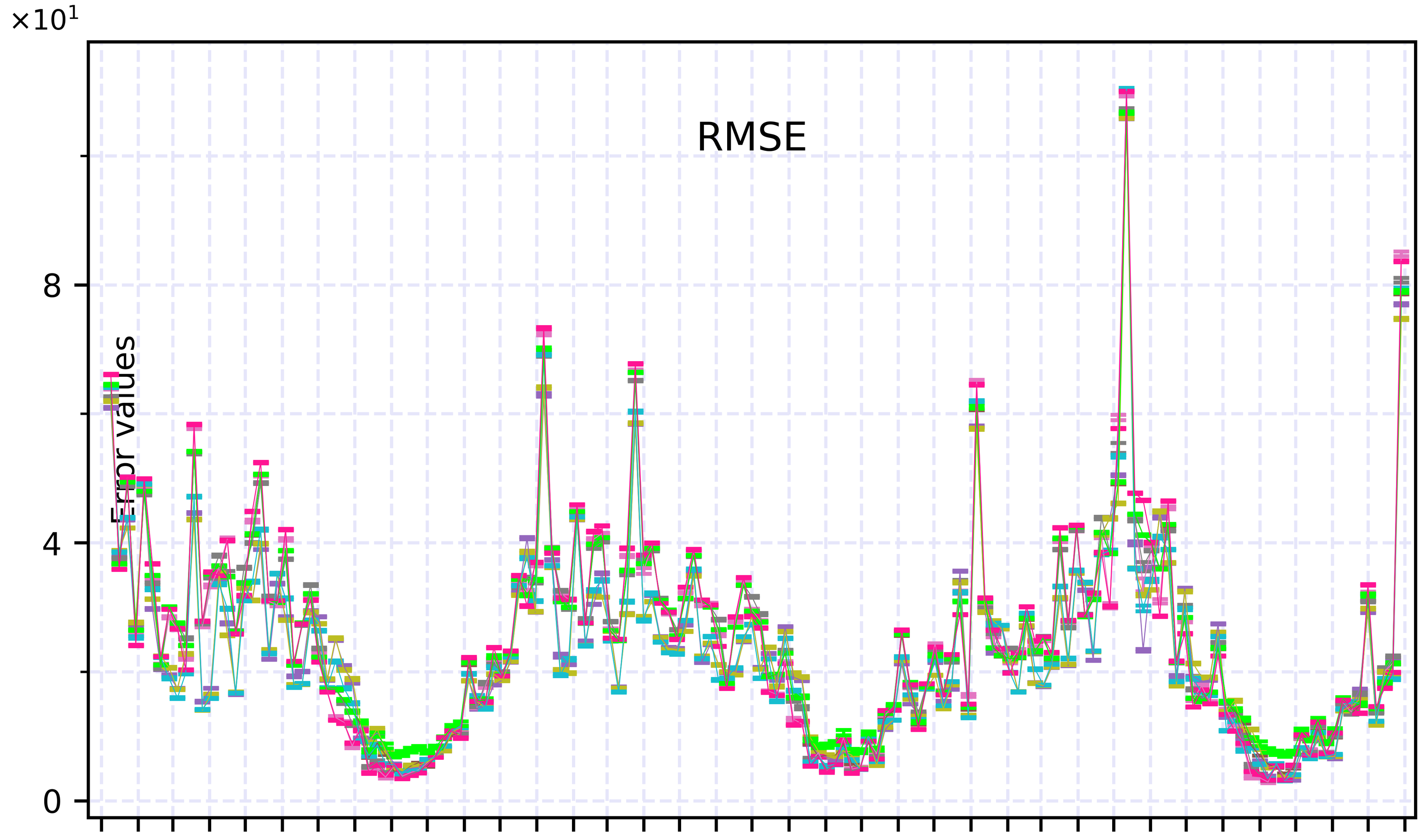}\par
\includegraphics[width=0.95\textwidth]{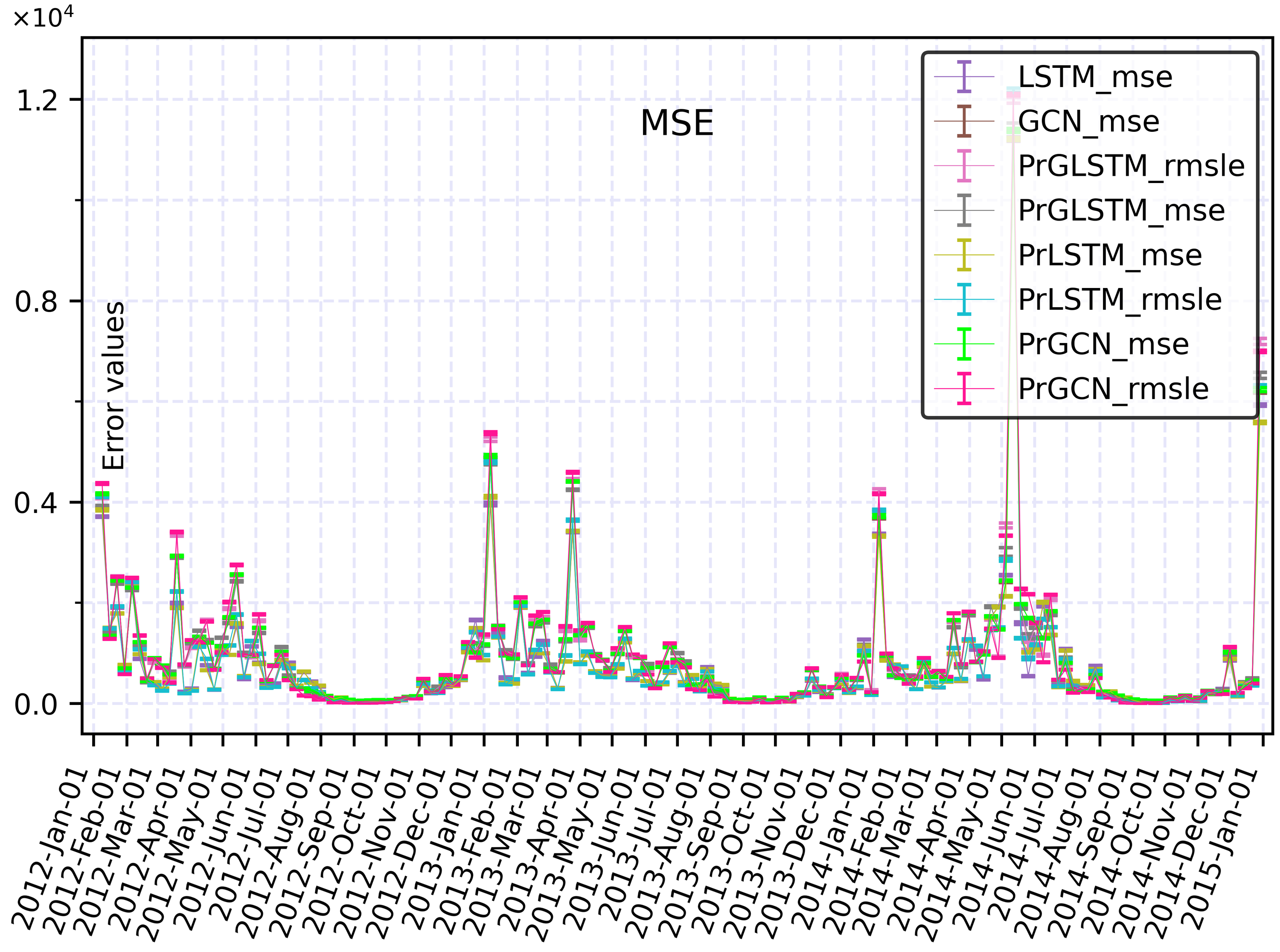}\par

\caption{
Weekly evolution of global performance metrics across the proposed and ablation models for chickenpox predictions in Hungary (20 counties).
}
\label{fig4:errorablat_hun}
\end{minipage}
\end{figure}

Figures \ref{fig4:errorablat_us} and \ref{fig4:errorablat_hun} illustrate the global metric values across temporal steps during the testing phase for the prediction of COVID-19 and chickenpox, respectively. The values are given for the proposed models and their ablations.

\begin{figure}%[!h]
\centering

\begin{minipage}{1\textwidth}
\raggedleft

\includegraphics[width=1\textwidth]{320_Benchmark_trend_n320_240304_1209_labXFalse_legTrue.png}
\includegraphics[width=1\textwidth]{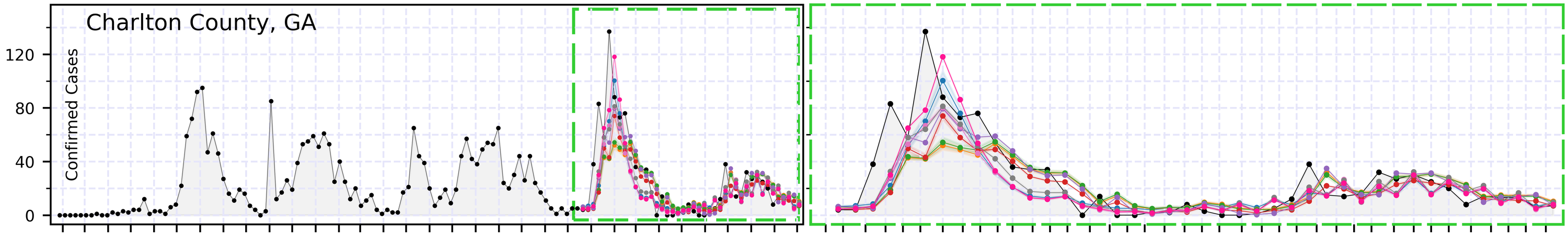}
\includegraphics[width=1\textwidth]{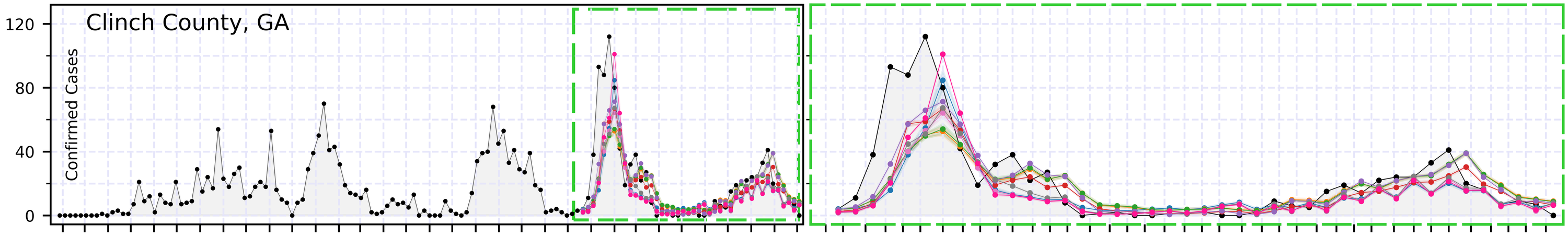}
\includegraphics[width=1\textwidth]{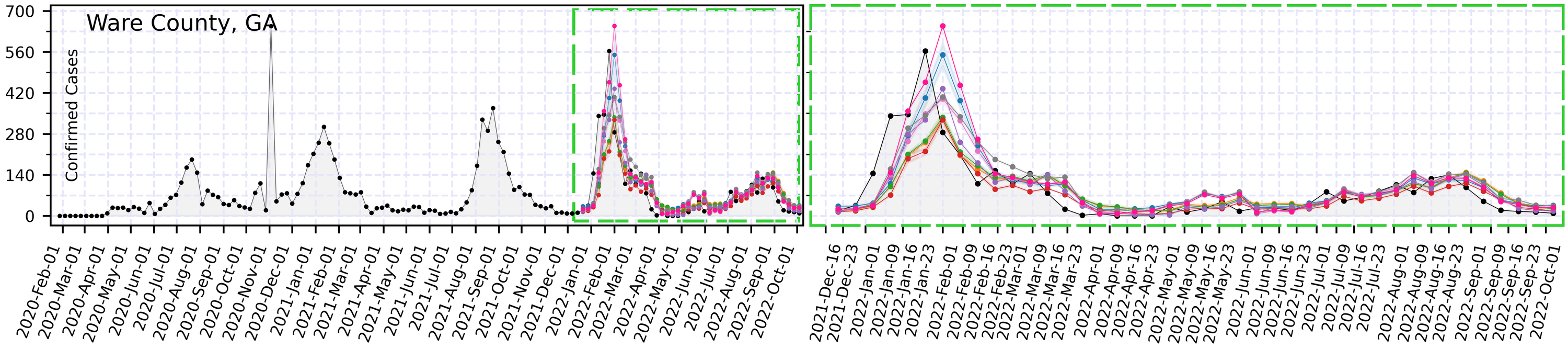}

\end{minipage}

\caption{Observed time-series data and model predictions for COVID-19 in Baker county (FL) and neighbouring counties in Georgia state (GA), US.}

\label{fig4:bakeraround_2}
\end{figure}

Figure \ref{fig4:bakeraround_2} presents the observed time-series incidence data and predictions for COVID-19 in Baker county (FL) and neighbouring counties in Georgia state (GA), US. We can observe the different patterns in the training data across the neighbourhood that were part of the model aggregation process to predict the incidence trajectories of Baker county. The prediction values are given for the proposed models and their benchmarks.

\begin{figure}%[!h]
\centering

\begin{minipage}{1\textwidth}
\raggedleft

\includegraphics[width=0.994\textwidth]{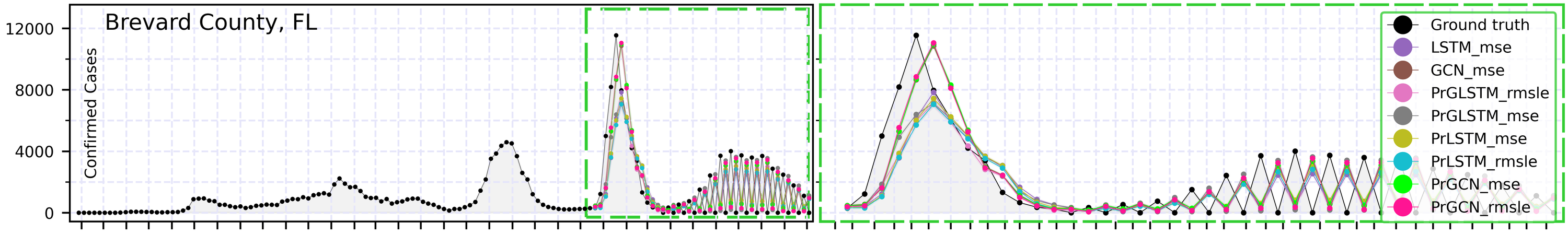}
\includegraphics[width=1\textwidth]{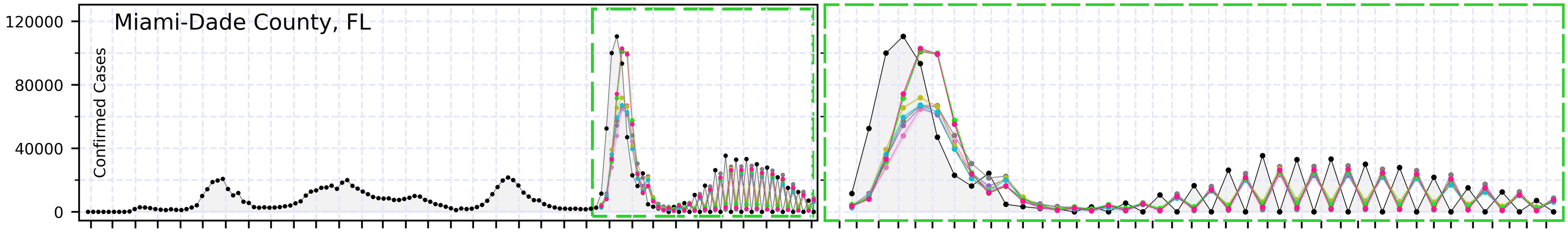}
\includegraphics[width=0.994\textwidth]{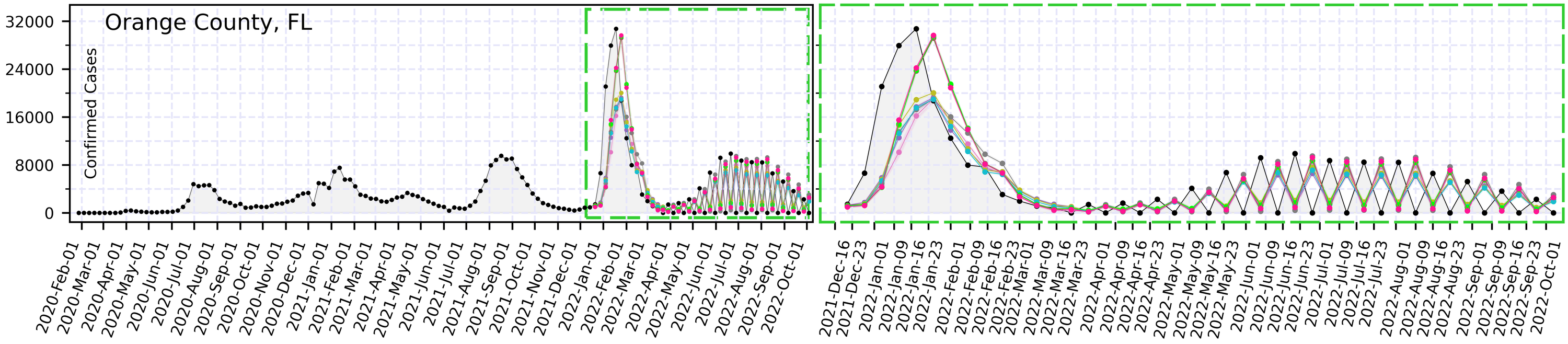}

\end{minipage}

\caption{
Observed time-series data and model predictions for COVID-19 in FL counties using the proposed and ablation models during a fluctuating testing period.
}
\label{fig4:trend_FLc}
\end{figure}

\begin{figure}%[!h]
\centering

\begin{minipage}{1\textwidth}
\raggedleft

\includegraphics[width=1\textwidth]{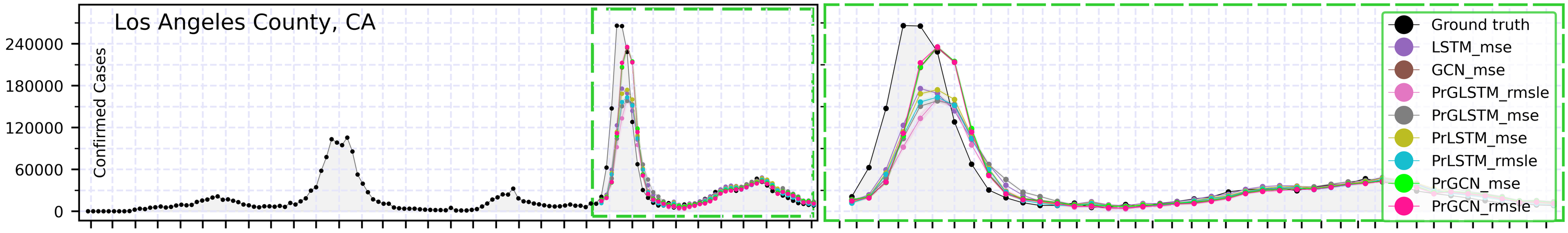}
\includegraphics[width=1\textwidth]{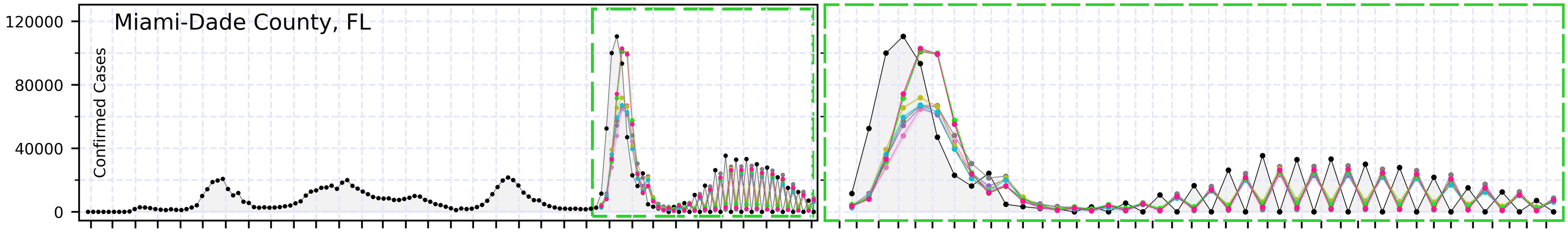}
\includegraphics[width=1\textwidth]{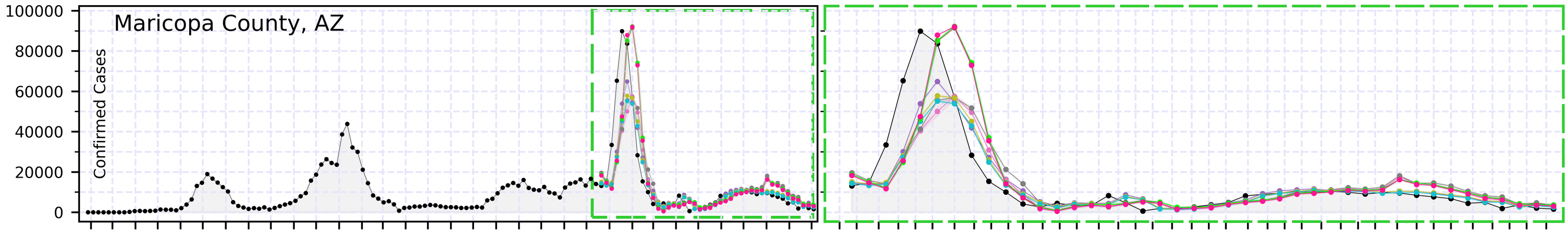}
\includegraphics[width=1\textwidth]
{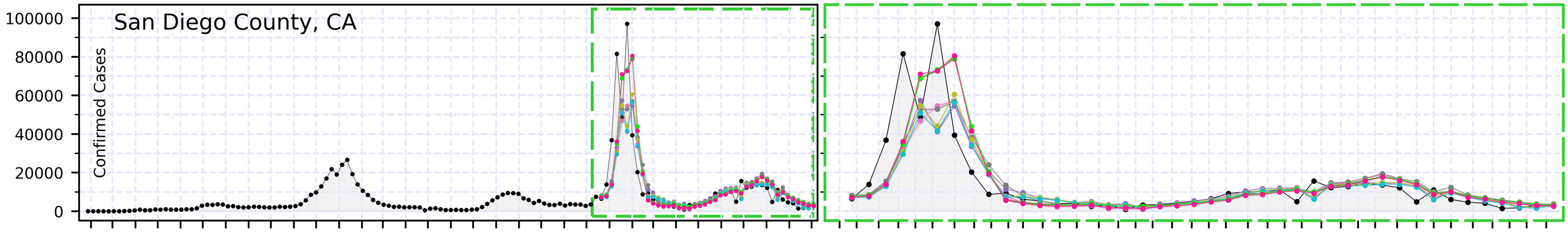}
\includegraphics[width=0.994\textwidth]
{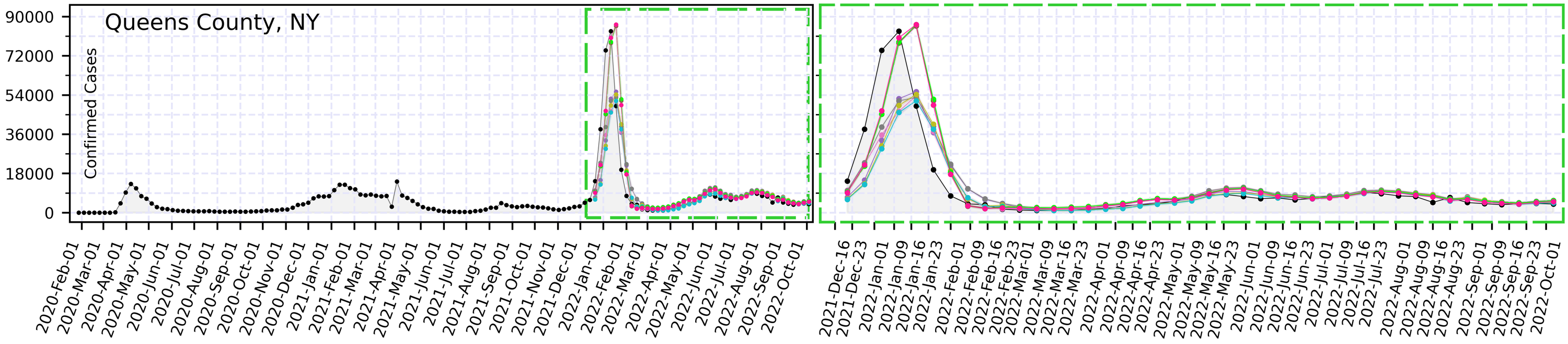}

\end{minipage}

\caption{
Observed time-series data and model predictions for COVID-19 in US counties with the largest weekly incidence using the proposed and ablation models.
}
\label{fig4:trend_best_us_ablation}
\end{figure}

\begin{figure}%[!h]
\centering

\begin{minipage}{1\textwidth}
\raggedright

\includegraphics[width=0.99\textwidth]{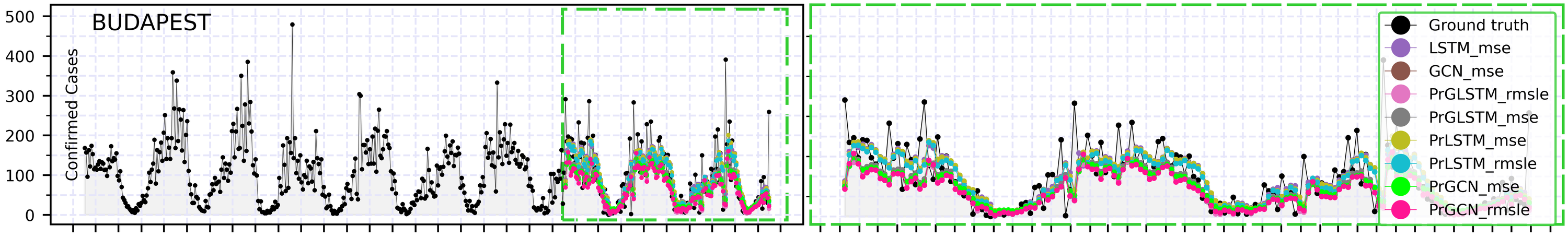}
\includegraphics[width=0.99\textwidth]{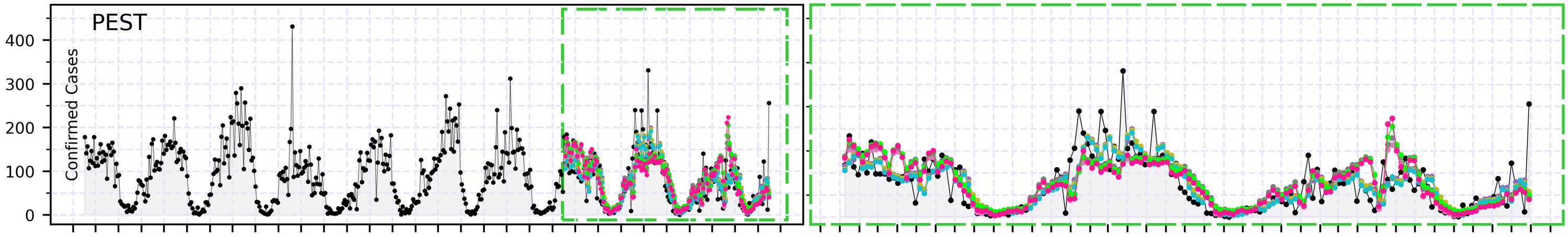}
\includegraphics[width=0.99\textwidth]{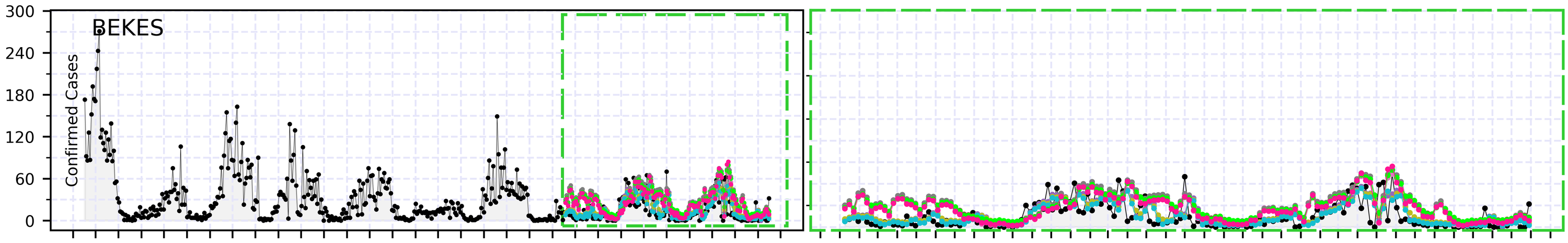}
\includegraphics[width=0.992\textwidth]{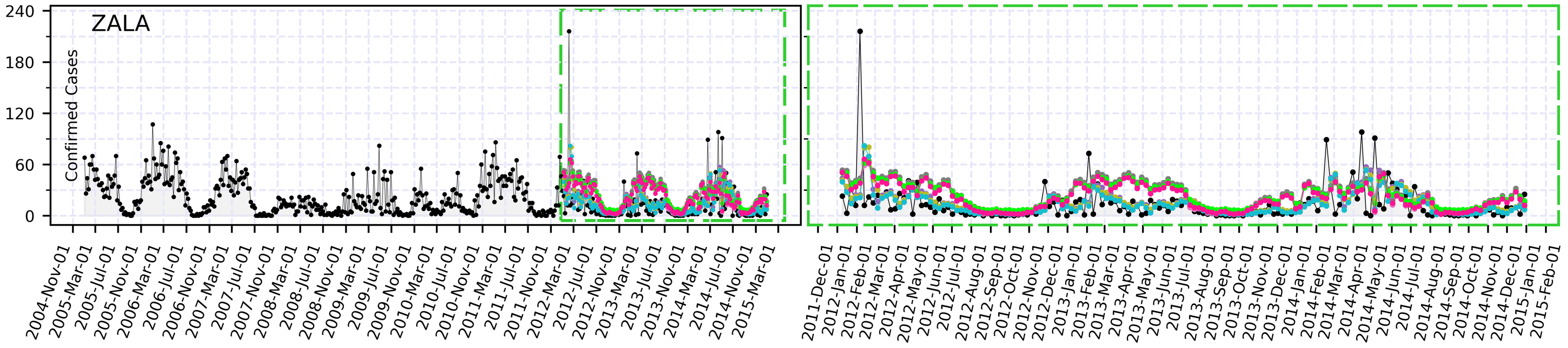}

\end{minipage}

\caption{
Observed time-series data and model predictions for chickenpox in Hungarian counties with the largest weekly incidence using the proposed and ablation models.
}
\label{fig4:trend_best_hun_ablation}
\end{figure}

\begin{figure}%[!h]
\centering
\begin{minipage}{1\textwidth}
\raggedleft

\includegraphics[width=1\textwidth]{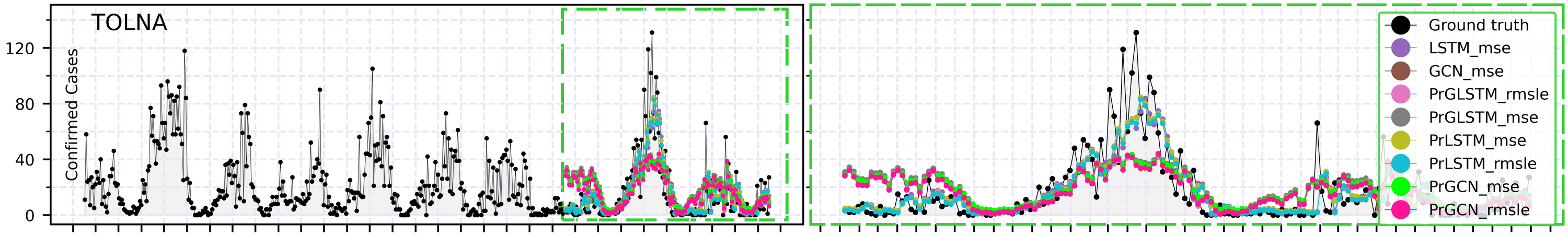}
\includegraphics[width=1\textwidth]{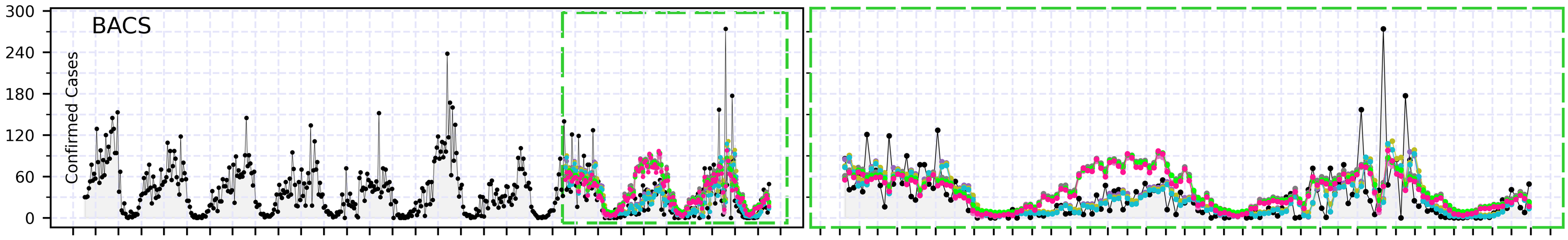}
\includegraphics[width=1\textwidth]{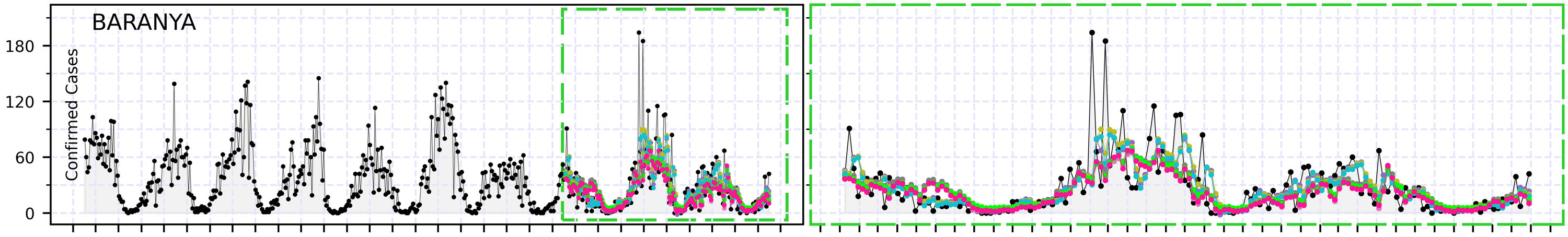}
\includegraphics[width=1\textwidth]{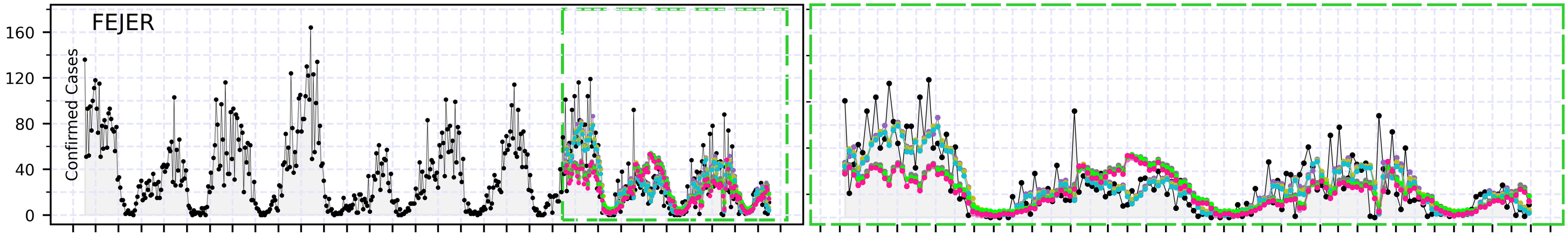}
\includegraphics[width=1\textwidth]{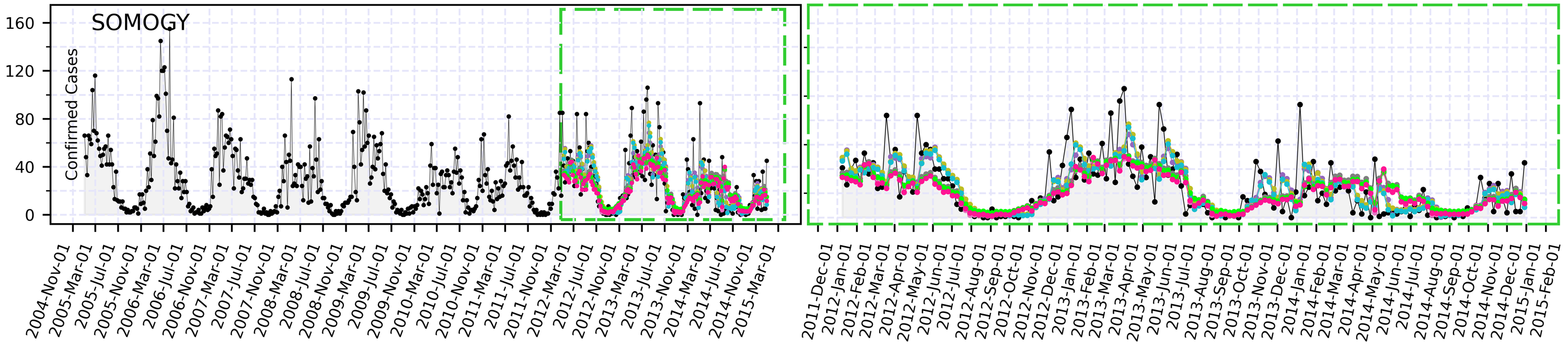}

\end{minipage}

\caption{
Observed time-series data and model predictions for chickenpox in Tolna and neighbouring counties, Hungary, using the proposed and ablation models.
}

\label{fig4:tolnaaround_ablation}
\end{figure}

\begin{figure}%[!h]
\centering

\begin{minipage}{1\textwidth}
\raggedleft

\includegraphics[width=1\textwidth]{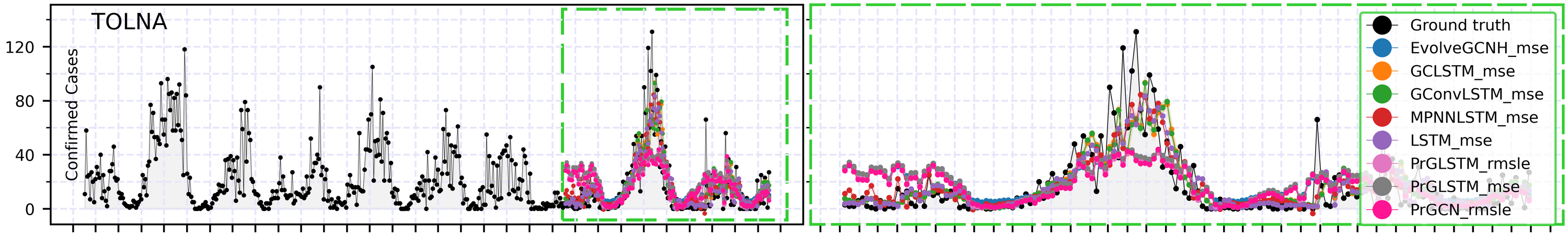}
\includegraphics[width=1\textwidth]{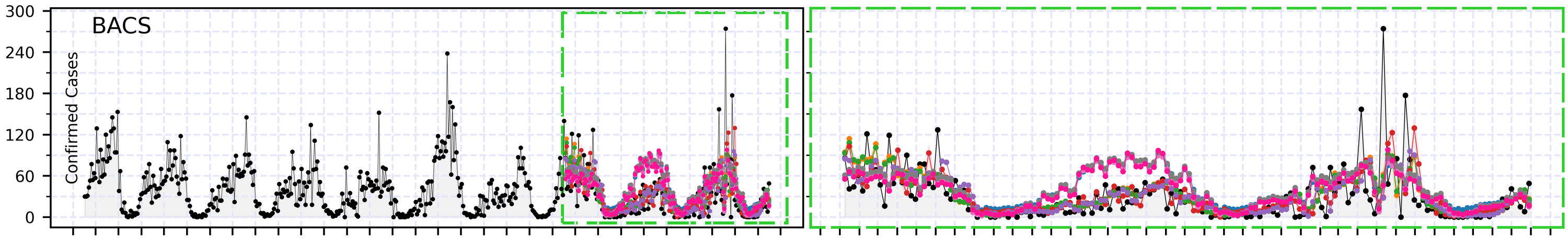}
\includegraphics[width=1\textwidth]{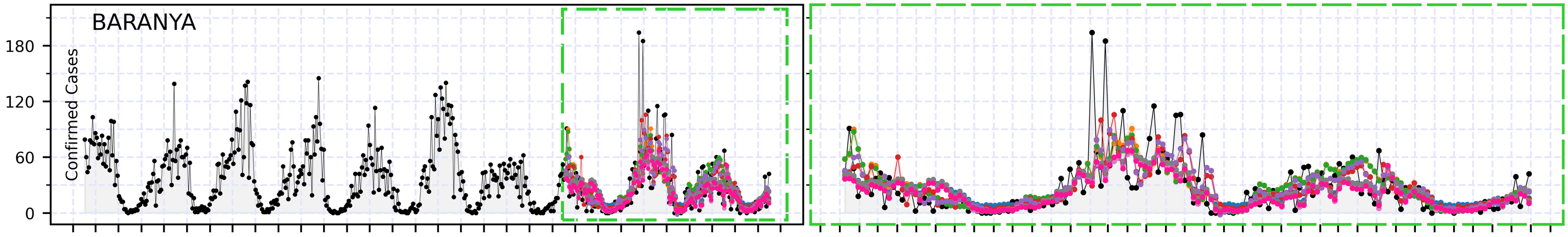}
\includegraphics[width=1\textwidth]{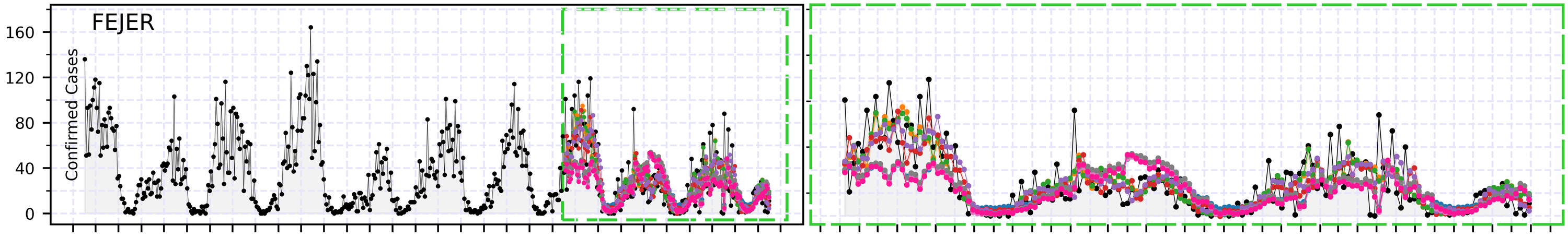}
\includegraphics[width=1\textwidth]{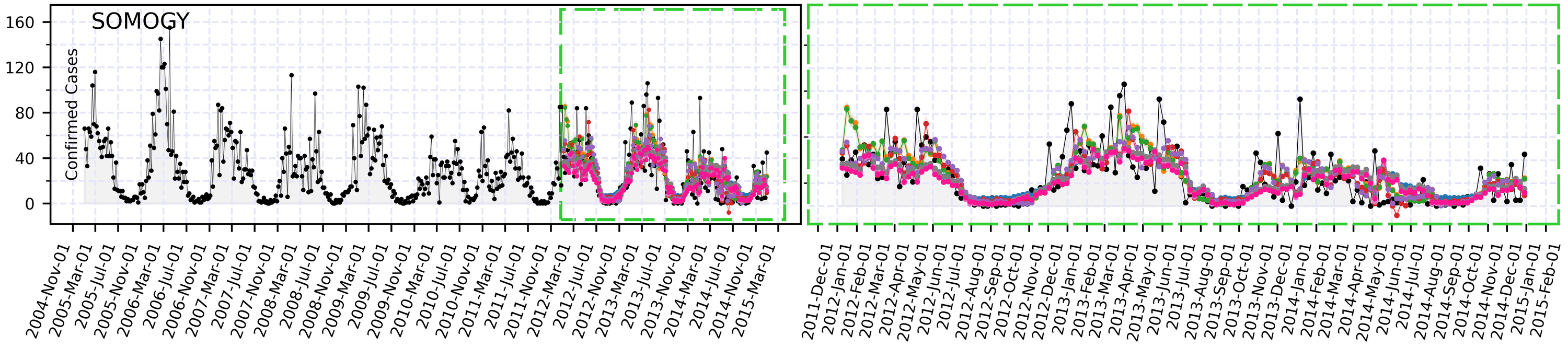}

\end{minipage}

\caption{
Observed time-series data and model predictions for chickenpox in Tolna and neighbouring counties, Hungary, using the proposed and benchmark models.
}

\label{fig4:tolnaaround_bench}
\end{figure}

Figures \ref{fig4:trend_FLc} and \ref{fig4:trend_best_us_ablation} display the prediction results for COVID-19 in the US using the proposed and ablation models, while Figures \ref{fig4:trend_best_hun_ablation}, \ref{fig4:tolnaaround_ablation} and \ref{fig4:tolnaaround_bench} visualise the predictions for chickenpox in Hungary.

\clearpage

\newpage

\section{Supplementary related work}\label{apd4:prework}

Table \ref{tab4:modelcomparison} summarises the features of graph-based learning models, including structure details and error metrics used to evaluate model performance.

\begin{table}[hb!]
    \caption{\label{tab4:modelcomparison} Prior work in graph-based learning models for infectious disease applications and the proposed version PrGLSTM.}
    \small
    \centering
    
    %\resizebox{\columnwidth}{!}{%
    %\resizebox{\textwidth}{!}{%
    {\fontsize{6pt}{7pt}\selectfont
    \begin{tabular}{lllllllll}
        %\toprule
        \\%[-1.25em]
        \multicolumn{1}{l}{Model} & \multicolumn{1}{l}{Architecture} & \multicolumn{1}{l}{ Parameters} & \multicolumn{1}{l}{Advantages} & \multicolumn{1}{l}{Limitations} & \multicolumn{1}{l}{Metrics} & \multicolumn{1}{l}{Datasets}\\ 
        
        \\[-1.00em]
        \hline
        \\[-0.25em]
        \begin{tabular}{@{}l@{}}MPNNLSTM \\ \cite{pan21} \end{tabular} & \begin{tabular}{@{}l@{}} Multiple layers \\of MPNN, \\ LSTM, MLP\end{tabular} & 22 405* & \begin{tabular}{@{}l@{}} Temporal and spatial \\ dependencies \\ Mobility features \end{tabular} & \begin{tabular}{@{}l@{}} Large amount of  \\ data for mobility \\ Large number of \\ internal parameters  \end{tabular}  & \begin{tabular}{@{}l@{}}MAE \\ Relative \\ error\end{tabular} & \begin{tabular}{@{}l@{}}COVID-19 cases, and \\ Data for Good mobility \\ for Italy, England, Spain \\ and France (105/129/35 \\ and 81 regions) \end{tabular}  \\
        \\[-0.25em] 
        GNN \textit{ZIP} \cite{fritz22} & \begin{tabular}{@{}l@{}} Distributional \\ regression \\ (ZIP and NB), \\ multiple layers \\ of GCN, MLP\end{tabular} & \begin{tabular}{@{}l@{}} Not \\ indicated  \end{tabular} & \begin{tabular}{@{}l@{}}Spatial dependencies \\ Mobility structures \\  Deep ensembles for \\ uncertainty \end{tabular} & \begin{tabular}{@{}l@{}} Large amount of  \\ mobility data  \\ Limited temporal \\ information  \end{tabular}  & RMSE & \begin{tabular}{@{}l@{}}COVID-19 cases, \\ Facebook mobility \\ for Germany \\ (401 districts) \end{tabular}  \\
        \\[-0.25em] 
         GNN \cite{kap20} & \begin{tabular}{@{}l@{}} Multiple layers \\ of GCN, MLP\end{tabular} & \begin{tabular}{@{}l@{}} Not \\ indicated  \end{tabular} & \begin{tabular}{@{}l@{}} Temporal and spatial \\ dependencies through  \\ edge attributes \\ Mobility structures \end{tabular} & \begin{tabular}{@{}l@{}} Large amount of \\ mobility data \\ Limited temporal \\ information \end{tabular}  & \begin{tabular}{@{}l@{}}PCC \\ RMSLE \end{tabular} & \begin{tabular}{@{}l@{}}COVID-19 NYT cases, \\ Google Aggregated Mobility \\ Research, and Google \\ Community Mobility Reports \\ for US (20 counties) \end{tabular}  \\
        \\[-0.25em] 
        STEP \cite{yu23} & \begin{tabular}{@{}l@{}} Multiple layers \\ of GCN with \\ Attention, GRU \\ MLP \end{tabular} & \begin{tabular}{@{}l@{}} Not \\ indicated  \end{tabular} & \begin{tabular}{@{}l@{}} Temporal and spatial \\ dependencies, attention \\ mechanisms for edge \\ importance, additional \\ attributes conditions, \\ SIR features \end{tabular} & \begin{tabular}{@{}l@{}} Large amount of \\  data for population\\ attributes \end{tabular}  & \begin{tabular}{@{}l@{}}MAE \\ RMSE \end{tabular} & \begin{tabular}{@{}l@{}}COVID-19 NHC, FCI NYT, \\ US population statistics, \\ Google maps, NOAA \\ weather for US (51 states) \end{tabular}  \\
         \\[-0.25em]
        GNN \cite{mur21} & \begin{tabular}{@{}l@{}}Multiple layers \\ of MLP, RNN, \\ and GAT\end{tabular} & 6 698*  & \begin{tabular}{@{}l@{}} Temporal and spatial \\ context \\ Mobility and SIR features \\ Attention mechanisms for \\ varying edge importance \end{tabular} & \begin{tabular}{@{}l@{}} Large amount of  \\ mobility data   \\ Complexity increases  \\ with graph size and \\ features \end{tabular}  & \begin{tabular}{@{}l@{}}PCC \\ MSE\end{tabular} & \begin{tabular}{@{}l@{}}COVID-19 cases and \\ mobility for Spain \\ (52 provinces)\end{tabular}  \\
        \\[-0.25em]
        Cola-GNN \cite{den20} & \begin{tabular}{@{}l@{}} Multiple layers \\ of RNN, CNN, \\ Attention, GNN \end{tabular} & 3 000* & \begin{tabular}{@{}l@{}} Temporal and spatial \\ dependencies \end{tabular} & \begin{tabular}{@{}l@{}} Large amount of \\ influenza data \end{tabular}  & \begin{tabular}{@{}l@{}}PCC \\ RMSE \\ MAE \end{tabular} & \begin{tabular}{@{}l@{}}Influenza statistics IDWR \\ for Japan (47 prefectures), \\ ILI CDC for US (49 states), \\ ILINet of US-HHS for US \\ (10 HHS regions)  \end{tabular}  \\
        \\[-0.25em] 
        STAN \cite{gao21} & \begin{tabular}{@{}l@{}} Multiple layers \\ of GAT, GRU, \\ and MLP\end{tabular} & \begin{tabular}{@{}l@{}} Not \\ indicated  \end{tabular} & \begin{tabular}{@{}l@{}} Temporal and spatial \\ dependencies with SIR \\ features, User data \\ structures, attention \\ mechanisms for varying \\ edge importance \end{tabular} & \begin{tabular}{@{}l@{}} Large amount of \\ user data, locations \\ with 1000 confirmed \\ cases or more  \end{tabular}  & \begin{tabular}{@{}l@{}}MAE \\ MSE \\ CCC \end{tabular} & \begin{tabular}{@{}l@{}}COVID-19 JHU cases, \\ and User data IQVIA US9 \\ for US (45 states and \\ 193 counties) \end{tabular}  \\
        \\[-0.25em] 
        CausalGNN \cite{wang22} & \begin{tabular}{@{}l@{}} Multiple layers \\ of GCN with \\ attention matrix, \\ temporal encoder \\ and decoder \end{tabular} & \begin{tabular}{@{}l@{}} Not \\ indicated  \end{tabular} & \begin{tabular}{@{}l@{}} Temporal and spatial \\ dependencies, attention \\ mechanisms for varying \\ edge importance, and \\ SIRD features, ensembles \\ for uncertainty \end{tabular} & \begin{tabular}{@{}l@{}} Locations with 3000 \\ confirmed cases or \\ more, daily \\ cumulative counts, \\ specific population \\ size \end{tabular}  & \begin{tabular}{@{}l@{}}MAE \\ MAPE \end{tabular} & \begin{tabular}{@{}l@{}}COVID-19 data for 93 \\ countries, US population \\ statistics (2019), state and \\ county adjacency data for \\ US (52 states and 1351 \\ counties) \end{tabular}  \\
        \\[-0.25em] 
        \textbf{PrGLSTM} & \begin{tabular}{@{}l@{}} A total of 4 layers \\ of GCN, LSTM, \\ MLP, Stochastic \\ module \end{tabular} & 4586 & \begin{tabular}{@{}l@{}} Temporal, spatial, and \\  stochastic dependencies, \\ ensembles for uncertainty \end{tabular} & \begin{tabular}{@{}l@{}} Good quality of \\ disease data with \\ no missing data \end{tabular}  & \begin{tabular}{@{}l@{}}RMSE \\ MAE \\ MSE \\ Huber \\ RMSLE  \end{tabular} & \begin{tabular}{@{}l@{}}COVID-19 JHU cases and \\ geographical data for \\ US (3,218 counties), \\ chickenpox data for \\ Hungary (20 counties) \end{tabular}  \\
        \\[-0.75em]

        \hline
    \end{tabular}%
    }
    \footnotesize
    {
    \parbox{1\linewidth}{
    * Indicates minimum number of internal trainable parameters.\\
    \textit{Abbreviations.} PCC: Pearson correlation coefficient; RMSE: root mean squared error; MAE: mean absolute error; MAPE: mean absolute percentage error; MSE: mean squared error; CCC: concordance correlation coefficient; RMSLE: root mean squared logarithmic error; MPNNLSTM: message passing neural network long short-term memory; GNN ZIP: graph neural network with zero-inflated Poisson; GNN: graph neural network; STEP: spatio-temporal epidemic prediction framework; Cola-GNN: cross-location attention based graph neural network; STAN: spatio-temporal attention network; CausalGNN: causal-based graph neural network; MPNN: message passing neural network; LSTM: long short-term memory; MLP: multi-layer perceptron; ZIP: zero-inflated Poisson; NB: negative binomial; GCN:graph convolutional neural network; GRU: gated recurrent unit; RNN: recurrent neural network; GAT: graph attention network; CNN: convolutional neural network; SIR: susceptible, infected and recovered; SIRD: susceptible, infected, recovered and death; NYT: The New York Times; US: United States of America; NHC: National Health Commission; NOAA: National Oceanic and Atmospheric Administration; ILI: influenza-like illness; IDWR: infectious diseases weekly report; CDC: Center for Disease Control; HHS: Department of Health and Human Services; JHU: Johns Hopkins University.
    }
    }    
\end{table}

\end{document}